%% file: 2018JMLR.tex
\documentclass[twoside,11pt, dvipsnames,abbrvbib]{article}

\usepackage{header}
\usepackage{lipsum}
\usepackage{csquotes}
\usepackage{jmlr2e}

\makeatletter
\def\@opargbegintheorem#1#2#3{\trivlist
   \item[]{\bfseries #1\ #2\ (#3)} \itshape}
\makeatother

\ShortHeadings{Bandits for Correlated Markovian Environments}{Fiez, Sekar, and Ratliff}
\firstpageno{1}

\begin{document}

\title{Multi-Armed Bandits for Correlated Markovian Environments with Smoothed Reward Feedback}

\author{\name Tanner Fiez \email fiezt@uw.edu \\
       \addr Department of Electrical and Computer Engineering\\
       University of Washington\\
       Seattle, WA 98195-4322, USA
       \AND
       \name Shreyas Sekar \email ssekar@hbs.edu \\
       \addr Harvard University\\
       Cambridge, MA 02138, USA
\AND
       \name Lillian J. Ratliff \email ratliffl@uw.edu \\
       \addr Department of Electrical and Computer Engineering\\
       University of Washington\\
       Seattle, WA 98195-4322, USA
}

\editor{}

\maketitle

\begin{abstract}
\input{Main_Sections/abstract}
\end{abstract}


\section{Introduction}
\label{sec:intro}
\input{Main_Sections/introduction}

\subsection{Contributions and Organization}
\input{Main_Sections/contributions}

\subsection{Background and Related Work}
\input{Main_Sections/related_work}

\section{Preliminaries}
\label{sec:preliminaries}
\input{Main_Sections/preliminaries}

\section{Regret Analysis}
\label{sec:regretdecomp}
\input{Main_Sections/regretdecomp}

\subsection{Preliminaries for Algorithm--Based Regret Bounds}
\label{sec:alg_preliminaries}
\input{Main_Sections/algorithm_preliminaries}

\subsection{EpochUCB Algorithm Analysis}
\label{sec:ucb}
\input{Main_Sections/ucb}

\subsection{EpochGreedy Algorithm Analysis}
\label{sec:greedy}
\input{Main_Sections/greedy}

\section{Experiments}
\label{sec:experiments}
\input{Main_Sections/experiments}

\section{Conclusion and Future Work}
\label{sec:conclusion}
\input{Main_Sections/discussion}

\acks{This work is supported by National Science Foundation (NSF). Tanner Fiez was also supported in part by a National Defense Science and Engineering Graduate (NDSEG) Research Fellowship.}

\newpage
\appendix

\section{Notation Table}
This appendix contains the most important and frequently used notation in the paper. 
\label{app:notation}
\input{Appendix_Sections/notation_table}

\section{Proofs}\label{app:proofs}
This appendix contains proofs that were not included in the main body of the paper.

\subsection{Proof of Harmonic Bound}
\label{app:harmonic}
\input{Appendix_Sections/harmonic}

\subsection{Proof of Lemma~\ref{lem:discount1}}
\label{app:innerbound}
\input{Appendix_Sections/discountbound1}

\subsection{Proof of Lemma~\ref{lem:boundforAH}}
\label{app:lem:boundforAH}
\input{Appendix_Sections/discountbound2}

\subsection{Proof of Theorem~\ref{thm:regretbound}}
\label{app:thm:ucb}
\input{Appendix_Sections/ucb_proof}

\subsection{Proof of Corollary~\ref{corr:gap_independent}}
\label{app:corr:gap_independent}
\input{Appendix_Sections/gap_independent}

\subsection{Proof of Theorem~\ref{thm:greedy}}
\label{app:thm:greedy}
\input{Appendix_Sections/greedy_proof}

\section{Details of Comparison Algorithms}
\label{app:experiments}
\input{Appendix_Sections/experiments}

\section{Supplemental Discussion}
\label{app:commentary}
\input{Appendix_Sections/commentary}

\bibliography{bibtex}

\end{document}

%% file: Main_Sections/abstract.tex
We study a multi-armed bandit problem in a dynamic environment where arm
rewards evolve in a correlated fashion according to a Markov chain. 
Different than much of the work on related problems, in our formulation a learning algorithm does not have access to either \emph{a priori} information or observations of the state of the Markov chain and only observes smoothed reward feedback following time intervals we refer to as epochs. We demonstrate that existing methods such as UCB and $\vep$--greedy can suffer linear regret in such an environment. Employing mixing-time bounds on Markov chains, we develop algorithms called EpochUCB and EpochGreedy that draw inspiration from the aforementioned methods, yet which admit sublinear regret guarantees for the problem formulation. Our proposed algorithms proceed in epochs in which an arm is played repeatedly for a number of iterations that grows linearly as a function of the number of times an arm has been played in the past. We analyze these algorithms under two types of smoothed reward feedback at the end of each epoch: a reward that is the discount-average of the discounted rewards within an epoch, and a reward that is the time-average of the rewards within an epoch.

%% file: Main_Sections/introduction.tex
Online learning and the theory of multi-armed bandits play a key
role in shaping how digital platforms actively engage with users. A common theme underlying these platforms is the presence of mechanisms that target individuals with personalized options chosen from a pool of available actions. Tools from the theory of multi-armed bandits provide a systematic framework for synthesizing algorithms where a \emph{decision-maker} interacts with an \emph{agent}, by balancing \emph{exploration} (learning how agents react to new alternatives) with \emph{exploitation} (repeatedly offering the best performing options to an agent). Typical applications where such trade-offs arise include recommender systems~\citep*{liCLS10,liKG16}, crowdsourcing~\citep{liu2017online,tran2014efficient}, and incentive design in e-commerce and physical retail\footnote{Although we motivate our work with applications pertaining to interactions between a decision-maker and an agent, an example being digital platforms that actively engage with users, our framework applies to any correlated Markovian environment, e.g., spectrum access applications~\citep{tekin:2012aa}.}~\citep{chakrabartiKRU08}. 
A rich stream of literature has investigated multi-armed bandit algorithms for such scenarios, obtaining near-optimal performance guarantees for a multitude of settings~\citep[see, e.g.,][]{BubeckC12, slivkins2017introduction, lattimore2018bandit}.

Despite their popularity, most of the bandit algorithms in this line of research fail to take into account a crucial component of human interaction prevalent in the aforementioned applications. Notably, the \emph{type} or \emph{state} of an agent at any given point in time depends primarily on their underlying beliefs, opinions, and preferences; as these states evolve, so do the rewards for pursuing each distinct action. For example, there is mounting empirical evidence that humans make decisions by comparing to evolving reference points such as the status quo, expectations about the future, or past
experiences~\citep{kahneman:1984aa}. Against this backdrop, we illuminate three
critical challenges that motivate our work and render current bandit approaches
ineffective in a dynamic setting.

\begin{description}[itemsep=-2pt, topsep=2pt, leftmargin=0pt]
    \item[\emph{Correlated Evolution of Rewards}:] The evolution of an agent's
        beliefs or reference point is inextricably tied to the mechanisms that they interact with. Hence, the preferences of an agent evolve in a correlated fashion, and consequently, so do the rewards for each action a decision-maker can select. As a result, standard techniques that ignore such correlations may misjudge the rewards of each action. It is worth noting that several previous works studying online decision-making problems have identified similar dynamic feedback loops in applications such as click-through rate prediction~\citep{graepelweb} and rating systems~\citep{herbrich2007trueskill}.
    \item[\emph{Lack of State Information}:] Digital platforms that engage with
        users often observe only their actions (e.g., click or no click) that
        can act as a proxy for a reward, and rarely have access to their
        underlying beliefs, opinions, or preferences that induce
        responses. Thus, it is imperative to design learning policies that are
        unaware of the agent's state, yet which remain cognizant of the general fact that rewards are drawn from evolving distributions over the agent's state.
    \item[\emph{Batch Feedback:}] In real world systems, it is often the case
        that reward feedback is processed in batches to update a learning
        policy because of various runtime constraints 
        ~\citep[see Chapter 8 of][]{slivkins2017introduction}. 
        Moreover, immediate feedback may not reflect the long-term mean reward of an action when agent preferences are rapidly evolving.
        In such scenarios, frequently adapting a policy can further complicate the task of distinguishing between the underlying value of actions.
\end{description}

The objective of the work in this paper is to develop a principled approach
to solving bandit problems in environments where: (\emph{i}) future rewards are correlated with past actions as a result of a dependence on the agent's underlying state, (\emph{ii}) the decision-maker does not have \emph{a priori} information regarding the state nor do they observe the state during their interaction with the agent, and (\emph{iii}) reward feedback cannot be collected immediately to update a learning policy. In order to capture these features, we consider a bandit setting with $m$ arms or actions and an underlying state $\theta \in \Theta$. The policy class is restricted such that a fixed action must be played within each \emph{epoch}---a time interval consisting of a number of iterations that grows linearly as a function of the number of times an arm has been played in the past. The observed feedback at the end of an epoch is a \emph{smoothed} reward. The smoothed reward can be modeled as a \emph{discount-averaged} reward or a \emph{time-averaged} reward of the instantaneous rewards at each iteration within an epoch. In the discount-averaged reward model, rewards carry more weight toward the end of an epoch, while in the time-averaged reward model, rewards are given equal weight within an epoch. The reward for an action at each iteration within an epoch is drawn from a distribution that depends on $\theta$. Moreover, the state $\theta$ evolves in each iteration according to a Markov chain whose transition matrix depends on the arm selection. 

Although our setup is a generalization of the classic bandit problem, popular
approaches such as UCB and $\vep$--greedy~\citep{auer:2002aa} perform poorly
here, often converging to suboptimal arms; we demonstrate this in Example~\ref{ex:cuteex} of Section~\ref{sec:preliminaries}. The failure of existing algorithms stems from the correlation between actions: since the evolution of $\theta$ depends on the action played, observed rewards are a function of past actions. Finally, it is worth noting that problems with Markovian rewards are traditionally studied through the lens of reinforcement learning~\citep{jaksch:2010aa,azar2013regret}. However, \emph{all} of the algorithms in this domain assume that the decision-maker has full or partial information on an agent's state. We believe that the  state-agnostic bandit algorithms developed in this work will serve as a bridge between the traditional bandit theory and techniques from reinforcement learning.

%% file: Main_Sections/contributions.tex
Given a multi-armed bandit problem where the arm rewards evolve in a correlated
fashion according to a Markov chain, the fundamental question studied in this
work is the following:
\emph{Can we design an algorithm that provably guarantees sublinear regret as a function of the time horizon in the absence of state observations and knowledge of the way in which the reward distributions are evolving?} 

The contributions of this work are now summarized. To demonstrate the necessity of our work, we show that traditional bandit algorithms such as UCB and $\vep$--greedy can misidentify the optimal arm and suffer linear regret as a result of underestimating the value of the optimal arm in a correlated Markovian environment. We then develop a general framework for analyzing epoch-based bandit algorithms for problems with correlated Markovian rewards. 
We join this framework with theory of Markov chain mixing times and existing bandit principles to design algorithms capturing the uncertainty in the empirical mean rewards arising from the unobserved, evolving state distribution and the stochasticity of the reward distributions.
This gives us the central contributions of the paper, bandit algorithms called \emph{EpochUCB} and \emph{EpochGreedy}. We prove $\mc{O}(\log(n))$ gap-dependent regret bounds for each algorithm under discount-averaged and time-averaged reward feedback. Given that no online learning algorithm can obtain sublogarithmic regret bounds, even in the case of independent and identically distributed rewards, our algorithms produce regret bounds that are asymptotically tight. Moreover, we obtain an $\mc{O}(\sqrt{n\log(n)})$ gap-independent regret bound for EpochUCB under discount-averaged and time-averaged reward feedback.
Finally, we present a number of simulations comparing the empirical and theoretical performance of our proposed algorithms. This is augmented with a set of illustrative experiments demonstrating that our proposed algorithms empirically outperform existing bandit and reinforcement learning algorithms in correlated Markovian environments. 

We now briefly characterize the challenges involved in designing online learning algorithms when the rewards evolve in a correlated fashion and outline our techniques to overcome them. Unlike a typical bandit problem, there are two sources of uncertainty in the empirical rewards: (\emph{i}) uncertainty regarding the state, and
(\emph{ii}) uncertainty from the stochasticity of the distributions from which rewards are drawn. As we demonstrate with illustrative examples, misjudging the expected reward of arm as an artifact of failing to take into account the multiple sources of uncertainty can lead to significant regret.
Moreover, the distribution from which an arm's reward is drawn can change even
when that arm is not selected since rewards depend on the evolving state. To
overcome these obstacles, our proposed algorithms leverage techniques from the
theory of mixing times for Markov chains, while retaining the spirit of the UCB
and $\vep$--greedy algorithms developed in~\citet{auer:2002aa}. Specifically, by
characterizing the mixing times, we are able to obtain estimates of the empirical mean rewards that closely approximate the expected stationary distribution rewards and  maintain accurate confidence windows representing the uncertainty in these estimates. The execution of each algorithm in epochs of growing length ensures that the uncertainty in the empirical mean reward estimates for each arm eventually dissipates.

Following a formalization of the problem we study in Section~\ref{sec:preliminaries}, we present our proposed EpochUCB and EpochGreedy algorithms and analyze their regret in Section~\ref{sec:regretdecomp}. In Section~\ref{sec:experiments}, we present simulation results. We conclude in Section~\ref{sec:conclusion} with a discussion of open questions and comments on future work. Finally, to promote readability, an appendix comes after the primary presentation of our work. In Appendix~\ref{app:notation}, we put forward a notation table that includes the most important and frequently used notation in the paper. The majority of the proofs backing our theoretical results are contained in Appendix~\ref{app:proofs}. However, proof sketches are provided immediately following the statements of our main results. In Section~\ref{app:experiments}, further details are given on the existing algorithms that we empirically compare to our proposed algorithms.

%% file: Main_Sections/related_work.tex
The two distinct features separating our model from previous
work on multi-armed bandits with Markov chains are that the rewards evolve in a correlated fashion and the decision-maker is fully unaware of the agent's state. These features are missing in the related \emph{rested} and \emph{restless} multi-armed bandit problems. 

In the rested~\citep{anantharam:1987ab,tekin:2010aa} and
restless~\citep{tekin:2012aa,ortner:2014aa} bandit literature,
there is an independent Markov chain tied to each arm and the reward for an arm
depends on the state of the Markov chain for that arm. In each model, when an
arm is selected the state of the Markov chain for that arm is observed and
transitions. The distinguishing characteristic between the problems is the
behavior of the Markov chains tied to arms that are not selected. In the rested
bandit model, the Markov chains associated with unselected arms remain unaltered; however, in the the restless bandit model, the Markov chains associated with unselected arms evolve precisely as they would had they been played. 

Hence, as is true in our model, in the restless bandit problem the
reward distribution on an arm can evolve even when the arm is not being played.
However, a key point of distinction is that the evolution of the reward
distribution on an arm is independent of the reward distribution for every other arm. In contrast, the problem setting we study is such that
there is a shared Markov chain between arms so that the evolution of the rewards on each arm is correlated. Moreover, in the setting under consideration there are no observations of the Markov chain state or distribution. Although these distinctions may appear subtle, the correlation between present actions and future rewards, along with the absence of state observations, results in a number of technical difficulties.

An exception to the formulations of the rested and restless bandit
problems is the manuscript of~\citet{mazumdar:2017aa}, which proposes a UCB-inspired strategy for expert selection in a Markovian environment. Our proposed UCB-inspired algorithm, called EpochUCB, improves upon the regret bound in that work. Moreover, we consider more general reward feedback structures and propose an $\varepsilon$--greedy inspired algorithm called EpochGreedy. Another work along this same vein is that of~\citet{azar2013regret}; in this work, an expert selection strategy is
proposed for finding policies in Markov decision processes with partial state feedback.

The problem we study bears some conceptual similarity to the traditional \emph{principal-agent} model from contract theory~\citep{bolton2005contract, laffont2009theory}. The standard principal-agent model is a one-shot interaction: a principal selects a signal $j \in [m]$ to send to an agent with a type variable $\theta$, and then the self-interested agent pursues an action depending on $(\theta,j)$. The reward the principal obtains is a function of the action of the agent, and consequently, the parameters $(\theta,j)$. Typically, there exists an information asymmetry between the principal and the agent. Prominent examples include \emph{adverse selection} (agent type is unobservable to the principal) and \emph{moral hazard} (agent action is unobservable to the principal). Recently, repeated principal-agent problems have begun to be studied as bandit problems where each round corresponds to the standard principal-agent formulation with adverse selection and moral hazard. A notable example of such a formulation is the work of~\cite{ho2016adaptive}. The caveat of this work is that in each round a brand-new agent arrives with a type variable sampled from an i.i.d.~distribution; following an interaction with the principal, an agent exits the system forever. In contrast, the problem we study can be viewed as a repeated principal-agent problem where a unique agent continuously interacts with the principal. However, while our formulation is an example of adverse selection since the state is unobserved and dynamically evolves, we do not model the strategic nature of the agent.

Finally, we remark that our setting is general enough to model a number of existing works, which we describe below: 
\begin{enumerate}[itemsep=-2pt, topsep=2pt]
\item \emph{Bandits with Positive Externalities}: The state $\theta$ could represent the agent's bias toward actions that have yielded high reward in the past as in~\citet{shahexternality}. The decision-maker receives a higher expected reward by selecting actions that the agent is positively predisposed toward.

\item \emph{Bandits Tracking Arm Selection History}: In recharging bandits~\citep{immorlicarecharging}, the reward on an arm is an increasing function of 
the time since it was last selected. The state $\theta$ could track such a time period. Along the same lines, in rotting bandits~\citep{levineCM17}, the reward on an arm is a decreasing function of the number of times it has been played in the past. The state $\theta$ could track the number of plays of each arm. 
\end{enumerate}

%% file: Main_Sections/preliminaries.tex
We now formalize the problem we study, detail technical challenges, and present an example demonstrating the insufficiency of existing algorithms that necessitates our work.

\subsection{Problem Statement}
Consider a decision-maker that faces the problem of repeatedly choosing an action to impact an agent over a finite time horizon. We refer to the set of possible actions as arms and use the notation $[m] = \{1,\ldots,m\}$ to index them. The agent is modeled as having state $\theta\in \Theta$, where $\Theta$ is a finite set, that evolves in time according to a Markov chain whose transition matrix depends on the arm selected by the decision-maker. In turn, the agent's state influences the rewards the decision-maker receives. The goal of the decision-maker is to construct a policy that sequentially selects arms in order to maximize the cumulative reward over a finite horizon.

We restrict the decision-maker's policy class to a specific type of multi-armed bandit algorithm that we refer to as an \emph{epoch mixing policy}. Such a policy $\alpha$ is executed over epochs indexed by $[n] = \{1,\ldots,n\}$. In each epoch $k \in [n]$, the policy selects an arm $\alpha(k) \in [m]$ and repeatedly `plays' this arm for $\taukr > 0$ iterations within this \emph{epoch}---where we use the superscript $\alpha$ to indicate that the epoch length depends on the arm selected---before receiving feedback in the form of a reward. When the decision-maker makes an arm choice at an epoch $k \in [n]$, the state of the agent $\theta$ evolves for $\taukr$ iterations. The reward the decision-maker observes at the end of the epoch is a function of the rewards at each iteration within the epoch. In short, an epoch mixing policy proceeds on two time scales: each selection of an arm corresponds to an epoch $k\in [n]$ that begins at time $t_k$ and ends at time $t_{k+1} = t_k + \taukr$ following $\taukr$ iterations. Our motivation for restricting the policy class in this way stems from the inherent challenges online platforms face to process immediate feedback in order to update learning algorithms frequently, and the necessity of observing feedback based upon periods of near-stationarity to obtain meaningful regret bounds.

Returning to the agent model, the agent's state $\theta$ is modeled as the state of a Markov chain. Let $\beta_{t_k}: \Theta \rar [0, 1]$ denote the state distribution at time $t_k$ and $\theta_{t_k}$ denote the random variable representing the agent's state at time $t_k$ having distribution $\beta_{t_k}$. Given that arm $\alpha(k) = j$ is selected at epoch $k \in [n]$, let the arm-specific transition matrix of the Markov chain be denoted as $P_{j}:\Theta\times\Theta\rightarrow [0,1]$. Then, the state distribution on each $\theta \in \Theta$ evolves between epochs $k$ and $k+1$ as follows:
\begin{equation*}
\textstyle \beta_{t_{k+1}}(\theta)=\sum_{\theta'\in \Theta} P_{j}^{\taukr}(\theta',\theta)\beta_{t_k}(\theta'),
\end{equation*}
where $P_{j}^{\taukr}(\theta', \theta)$ is the probability of the state
transition from $\theta'$ to $\theta$ in $\taukr$ iterations---that is,
$P_{j}^{\taukr}$ is the $\taukr$ composition of $P_{j}$. Observe that the agent model we adopt captures the fact that the agent's preferences depend on past interactions with the decision-maker since the state distribution when an epoch begins depends on previous epochs.
\begin{assumption}\label{assumption:ergodic}
For each arm $j\in [m]$, the transition matrix $P_j$ is irreducible and aperiodic.
\end{assumption}
Irreducible and aperiodic Markov chains are \emph{ergodic}.
Assumption~\ref{assumption:ergodic} implies that for each $j\in[m]$, the Markov
chain characterized by $P_j$ has a unique positive stationary distribution that we denote by $\pi_j:\theta\mapsto (0,1]$. 
Furthermore, let $\tilde{P}_j$ denote the \emph{time reversal} of $P_j$, defined as $\tilde{P}_j(\theta,\theta')=\frac{\pi_j(\theta')P_j(\theta',\theta)}{\pi_j(\theta)} \ \forall \ \theta, \theta' \in \Theta$. The time reversal matrix $\tilde{P}_j$ is also irreducible and aperiodic with unique positive stationary distribution $\pi_j$. Define the \emph{multiplicative reversiblization} of $P_j$ to be $M(P_j)=P_j\tilde{P_j}$, which is itself a reversible transition matrix. The eigenvalues of $M(P_j)$ are real and non-negative so that the second largest eigenvalue $\lambda_2(M(P_j))\in[0,1]$~\citep{fill:1991aa}.
\begin{assumption}\label{assumption:reverse}
For each arm $j\in [m]$, the transition matrix $P_j$ is such that $M(P_j)$ is irreducible.
\end{assumption}
This is a standard assumption in the bandit literature with Markov chains~\citep[see][and the references therein]{tekin:2012aa} that implies $\lambda_2(M(P_j))\in[0,1)$. Recall that Assumption~\ref{assumption:ergodic} on the transition matrices ensures $\lambda_2(M(P_j)) \in [0, 1]$. Hence, Assumption~\ref{assumption:reverse} on the transition matrices only restricts the boundary case when $\lambda_2(M(P_j))=1$. The assumption is necessary to derive meaningful bounds on the deviation between the expected reward feedback and the expected stationary distribution reward of an arm.

\begin{remark}
A permissive sufficient condition on an ergodic transition matrix $P$ that ensures $M(P)$ is irreducible is $P(\theta, \theta) > 0 \ \forall \ \theta \in \Theta$~\citep{tekin:2012aa}. This condition holds naturally for a large class of applications. A much more restrictive, yet still relevant sufficient condition on an ergodic transition matrix $P$ certifying that $M(P)$ is irreducible would be that $P$ is also reversible. We emphasize that each of the sufficient conditions are not necessary conditions for Assumption~\ref{assumption:reverse}.
\end{remark}

\noindent
\textbf{Reward Models:} The reward the decision-maker receives is dependent on the state of the agent. Let $\theta_k = \{\theta_t\}_{t=t_k}^{t_{k+1}-1}$ be the sequence of random state variables in epoch $k \in [n]$ and let $\br{k}{\alpha(k)}$ be the observed reward at the end of the epoch, where $\alpha(k)$ and $\theta_k$ denote the dependence on the arm selected and the state, respectively. Similarly, let $\nr{t}{\alpha(k)}$ denote the instantaneous reward at an iteration $t\in[t_k,t_{k+1}-1]$ during epoch $k \in [n]$. While a number of different functions of the instantaneous rewards in each epoch could be considered, we restrict our attention to the case where $\br{k}{\alpha(k)}$ is a \emph{smoothed} reward over the epoch. We consider the observed smoothed reward to be a \emph{discount-averaged} or \emph{time-averaged} reward of the instantaneous rewards in an epoch. For a discount factor $\gamma \in (0, 1)$ selected by the decision-maker, the discount-averaged reward is defined as
\begin{equation*}
\textstyle \br{k}{\alpha(k)}=\frac{1}{\bsgr_k}\sum_{t=t_k}^{t_{k+1}-1}(\gamma)^{t_{k+1}-1-t}\nr{t}{\alpha(k)},
\end{equation*}
where 
\begin{equation*}
\textstyle \bsgr_k = \sum_{t=t_k}^{t_{k+1}-1}(\gamma)^{t_{k+1}-1-t} 
\end{equation*} 
denotes the sum of the discount factors in the epoch. In the special case that
the discount factor $\gamma = 1$, the observed reward is a time-averaged reward:
\begin{equation*}
\textstyle \br{k}{\alpha(k)} = \frac{1}{\bsgr_k}\sum_{t=t_k}^{t_{k+1}-1}\gamma^{t_{k+1}-1-t}\nr{t}{\alpha(k)} =\frac{1}{\taukr}\sum_{t=t_k}^{t_{k+1}-1}\nr{t}{\alpha(k)}.
\label{eq:time_averaged}
\end{equation*}
The rewards at an iteration $\nr{t}{\alpha(k)}$ are assumed bounded; without loss of generality, $\nr{t}{\alpha(k)}\in[0,1]$. Moreover, they are stochastic with stochastic kernel $\mc{T}_r(\theta, \alpha(k))\in
\mc{P}([0,1])$ such that  $\nr{t}{\alpha(k)}\sim \mc{T}_r(\theta_t, \alpha(k))$ and where
$\mc{P}([0,1])$ denotes the space of probability distributions on $[0,1]$.

\begin{remark}
Observe that when $\gamma \in (0, 1)$, the discount factors are growing within an epoch so that rewards are given more weight toward the end of an epoch, and when $\gamma=1$, the rewards within an epoch are given equal weight. 
This general framework allows us to model a variety of objectives. 
For instance, agents are likely to have recency bias and hence, if a decision-maker's instantaneous reward depends on some measure of agent happiness or satisfaction, then discounting rewards over the epoch is reasonable. On the other hand, if the decision-maker's instantaneous reward measures revenue or profit, then equally weighting all rewards accrued in an epoch is reasonable.
\end{remark}

Given Assumption~\ref{assumption:ergodic}, if an arm $j \in [m]$ is chosen at every iteration, then the Markov chain would eventually converge toward its stationary distribution $\pi_j$. This would, in turn, give rise to a fixed reward distribution. We define the expected stationary distribution reward $\mu_j$ for arm $j\in[m]$ to be
\begin{equation*}
\textstyle \mu_j=\mb{E}\big[\sum_{\theta\in \Theta} r^\theta_j\pi_j(\theta)\big],
\end{equation*}
where the expectation is with respect to $\mc{T}_r(\theta, j)$. Likewise,
we define the \emph{optimal arm}, indexed by $j_\ast\in [m]$, and denoted as $\ast$ when used in a subscript, to be the arm that yields the highest expected reward $\mu_j$ from its stationary distribution $\pi_j$. Hence, the expected reward of the optimal arm $j_\ast$ is 
\begin{equation*}
\textstyle   \mu_\ast=\max_j\mb{E}\big[\sum_{\theta\in \Theta}
    r^\theta_j\pi_j(\theta)\big].    
\end{equation*}


We use a notion of \emph{regret} as a performance metric
that compares the cumulative expected reward over a finite horizon of a benchmark policy and that of the policy $\alpha$. The benchmark policy we compare to is the best fixed arm in hindsight on the stationary distribution rewards. That is, we compare to the policy that plays the optimal arm $j_\ast$ at each epoch and receives rewards drawn from its stationary distribution.
\begin{definition}[Cumulative Regret]
The cumulative regret after $n$ epochs of policy $\alpha$ is given by
\begin{equation}
\textstyle   R^\alpha(n)=n\mu_\ast-\mb{E}\big[\sum_{k=1}^{n}\br{k}{\alpha(k)}\big],
\label{eq:cum_regret}
\end{equation}
where the expectation is with respect to the random draw of the rewards
through $\mc{T}_r(\theta, \alpha(k))$, arms selected by the decision-maker
using $\alpha$, and the state $\theta$.
\label{def:regret}
\end{definition}

\subsubsection{Discussion of Regret Notion}
Let us briefly comment on the regret notion we consider. It is worth noting that the benchmark policy being compared to is the optimal policy within the policy class that is restricted to a fixed arm being played. In general, however, the globally optimal policy for a given problem instance may not always play a fixed arm at each epoch. Simply put, the globally optimal policy may select an arm dependent on the state of the Markov chain. In fact, we would expect the globally optimal policy to be a deterministic policy in each state. Meaning that, conditioning on the state, the globally optimal policy would play the best arm for that state. This is because in the full information case, where the decision-maker observes the initial state distribution, the dynamics, etc., the decision-maker simply faces a Markov decision process---which are known to have deterministic state-dependent optimal policy~\citep{bellman1957dynamic}. Of course, since in our problem the state is fully unobserved and no prior on the distribution is available, finding such a policy is infeasible. Owing to this basis, measuring the regret with respect to the best fixed arm in hindsight on the stationary distribution rewards is standard in multi-armed bandit problems with Markov chains~\citep[see, e.g.,][]{tekin:2010aa, tekin:2012aa, gai:2011aa}. The regret notion we adopt---comparing to the best fixed arm in hindsight when the globally optimal policy may not always play a fixed arm---is often referred to as
\emph{weak regret}~\citep{auer2002nonstochastic}.

\subsection{Details on Technical Challenges and Insufficiency of Existing Methods}
The key technical challenge stems from the dynamic nature of the reward distributions. Indeed, the rewards depend on an underlying state distribution which is common across arms; the initial distribution of the Markov chain when an arm is pulled is the distribution at the end of the preceding arm pull. The consequences of the evolving nature of the state distribution are two-fold: (\emph{i}) the reward distribution on any given arm can evolve even when the arm is not being played by the algorithm and (\emph{ii}) the fashion in which the reward distribution on each arm evolves when not being played depends on the current arm being played by the algorithm. That is, future reward distributions on each arm are \emph{correlated} with the present actions of an algorithm. The manner in which the reward distributions evolve is precisely where the problem deviates from the rested and restless bandit problems and becomes more challenging.

Since rewards are \emph{dependent} on the algorithm, i.i.d.~reward assumptions, such as those found in the stochastic bandit problem, fail to hold. Despite this, a natural question may be whether or not naively employing algorithms from this literature, such as UCB and $\vep$--greedy, is sufficient in a correlated Markovian environment. We consider a simple example that indicates it is not. 
\begin{example}[Failure of UCB and $\vep$--greedy]
\label{ex:cuteex}
Consider a problem instance with two arms $[m]=\{1,2\}$ and two states $\Theta=\{\theta_1, \theta_2\}$. The state transition matrix and reward structure for each arm are depicted in~Figure~\ref{fig:M1}. We assume that $\epsilon > 0$ is a sufficiently small constant. The stationary distribution for arm $1$ is given by $\pi_1(\theta_1) = \epsilon/(\epsilon+1) \approx 0$ and $\pi_1(\theta_2) = 1/(1+\epsilon) \approx 1$, meaning at the stationary distribution of arm $1$ the state is $\theta_2$ almost surely, and vice-versa for arm $2$. The deterministic reward $r_j^{\theta_i}$ for each (arm, state) pair $(j,\theta_{i})$ with $j \in [m]$ and $i \in \{1, 2\}$ is provided under the state.  Clearly, the optimal strategy is to play arm $1$ repeatedly to obtain a per arm selection reward of nearly one. 
\begin{figure}[h]
\centering
\subfloat[][]{
\input{Figs/arms}
\label{fig:M1}}
\subfloat[][]{\includegraphics[width=0.33\textwidth]{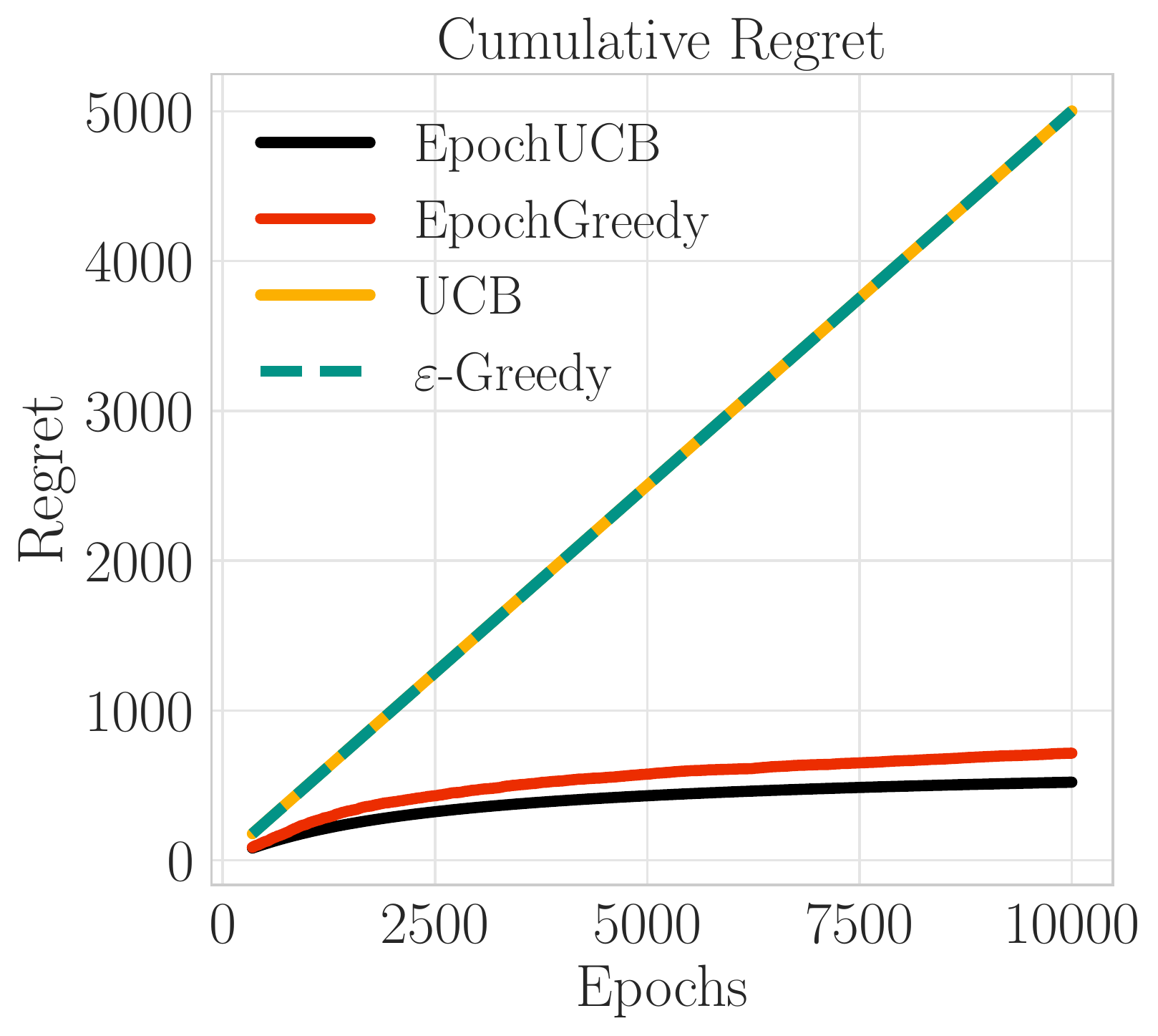}
\label{fig:badex}}
\caption{(a) State transition diagram and reward structure for each arm. (b) Regret for UCB and $\vep$--greedy versus our proposed algorithms EpochUCB and EpochGreedy. }
\end{figure}

Suppose that the initial state distribution is given by $\beta_1 = [1, 0]$.
Every time UCB and $\vep$--greedy play arm $1$, the agent is in state $\theta_1$ with high probability; consequently the reward for arm $1$ is estimated to be
close to zero. Therefore, UCB and $\vep$--greedy underestimate the reward for arm $1$ and misidentify arm $2$ as the optimal arm since the agent almost
always remains in state $\theta_1$ as a result of the induced Markov chain. 
Simulations support this finding as demonstrated in~Figure~\ref{fig:badex}. Indeed, UCB and $\vep$--greedy converge to the suboptimal arm and suffer linear regret. In contrast, our proposed algorithms, EpochUCB (Section~\ref{sec:ucb}) and EpochGreedy (Section~\ref{sec:greedy}), identify the optimal arm rapidly and incur only sublinear regret.


\end{example}

%% file: Figs/arms.tex
\begin{tikzpicture}[font=\sffamily, scale=0.8, every node/.style={transform shape}]
	
	\node[state,
	text=yellow,
	draw=none,
	fill=gray!50!black] (s) {$\theta_1$};
	\node[state,
	right=1.5cm of s,
	text=blue!30!white, 
	draw=none, 
	fill=gray!50!black] (r) {$\theta_2$};
	
    \node[below=0.0cm  of s] (som) {$r_1^{\theta_1}=0$}; 
		\node[below right=0.1cm and -0.6cm of r] 
        (som) {$r_1^{\theta_2}=1$}; 
	
	\draw[every loop,
	auto=right,
	line width=1mm,
	>=latex,
	draw=orange,
	fill=orange]
	(s) edge[bend right, auto=left]  node {1} (r)
	(r) edge[bend right, auto=right] node {$\epsilon$} (s)
    (s) edge[loop above]             node[left=0.1cm] {0} (s)
    (r) edge[loop above ]             node[left=0.1cm] {$1-\epsilon$} (r);
    \node[left=0.5cm of s] {Arm 1};
    \node[state,
text=yellow,
draw=none,
fill=gray!50!black, below=1.5cm of s] (s) {$\theta_1$};
\node[state,
right=1.5cm of s,
text=blue!30!white, 
draw=none, 
fill=gray!50!black] (r) {$\theta_2$};
\node[below =0.0cm of s] (som) {$r_2^{\theta_1}=0.5$}; 
\node[below =0.0cm of  r] (som) {$r_2^{\theta_2}=0.5$}; 

\draw[every loop,
auto=right,
line width=1mm,
>=latex,
draw=orange,
fill=orange]
(s) edge[bend right, auto=left]  node {$\epsilon$} (r)
(r) edge[bend right, auto=right] node {1} (s)
(s) edge[loop above]             node[left=0.1cm] {$1-\epsilon$} (s)
(r) edge[loop above]             node[left=0.1cm] {0} (r);
\node[left=0.5cm of s] {Arm 2};

\end{tikzpicture} 

%% file: Main_Sections/regretdecomp.tex
We begin this section by deriving a general framework for analyzing the regret of any multi-armed bandit policy interacting with a correlated Markovian environment in which the observed feedback is a smoothed (discount-averaged or time-averaged) reward over an epoch. We then introduce our proposed EpochUCB algorithm and present gap-dependent and gap-independent regret bounds for both discount-averaged and time-averaged reward feedback. This is followed by an exposition of our proposed EpochGreedy algorithm. For EpochGreedy, we prove a gap-dependent regret bound for both discount-averaged and time-averaged reward feedback.

\subsection{Regret Decomposition}\label{sec:regret_decomp_subsection}
In this section, we derive a regret bound for a generic policy $\alpha$ in terms of the expected number of plays of each suboptimal arm. While such regret decompositions are typical in the bandit literature, our bound is novel. This is owing to the fact that we need to employ results on the mixing times of Markov chains to decompose the regret into components arising from the selection of a suboptimal arm and that coming from the misalignment of the agent's state distribution with the stationary distribution.

Define $T_j^\alpha(n)=\sum_{k=1}^{n}I\{\alpha(k)=j\}$ to be the random variable
representing the number of epochs in which arm $j\in [m]$ was selected by algorithm $\alpha$ in the initial $n$ epochs. We use $I\{\cdot\}$ to denote the indicator function, meaning that $I\{\alpha(k)=j\}=1$ when $\alpha(k)=j$ and $I\{\alpha(k)=j\}=0$ when $\alpha(k)\neq j$. Moreover, observe that $\sum_{j\in[m]}T_j^\alpha(n)=n$. Our goal is to relate $\mb{E}_{\alpha}[T_j^\alpha(n)]$, where $\mb{E}_\alpha$ emphasizes the randomness in the algorithm and the rewards, to the regret $R^\alpha(n)$. Toward this end, define $\Delta_j=\mu_\ast-\mu_j$ for each $j\in [m]$ to be the \emph{reward gap}. We can add and subtract $\sum_{j\in [m]}\mb{E}_{\alpha}[T^\alpha_j(n)]\mu_j$ into~\eqref{eq:cum_regret} to obtain
\begin{align}
R^\alpha(n)&=\textstyle n\mu_\ast-\sum_{j\in[m]}\mb{E}_{\alpha}[T^\alpha_j(n)]\mu_j+\sum_{j\in [m]}\mb{E}_{\alpha}[T^\alpha_j(n)]\mu_j-\mb{E}\big[\sum_{k=1}^{n}\br{k}{\alpha(k)}\big]\notag\\
&\textstyle=\sum_{j\in [m]}\mb{E}_{\alpha}[T^\alpha_j(n)](\mu_\ast-\mu_j)+\mb{E}_{\alpha}\big[\sum_{k=1}^{n}\sum_{j\in[m]}I\{\alpha(k)=j\}\mu_j\big]\notag\\
&\mkern235mu\textstyle-\mb{E}\big[\sum_{k=1}^{n}\sum_{j\in [m]}I\{\alpha(k)=j\}\br{k}{j}\big]\notag\\
&=\textstyle\sum_{j\neq j_\ast}\mb{E}_{\alpha}[T_j^\alpha(n)]\Delta_j+\mb{E}\big[\sum_{k=1}^{n}\sum_{j\in [m]}I\{\alpha(k)=j\}(\mu_j-\br{k}{j})\big].
\label{eq:regret1}
\end{align}
Compared to the regret decomposition for the stochastic multi-armed bandit problem, which has the form $R^{\alpha}(n) = \sum_{j\neq j_\ast}\mb{E}_{\alpha}[T_j^\alpha(n)]\Delta_j$, the dynamic nature of the reward distributions in the problem leads \emph{any} algorithm to incur an additional regret penalty through the term
\begin{equation}
\textstyle\mb{E}\big[\sum_{k=1}^{n}\sum_{j\in[m]}I\{\alpha(k)=j\}(\mu_j-\br{k}{j})\big].
\label{eq:extraregret}
\end{equation}
Intuitively, this regret term is capturing the fact that the expected reward for selecting an arm at any given epoch can potentially deviate from the expected stationary distribution reward of the arm in an unfavorable way when the state distribution is not at the stationary distribution. 
\begin{example}[Markovian Regret Penalty]
Consider an optimal arm $j_\ast \in [m]$ with two states $\Theta = \{\theta_1, \theta_2\}$, stationary distribution given by $\pi_\ast(\theta_1) = \epsilon$ and $\pi_\ast(\theta_2) = 1-\epsilon$ where $\epsilon > 0$ is a small constant, and deterministic state-dependent rewards given by $r_\ast^{\theta_1} = 0$ and $r_\ast^{\theta_2} = 1$, so that the expected stationary distribution reward for the arm is nearly one. Moreover, suppose that the initial state distribution of a problem instance is given by $\beta_1 = [1, 0]$. The expected reward of arm $j_\ast$ is close to zero in the initial epoch for this problem instance, implying that the regret penalty for selecting arm $j_\ast$ in the epoch is almost one \emph{despite the reward gap being zero}. This example highlights precisely what~\eqref{eq:extraregret} elucidates in the regret decomposition: an arm selection can yield regret beyond the reward gap $\Delta_j$ when the state distribution departs from the stationary distribution of the chosen arm. We often refer to the regret term in~\eqref{eq:extraregret} as the \emph{Markovian regret penalty}.
\end{example}

In order to bound the Markovian regret penalty, we need some technical machinery. Let $\car{i}{j}$ be the reward received when arm $j \in [m]$ is chosen for the $i$--th time, where we include $\theta$ in the subscript to note the state dependence of the random reward. 
For each arm $j \in [m]$, define the $i$--th filtration: 
\begin{equation*}
\ft{j}{i}=\sigma(\car{1}{j}, \ldots, \car{i}{j},\thetarv{t_j^1}, \ldots, \thetarv{t_j^{i}}),
\label{eq:filtration}
\end{equation*}
where $t_j^i$ is the time instance at which arm $j \in [m]$ is chosen for the $i$-th time. That is, $\ft{j}{i}$ is the smallest $\sigma$-algebra generated by the random variables $(\car{1}{j}, \ldots, \car{i}{j},\thetarv{t_j^1}, \ldots, \thetarv{t_j^{i}})$. 
From the tower property of expectation, we have $\mb{E}\big[\sum_{k=1}^{n}\sum_{j\in[m]}I\{\alpha(k)=j\}(\mu_j-\br{k}{j})\big]$
\begin{align}
&=\textstyle\mb{E}_{\alpha}\big[\sum_{k=1}^{n}\sum_{j \in [m]}I\{\alpha(k)=j\}\mb{E}\big[\mu_j-\br{k}{j}\big|\ft{j}{T_j^{\alpha}(k)-1}\big]\big] \notag \\
&\leq\textstyle\mb{E}_{\alpha}\big[\sum_{k=1}^{n}\sum_{j \in [m]}I\{\alpha(k)=j\}\big|\mb{E}\big[\mu_j-\br{k}{j}\big|\ft{j}{T_j^{\alpha}(k)-1}\big]\big|\big].\label{eq:prelemma1}
\end{align}

Prior to continuing to bound the Markovian regret penalty, we introduce the epoch sequence $\{\taukr\}_{k=1}^n$ considered in this work.
Recall that $\alpha(k)$ represents the arm selected by the policy $\alpha$ at the beginning of epoch $k\in [n]$ and $T^{\alpha}_{\alpha(k)}(k-1)$ is the number of times this arm has been selected in previous epochs. At each epoch $k \in [n]$, the policy-dependent epoch length is 
\begin{equation}
 \taukr= \tauz+ \zeta T^{\alpha}_{\alpha(k)}(k-1),
 \label{eq:tauk}
\end{equation}
where $\tauz, \zeta\in \mb{Z}_{+}$ are constants selected by the decision-maker. We also use the notation $\taukj=\tauz+\zeta T_j^\alpha(k-1)$ to denote the epoch length when $\alpha(k)=j$ at epoch $k \in [n]$. It is important to recognize that the length of each epoch is random owing to the dependence on not only the epoch index, but also on the identity of the arm selected in the epoch. This is a reasonable model since a learning strategy should only be altered for an arm as a result of acquiring more information about the arm. The sequence $\{\taukr\}_{k=1}^n$ ensures that as an arm is repeatedly selected and the confidence in the expected stationary reward of the arm grows, so does the length of each epoch when the arm is selected. Consequently, once highly suboptimal arms are discarded, each epoch contains sufficiently many iterations to guarantee that the observed rewards closely approximate the stationary distribution rewards---this is crucial for discriminating between the optimal arm and nearly optimal arms.  
We remark that epochs of a fixed duration would lead to a regret bound that is linear in the time horizon under our analysis. This provides theoretical justification beyond Example~\ref{ex:cuteex} on the insufficiency of existing bandit algorithms for this problem. Informally, this is because algorithms that do not play arms an increasing number of times by design may never push the state distribution toward a stationary distribution and the rewards drawn from a distribution misaligned with a stationary distribution could be highly suboptimal. This observation serves as further motivation for the feedback model we study apart from relevant applications. More detail is provided on this point in Appendix~\ref{app:commentary}.

We now return to deriving a bound on the Markovian regret penalty. To do so, we adopt tools from the theory of Markov chains. Indeed, we need a classic result about the convergence rates of Markov chains.

\begin{proposition}[\citet{fill:1991aa}]Let $P$ be an irreducible and aperiodic transition matrix on a finite state space $\Theta$ and $\pi$ be the stationary distribution. Define the chi-squared distance from stationary at time $n$ as $\chi^2_n=\sum_{\theta}(\pi_n(\theta)-\pi(\theta))^2/\pi(\theta)$, where $\pi_n=\sum_{\theta}P^n(\theta, \cdot)\pi_0(\theta)$ and $\pi_0$ is the initial distribution of the Markov chain. 
Then, $4\|\pi_n-\pi\|^2\leq \chi^2_0(\lambda_2(M(P)))^n$.
Furthermore, 
\begin{equation}
\textstyle\max_{\pi_0\in \mc{P}(\Theta)}\big\|\sum_{\theta}P^n(\theta,\cdot)\pi_0(\theta)-\pi(\cdot)\big\|^2  \leq \textstyle\frac{1}{4}\big(1+\frac{(1-\min_\theta \pi(\theta))^2}{\min_\theta \pi(\theta)}\big)(\lambda_2(M(P)))^n,
\label{eq:boundoverall}
\end{equation}
where $\mc{P}(\Theta)$ is the space of probability distributions on $\Theta$.
\label{prop:convergence}
\end{proposition}
Noting that $\chi^2_0$ is always bounded above by $1+(1-\min_\theta \pi(\theta))^2(\min_\theta \pi(\theta))^{-1}$, Equation~\ref{eq:boundoverall} is easily derived. 
\begin{remark}
Proposition~\ref{prop:convergence} provides a bound certifying that the state distribution of a Markov chain with an ergodic transition matrix $P$ will converge toward its stationary distribution at least at a geometric rate in $\lambda_2(M(P))$ when $\lambda_2(M(P)) \in [0, 1)$. Recall that when $M(P)$ is irreducible, $\lambda_2(M(P)) \in [0, 1)$. 
\label{remark:r1}
\end{remark}

The ensuing lemma translates Proposition~\ref{prop:convergence} into a bound on the deviation of the expected reward of an arm selection from the expected stationary distribution reward of the arm; the deviation decays geometrically as a function of the epoch length. Beforehand, for each arm $j \in [m]$, define the following constants: 
\begin{equation*}
\lambda_j=(\lambda_2(M(P_j)))^{1/2}, \quad \eta_j = \min\{\gamma, \lambda_j\},\quad \phi_j = \max\{\gamma, \lambda_j\}, \quad \psi_j = \eta_j/\phi_j.
\label{eq:constants}
\end{equation*}
\begin{lemma}[Convergence of Expected Reward to Expected Stationary Reward]
Suppose Assumptions~\ref{assumption:ergodic} and~\ref{assumption:reverse}  hold and $\alpha(k)=j \in [m]$ at epoch $k\in[n]$. Then, 
\begin{equation*}
\textstyle \big|\mb{E}[\mu_j-\br{k}{j}\big|\ft{j}{T_j^{\alpha}(k)-1}\big]
\big|\leq\frac{C_j\uj{j}{\taukj}}{\bsgr_k},
\end{equation*}
where 
\begin{equation*}
\textstyle C_j =  1/2(1+(1-\min_{\theta}\pi_j(\theta))^2/\min_{\theta}\pi_j(\theta))^{1/2}
\end{equation*}
and $\uj{j}{\taukj}$ is defined as follows depending on the type of reward feedback:
\begin{enumerate}[topsep=0pt,itemsep=-2pt]
\item Discount-Averaged Reward Feedback: $\gamma\in(0,1)$.
\begin{equation}
\uj{j}{\taukj} = 
\begin{cases} 
\frac{(\phi_j)^{\taukj-1}(1 - (\psi_j)^{\taukj})}{(1-\psi_j)}, & \text{if} \
\gamma \neq \lambda_j \\
(\phi_j)^{\taukj-1}\taukj, & \text{otherwise} 
\end{cases}.
\label{eq:upsilon_discount}
\end{equation}
\item Time-Averaged Reward Feedback: $\gamma=1$. 
\begin{equation}
\textstyle\uj{j}{\taukj}=\frac{1-(\lambda_j)^{\taukj}}{1-\lambda_j}.
\label{eq:upsilon_time}
\end{equation}
\end{enumerate}
\label{lem:discount1}
\end{lemma}
The proof of Lemma~\ref{lem:discount1} is primarily a consequence of Proposition~\ref{prop:convergence} and can be found in Appendix~\ref{app:innerbound}. Observe that since Proposition~\ref{prop:convergence} holds for any initial state distribution, Lemma~\ref{lem:discount1} holds irrespective of the state distribution at the beginning of an epoch, and hence, is independent of the algorithm.
\begin{remark}
Lemma~\ref{lem:discount1} contains a discount factor dependent definition for $\textstyle\uj{j}{\taukj}$ under discount-averaged reward feedback. To be precise, the definition of $\textstyle\uj{j}{\taukj}$ depends on if $\gamma = \lambda_j$. The definition of $\textstyle\uj{j}{\taukj}$ provided for the case that $\gamma = \lambda_j$ holds even when $\gamma \neq \lambda_j$. However, the bound specified for when $\gamma \neq \lambda_j$ is tighter than that specified for when $\gamma = \lambda_j$. More generally, each bound we give in this paper for discount-averaged reward feedback contains similar discount factor dependent definitions; it will always be the case that the bounds provided for the event in which $\gamma = \lambda_j$ for some $j \in [m]$ hold when $\gamma \neq \lambda_j$ for each $j \in [m]$, but the latter bounds are stronger.
\end{remark}

Returning to the regret decomposition, we apply Lemma~\ref{lem:discount1}
to~\eqref{eq:prelemma1} and obtain
\begin{equation}
\textstyle\mb{E}_{\alpha}[\sum\limits_{k=1}^{n}\sum\limits_{j \in [m]}I\{\alpha(k)=j\}|\mb{E}[\mu_j-\br{k}{j}|\ft{j}{T_j^{\alpha}(k)-1}
]|]\leq \mb{E}_\alpha[\sum\limits_{k=1}^{n}\sum\limits_{j \in [m]}I\{\alpha(k)=j\} \uj{j}{\taukj}].
\label{eq:precases}
\end{equation}
We now derive a bound on~\eqref{eq:precases} dependent on the type of reward feedback. Building on Lemma~\ref{lem:discount1}, we need to consider several cases: 1) $\gamma\in (0,1)$ and $\gamma\neq \lambda_j$ for all $j\in[m]$, 2) $\gamma\in (0,1)$ and $\gamma=\lambda_j$ for some $j\in [m]$, and 3) $\gamma=1$.
\newline

\noindent
\textbf{Case 1.} $\gamma\in(0,1)$ and $\gamma \neq \lambda_j \ \forall \ j \in [m]$.
\begin{align}
    \textstyle\mb{E}_\alpha\big[\sum_{k=1}^{n}\sum_{j \in [m]}I\{\alpha(k)=j\}
\uj{j}{\taukj}\big] &= \textstyle\mb{E}_\alpha\big[\sum_{k=1}^{n}\sum_{j \in
[m]}I\{\alpha(k)=j\}\frac{C_j
(\phi_j)^{\taukj-1}(1-(\psi_j)^{\taukj})}{\bsgr_k(1-\psi_j)} \big] \notag \\
&\textstyle\leq \mb{E}_\alpha\big[\sum_{j \in
[m]}\frac{C_j}{1-\psi_j}\sum_{k=1}^{n}I\{\alpha(k)=j\}\frac{(\phi_j)^{\taukj-1}}{\bsgr_k}
\big] \notag \\
&\textstyle\leq \mb{E}_\alpha\big[\sum_{j \in[m]}\frac{C_j}{1-\psi_j}\sum_{k=1}^{n}I\{\alpha(k)=j\}(\phi_j)^{\taukj-1} \big] \notag \\
&\textstyle= \mb{E}_\alpha\big[\sum_{j \in
[m]}\frac{C_j}{1-\psi_j}\sum_{i=1}^{T_j^{\alpha}(n)}(\phi_j)^{\tauz+\zeta
(i-1) -1} \big] \notag \\
&\textstyle\leq \sum_{j \in [m]}\frac{C_j}{1-\psi_j}\sum_{i=1}^{n}(\phi_j)^{\tauz + \zeta (i-1) - 1} \label{eq:c1sum}\\
&\textstyle= \sum_{j \in [m]}C_j\big(\frac{(\phi_j)^{\tauz}}{\phi_j-\eta_j}\big)\big(\frac{1 - (\phi_j)^{\zeta n}}{1- (\phi_j)^{\zeta}}\big) \notag
\end{align}
Observe that as $n \rightarrow \infty$, the inner sum found in~\eqref{eq:c1sum} approaches the constant given as
$(\phi_j)^{\tauz}(\phi_j-\eta_j)^{-1}(1- (\phi_j)^{\zeta})^{-1}$ since it is a geometric series.
\newline

\noindent
\textbf{Case 2.} $\gamma\in(0,1)$ and $\gamma = \lambda_j \ \text{for some} \ j \in [m]$. 
\begin{align}
\textstyle\mb{E}_\alpha\big[\sum_{k=1}^{n}\sum_{j \in [m]}I\{\alpha(k)=j\}&
\uj{j}{\taukj}\big] \notag\\
&\textstyle= \mb{E}_\alpha\big[\sum_{k=1}^{n}\sum_{j \in
[m]}I\{\alpha(k)=j\}\frac{C_j(\phi_j)^{\taukj-1}\taukj}{\bsgr_k} \big] \notag \\
&\textstyle\leq \mb{E}_\alpha\big[\sum_{j \in
[m]}C_j\sum_{k=1}^{n}I\{\alpha(k)=j\}(\phi_j)^{\taukj-1}\taukj \big] \notag \\
&\textstyle= \mb{E}_\alpha\big[\sum_{j \in
[m]}C_j\sum_{i=1}^{T_j^{\alpha}(n)}(\phi_j)^{\tauz + \zeta(i-1)-1}(\tauz +
\zeta (i-1)) \big] \notag \\
&\textstyle\leq \sum_{j \in [m]}C_j\sum_{i=1}^{n}(\phi_j)^{\tauz + \zeta(i-1)-1}(\tauz + \zeta(i-1))\label{eq:arithgeo} \\
&\textstyle= \sum_{j \in [m]}C_j (\phi_j)^{\tauz - 1}\big(\frac{\tauz - (\phi_j)^{\zeta n}(\tauz + \zeta n)}{1 -(\phi_j)^{\zeta}} + \frac{\zeta (\phi_j)^{\zeta}(1 - (\phi_j)^{\zeta n})}{(1 - (\phi_j)^{\zeta})^2}\big) \notag
\end{align}
The final equality follows from recognizing that the inner sum contained in~\eqref{eq:arithgeo} is an arithmetico-geometric series and substituting the expression for the finite sum. Observe that as $n\rightarrow \infty$, the arithmetico-geometric series in~\eqref{eq:arithgeo} approaches the constant given as $\tauz(1 -(\phi_j)^{\zeta})^{-1} + \zeta (\phi_j)^{\zeta}((1 - (\phi_j)^{\zeta})^2)^{-1}$. 
For more details on this series, see Appendix~\ref{app:lem:boundforAH}. 
\newline

\noindent
\textbf{Case 3:} $\gamma = 1$.
\begin{align}
    \textstyle\mb{E}_\alpha\big[\sum_{k=1}^{n}\sum_{j \in [m]}I\{\alpha(k)=j\}
\uj{j}{\taukj}\big] &=\textstyle \mb{E}_\alpha\big[\sum_{k=1}^{n}\sum_{j \in
[m]}I\{\alpha(k)=j\}\frac{C_j(1-(\lambda_j)^{\taukj})}{\taukj(1-\lambda_j)}\big] \notag \\
&\textstyle\leq \mb{E}_\alpha\big[\sum_{j \in
[m]}\frac{C_j}{1-\lambda_j}\sum_{k=1}^{n}I\{\alpha(k)=j\}\frac{1}{\taukj}\big] \notag \\
&=\textstyle \mb{E}_\alpha\big[\sum_{j \in
[m]}\frac{C_j}{1-\lambda_j}\sum_{i=1}^{T_j^{\alpha}(n)}\frac{1}{\tauz + \zeta
(i-1)}\big] \notag \\
&\textstyle\leq \sum_{j \in [m]}\frac{C_j}{1-\lambda_j}\sum_{i=1}^{n}\frac{1}{\tauz + \zeta (i-1)} \label{eq:preharmonic} \\
&\textstyle\leq \sum_{j \in [m]}\frac{C_j}{1-\lambda_j}\big(\frac{1}{\tauz} + \frac{1}{\zeta}\log\big(1+\frac{\zeta n}{\tauz}\big)\big) \notag
\end{align}
The final inequality is obtained from the observation that the inner sum found in~\eqref{eq:preharmonic} is a harmonic sum that can be bound with standard techniques. We include the derivation in Appendix~\ref{app:harmonic}. 

The bounds we just derived give our final bounds on the Markovian regret penalty. Hence, plugging the bounds on the Markovian regret penalty back into the initial expression for the regret decomposition found in~\eqref{eq:regret1} gives rise to the following proposition.  
\begin{proposition}[Regret Decomposition] Suppose Assumptions~\ref{assumption:ergodic} and~\ref{assumption:reverse} hold. Then, for any given algorithm $\alpha$ with corresponding epoch length sequence $\{\taukr\}_{k=1}^n$ as given in~\eqref{eq:tauk}:
\begin{equation*}
\textstyle R^\alpha(n) \leq\textstyle  \sum_{j\neq j_\ast }\mb{E}_\alpha\big[
T_j^\alpha(n)\big]\Delta_j+ \sum_{j\in [m]} \lj{j}{n},
\end{equation*}
where $\lj{j}{n}$ is defined as follows depending on the type of reward feedback:
\begin{enumerate}[topsep=0pt,itemsep=-2pt]
\item Discount-Averaged Reward Feedback: $\gamma\in(0,1)$.
\begin{equation}
\lj{j}{n} = 
\begin{cases}
\textstyle C_j\big(\frac{(\phi_j)^{\tauz}}{\phi_j-\eta_j}\big)\big(\frac{1 - (\phi_j)^{\zeta n}}{1-(\phi_j)^\zeta}\big), & \text{if} \ \gamma \neq \lambda_j \  \forall \ j \in [m]\\
\textstyle C_j(\phi_j)^{\tauz - 1}\big(\frac{\tauz - (\phi_j)^{\zeta n}(\tauz + \zeta n)}{1 -(\phi_j)^{\zeta}} + \frac{\zeta (\phi_j)^{\zeta}(1 - (\phi_j)^{\zeta n})}{(1 - (\phi_j)^{\zeta})^2}\big), & \text{otherwise} 
\end{cases}.
\label{eq:lj1}
\end{equation}
\item Time-Averaged Reward Feedback: $\gamma=1$.
\begin{equation}
\textstyle \lj{j}{n}=\frac{C_j}{1-\lambda_j}\big(\frac{1}{\tauz}+\frac{1}{\zeta}\log\big(1+\frac{\zeta n}{\tauz} \big)\big).
\label{eq:lj12}
\end{equation}
\end{enumerate}
\label{thm:regretdecomp1}
\end{proposition}

\begin{remark}\label{remark:lj_size}
The type of reward feedback (discount-averaged or time-averaged) for which the Markovian regret penalty of $\sum_{j\in[m]}\lj{j}{n}$ is not as costly depends on the precise discount factor under discount-averaged reward feedback, the Markov chain statistics $(C_j, \lambda_j)$ for each $j\in[m]$, and the time horizon $n$. Typically however, 
the Markovian regret penalty will be smaller under discount-averaged reward feedback than under time-averaged reward feedback. In most cases, this is to be expected since the rewards are given increased weight as the state distribution tends closer to a stationary distribution.
\end{remark}

\subsubsection{Discussion of Regret Decomposition}\label{sec:regret_decomp_discussion}
The dynamic and evolving reward structure present in the problem we study leads any algorithm to incur regret beyond the usual penalty for playing suboptimal arms via, what we refer to as, the \emph{Markovian regret penalty} (see~\ref{eq:extraregret}). Leveraging classic results on mixing of Markov chains and the construction of the epoch length sequence $\{\taukr\}_{k=1}^n$ considered in this work, we bounded the Markovian regret penalty with $\sum_{j\in [m]}\lj{j}{n}$. In essence, this bound limits the regret arising from the rewards on an arm being drawn from an evolving distribution to a term that quickly approaches a constant as the time horizon grows in the case of discount-averaged reward feedback and a term that grows only logarithmically in the time horizon in the case of time-averaged reward feedback. The regret decomposition allows us to now focus soley on the selection of suboptimal arms.


%% file: Main_Sections/algorithm_preliminaries.tex
Given Proposition~\ref{thm:regretdecomp1}, in order to obtain a bound on the regret for a particular algorithm $\alpha$, we need to limit $\mb{E}_\alpha[T^{\alpha}_j(n)]$ for each $j \neq j_\ast$. To do so, it is important to characterize the uncertainty in the empirical mean reward of each arm as a function of the number of times the arm has been pulled. Fundamentally, there are two sources of uncertainty in the observed rewards: 
\begin{enumerate}
\item The reward distribution on each arm is dynamic owing to the dependence on the unobserved and evolving state distribution.
\item The observed rewards derive from stochastic reward
distributions.
\end{enumerate}
Hence, in contrast to the conventional stochastic multi-armed bandit problem,
where the stochasticity of the observed rewards is the only source of uncertainty, we must also carefully consider how much uncertainty arises from the dynamic nature of the reward distributions as an artifact of the unobserved and evolving state distribution. 

From Lemma~\ref{lem:discount1}, we can observe that the upper bound on the deviation between the expected reward of an arm selection and the expected stationary distribution reward for that arm decays as a function of the number of times the arm has been selected---since epochs grow linearly in the number of times an arm has been pulled in the past. Consequently, the mean of these deviations vanishes as the number of times the arm has been pulled grows. Using this observation, the following lemma delineates the maximum amount of uncertainty in the empirical mean reward of an arm arising from the dynamic nature of the reward distribution on the arm from that coming out of the stochasticity of the rewards. Precisely, Lemma~\ref{lem:boundforAH} provides a bound on the deviation between the expected mean reward and the expected stationary distribution reward for an arm $j \in [m]$ after it has been selected $T_j$ times. 
\begin{lemma}[Convergence of Expected Mean Reward to Expected Stationary Reward]
Suppose Assumptions~\ref{assumption:ergodic} and~\ref{assumption:reverse} hold. Then, after an arm $j \in[m]$ has been played $T_j$ times by an algorithm $\alpha$ with corresponding epoch length sequence $\{\taukr\}_{k=1}^n$ as given in~\eqref{eq:tauk}, 
\setlength{\belowdisplayskip}{0pt}\setlength{\belowdisplayshortskip}{0pt}
\begin{equation*}
\textstyle
\big|\mu_j-\frac{1}{T_j}\sum_{i=1}^{T_j}\mb{E}[\car{i}{j}|\ft{j}{i-1}]\big|\leq \frac{\lj{j}{T_j}}{T_j}.
\end{equation*}
\label{lem:boundforAH}
\end{lemma} 
The proof of Lemma~\ref{lem:boundforAH} follows from manipulating the expression that needs to be bounded into a sum over terms that can each be bounded using Lemma~\ref{lem:discount1} and then applying similar analysis to that which was used to bound~\eqref{eq:precases} when deriving Proposition~\ref{thm:regretdecomp1}. The full proof can be found in Appendix~\ref{app:lem:boundforAH}.

\begin{remark}
In a similar manner to how we were able to limit the Markovian regret penalty, 
Lemma~\ref{lem:boundforAH} limits the amount of uncertainty stemming from the dynamic nature of the reward distribution on an arm to a term that tends toward zero quickly as a function of the number of times the arm has been pulled. 
\end{remark}

Given that Lemma~\ref{lem:boundforAH} characterizes the maximum amount of uncertainty coming solely from the evolution of the reward distributions in time,  we are left to identify the uncertainty arising from the stochasticity in the rewards. To do so, we need a concentration inequality that does not require independence in the observed rewards of an arm since the underlying Markov chain that generates the rewards is common across the arms. On that account, an important technical tool for our impending algorithm-based regret analysis is the Azuma-Hoeffding inequality.
\begin{proposition}[Azuma-Hoeffding Inequality~\citep{azuma:1967aa,hoeffding:1963aa}]
Suppose $(Z_i)_{i\in \mb{Z}_+}$ is a martingale with respect to the filtration $(\mc{F}_i)_{i\in \mb{Z}_+}$ and there are finite, non-negative constants $c_i$, such that $|Z_i-Z_{i-1}|<c_i$ almost surely for all $i \geq 1$. Then for all $\epsilon > 0$ 
\setlength{\belowdisplayskip}{0pt}\setlength{\belowdisplayshortskip}{0pt}
\begin{equation*}
\textstyle P(Z_n-\mb{E}[Z_n]\leq -\epsilon)\leq \exp\Big(-\frac{\epsilon^2}{2\sum_{i=1}^n c_i^2} \Big).
\end{equation*}
\label{prop:AH}
\end{proposition}
To apply the Azuma-Hoeffding inequality, we need to formulate our problem as a
Martingale difference sequence. 
Toward this end, define the random variables
\begin{equation*}
\textstyle X_{j,i}=\car{i}{j}-\mb{E}[\car{i}{j}|\ft{j}{i-1}],
\label{eq:xk}
\end{equation*} 
where the expectation is taken with respect to $\mc{T}_r(\theta, j)$, and
\begin{equation}
\textstyle Y_{j,T_j}=\sum_{i=1}^{T_j}X_{j,i},
\label{eq:yk}
\end{equation}
where $T_j$ denotes number of times arm $j$ has been played. Note that  $Y_{j,T_j}$ is a martingale; indeed, since $Y_{j,T_j}$ is $\ft{j}{T_j}$--measurable by construction,
\begin{equation*}
\mb{E}[Y_{j,{T_j+1}}|\ft{j}{T_j}]=\mb{E}[X_{j,{T_j+1}}|\ft{j}{T_j}]+\mb{E}[Y_{j,T_j}|\ft{j}{T_j}]=Y_{j,T_j}
\end{equation*}
and $\mb{E}[|Y_{j,T_j}|]<\infty$ since rewards are bounded. Moreover, the
boundedness of the rewards also implies the martingale $Y_{j,T_j}$ has bounded
differences: $|Y_{j,T_j}-Y_{j,{T_j-1}}|=|X_{j,T_j}|\leq 1$ almost surely since rewards are normalized to be on the interval $[0,1]$, without loss of generality. 

The remainder of this section is devoted to presenting our proposed EpochUCB and EpochGreedy algorithms along with the regret bound guarantees we obtain for each of these algorithms. The environment simulation procedure for the algorithms is given in Algorithm~\ref{alg:armpull}. To derive the algorithm-based regret bounds, we make use of the techniques we have developed to reason about the uncertainty in the empirical mean reward of each arm in conjunction with the proof techniques developed to analyze the UCB and $\vep$--greedy algorithms.

\input{Algorithms/pullarm}

%% file: Algorithms/pullarm.tex
\begin{algorithm}[htbp]
\caption{Environment Implementation for Pulling an Arm}
\label{alg:armpull}
\begin{algorithmic}[1] 
\Function{pullarm}{$i$, $k$, $\gamma$, $t_k$, $\taukr$} 
\State $\br{k}{i} \gets 0$, $\bsgr_k \gets 0$
\For{$t \in [t_k,t_k+\taukr)$} \Comment{Pull arm $i$}
    \State $\nr{t}{i}\sim \mc{T}_r(\theta_t, i)$, $\thetarv{t}\sim\beta_{t}$ 
    \State $\br{k}{i}\gets \br{k}{i} + (\gamma)^{t_k + \taukr-1-t}\nr{t}{i}$
    \State $\bsgr_k \gets \bsgr_k + (\gamma)^{t_k + \taukr -1-t}$
    \State $\textstyle \beta_{t+1}(\theta)=\sum_{\theta'\in \Theta} P_{i}(\theta',\theta)\beta_{t}(\theta') \ \forall \ \theta \in \Theta$
\EndFor
\State $\br{k}{i}\gets \br{k}{i}/\bsgr_k$
\State\Return $\br{k}{i}$
\EndFunction
\end{algorithmic}
\end{algorithm}

%% file: Main_Sections/ucb.tex
In this section, we analyze the regret of EpochUCB (Algorithm~\ref{alg:ucb}). At a high level, EpochUCB plays the arm that maximizes the sum of the empirical mean reward and the confidence window at each epoch for a time period that grows linearly as a function of the number of times the arm selection has been chosen in the past. More formally, for each arm $j \in [m]$, define the empirical mean reward  after $k-1$ epochs to be
\begin{equation*}
\textstyle\barR{T_j^{\alpha}(k-1)}{j}=\frac{1}{T_j^{\alpha}(k-1)}\sum_{i=1}^{T_j^{\alpha}(k-1)}\car{i}{j},
\end{equation*}
and the confidence window at epoch $k \in [n]$ to be
\begin{equation}
\textstyle\cf{k}{T_j^{\alpha}(k-1)}{j}=\frac{\lj{j}{T_j^{\alpha}(k-1)}}{T_j^{\alpha}(k-1)}+\sqrt{\frac{6\log(k)}{T_j^{\alpha}(k-1)}}.
\label{eq:confidence_window}
\end{equation}
Following an initialization round in which each arm is played once, the algorithm selects the arm $\alpha(k)$ at epoch $k\in [n]$ such that: \begin{equation}
\textstyle \alpha(k) = \arg \max_{j\in [m]} \barR{T_j^{\alpha}(k-1)}{j} + \cf{k}{T_j^{\alpha}(k-1)}{j}.
\label{eq:epochucb_policy}
\end{equation}
\input{Algorithms/ucbalg}

Our algorithm bears conceptual similarity to the UCB2 algorithm introduced in~\citet{auer:2002aa}. However, the crucial ingredients in our method are the careful choice of the confidence window that captures multiple sources of uncertainty, and the way we exploit the linearly increasing epoch length sequence to ensure that the window of uncertainty is in fact diminishing after multiple plays of an arm. The following theorem provides an upper bound on the number of times any suboptimal arm will be played by the EpochUCB algorithm.

\begin{theorem}[Bound on Suboptimal Plays for EpochUCB]
Suppose Assumptions~\ref{assumption:ergodic} and~\ref{assumption:reverse} hold. Let $\alpha$ be the EpochUCB algorithm with corresponding epoch length sequence
$\{\taukr\}_{k=1}^n$ as given in~\eqref{eq:tauk}. Then, for each suboptimal arm $j \in [m]$,
\begin{equation*}
\textstyle\mb{E}_\alpha[T_j^{\alpha}(n)]\leq \frac{4}{\Delta_j^2}\big(\rj{j}+\sqrt{6\log(n)} \big)^2+3+2\log(n),
\end{equation*}
where $\rj{j}$ is a time-invariant constant defined as follows depending on the type of reward feedback:
\begin{enumerate}[topsep=0pt,itemsep=-2pt]
\item Discount-Averaged Reward Feedback: $\gamma \in (0, 1)$.
\begin{equation}
\rj{j} = 
\begin{cases}
\textstyle C_j\big(\frac{(\phi_j)^{\tauz}}{\phi_j - \eta_j}\big)\big(\frac{1}{1-(\phi_j)^{\zeta}}\big), & \text{if} \ \gamma \neq \lambda_j \ \forall \ j \in [m] \\
\textstyle C_j(\phi_j)^{\tauz-1}\big(\frac{\tauz }{1 - (\phi_j)^{\zeta}}+\frac{\zeta(\phi_j)^{\zeta}}{(1 - (\phi_j)^{\zeta})^2}\big), & \text{otherwise}
\end{cases}.
\label{eq:rhoj}
\end{equation}
\item Time-Averaged Reward Feedback: $\gamma=1$.
\begin{equation}
\textstyle \rj{j}=\frac{C_j}{\sqrt{\zeta \tauz}(1-\lambda_j)}\big(1 + \frac{\zeta}{\tauz} \big).
\label{eq:rhoj-avg}
\end{equation}
\end{enumerate}
\label{thm:regretbound}
\end{theorem}
\noindent
The complete proof can be found in Appendix~\ref{app:thm:ucb}.

\begin{proof}(\emph{sketch.}) 
The intuition behind the proof is that the algorithm can play a suboptimal arm when: (\emph{i}) the confidence bounds on the stationary distribution rewards for the optimal arm or a suboptimal arm fail, or (\emph{ii}) the optimal arm and a suboptimal arm have been sampled insufficiently to distinguish between the respective stationary distribution rewards. 
The crux of the proof is the derivation of the confidence window, which must capture the maximum amount of uncertainty in the observed rewards. Since Lemma~\ref{lem:boundforAH} provides an upper bound on the uncertainty from the dynamic nature of the reward distributions, we are only left to characterize the uncertainty from the stochasticity in the rewards.   

We previously formulated the arm-based reward observations as a Martingale difference sequence so that we can apply the Azuma-Hoeffing inequality (Proposition~\ref{prop:AH}) to bound the uncertainty arising from the stochasticity in the rewards. Toward this end, suppose an arm $j \in [m]$ has been played $T_j$ times prior to an epoch $k \in [n]$. We need to relate the martingale $Y_{j,T_j}$ as defined in~\eqref{eq:yk} to the deviation between the empirical mean reward and the expected stationary distribution reward for the arm, so that we can derive a bound on this quantity of interest and obtain our confidence window. 

To do so, define the event $\omega = \{\mu_j - \barR{T_j}{j} \geq \epsilon\}$ for some arbitrary $\epsilon > 0$. We can equivalently express this event as
\begin{align*}
\omega&=\textstyle\big\{\mu_j-\frac{1}{T_j}\sum_{i=1}^{T_j}
\mb{E}[\car{i}{j}|\ft{j}{i-1}]+\frac{1}{T_j}\sum_{i=1}^{T_j}
\mb{E}[\car{i}{j}|\ft{j}{i-1}]-\barR{T_j}{j}\geq \epsilon \big\}\\
&=\textstyle\big\{\mu_j- \frac{1}{T_j}\sum_{i=1}^{T_j}
\mb{E}[\car{i}{j}|\ft{j}{i-1}]-\frac{Y_{j,T_j}}{T_j}\geq \epsilon\big\}.
\end{align*}
This representation is obtained from adding and subtracting $\frac{1}{T_j}\sum_{i=1}^{T_j}\mb{E}[\car{i}{j}|\ft{j}{i-1}]$ into the event $\omega$ and invoking the definition of $Y_{j,T_j}$ found in~\eqref{eq:yk}. From Lemma~\ref{lem:boundforAH}, we obtain
\begin{equation*}
\textstyle \omega\subset\big\{\frac{\lj{j}{T_j}}{T_j}-\frac{Y_{j,T_j}}{T_j}\geq \epsilon \big\}=\big\{\frac{Y_{j,T_j}}{T_j}\leq \frac{\lj{j}{T_j}}{T_j}-\epsilon \big\}.
\end{equation*}
Hence, applying Proposition~\ref{prop:AH},
\begin{equation*}
\textstyle P(\mu_j- \barR{T_j}{j}\geq \epsilon)\leq
P\big(\frac{Y_{j,T_j}}{T_j}\leq \frac{\lj{j}{T_j}}{T_j}-\epsilon\big)\leq 
\exp\big(-\frac{T_j}{2}\big(\epsilon-\frac{\lj{j}{T_j}}{T_j} \big)^2\big).
\end{equation*}
Defining $\epsilon=\sqrt{\frac{2}{T_j}\log\big(\frac{1}{\delta}\big)}+\frac{\lj{j}{T_j}}{T_j}$, we determine that for any fixed $\delta>0$,
\begin{equation*}
\textstyle P\big(\mu_j-\barR{T_j}{j}\geq
\sqrt{\frac{2}{T_j}\log\big(\frac{1}{\delta}\big)}+\frac{\lj{j}{T_j}}{T_j}\big)\leq \delta.
\end{equation*}
Selecting $\delta(k)=k^{-3}$, we recover the definition of the confidence window found in~\eqref{eq:confidence_window} and get the ensuing confidence bounds on the expected stationary distribution reward of an arm $j \in [m]$ that hold with growing probability as the algorithm proceeds:
\begin{equation*}
P(\mu_j \leq \barR{T_j}{j} + \cf{k}{T_j}{j})\geq 1 - k^{-3} \quad \text{and} \quad P(\mu_j \geq \barR{T_j}{j} - \cf{k}{T_j}{j})\geq 1 - k^{-3}.
\end{equation*}
Since the probability of the confidence bounds failing diminishes as the algorithm progresses, we can show that the number of times a suboptimal arm is played as an artifact of the confidence bounds failing grows only logarithmically in the time horizon. The details of this argument are found in our full proof.

When the confidence bounds hold, a suboptimal arm can be only played when the optimal arm and a suboptimal arm have been sampled insufficiently to distinguish between the respective stationary distribution rewards.
To distinguish between the stationary distribution rewards of the optimal arm $j_{\ast}$ and that of a suboptimal arm $j$, it is adequate to find the smallest integer $\ell$ representing the number of samples of the arm, such that 
\begin{equation*}
\textstyle \mu_\ast - \mu_j - 2\cf{k}{\ell}{j} = \Delta_j - 2\big(\frac{\lj{j}{\ell}}{\ell}+\sqrt{\frac{6\log(k)}{\ell}} \big) > 0 
\end{equation*}
holds for every epoch $k \in [n]$. Indeed, we do so for each function $\lj{j}{\ell}$ can adopt dependent on the type of reward feedback, and show that the number of times each suboptimal arm must be sampled to identify the optimal arm grows only logarithmically in the time horizon. This component of the proof is is detailed in our full proof.
\end{proof}

The following corollary is a direct consequence of
Proposition~\ref{thm:regretdecomp1} and Theorem~\ref{thm:regretbound}.
\begin{corollary}[Gap-Dependent Regret Bound for EpochUCB]
Under the assumptions of Theorem~\ref{thm:regretbound},
\setlength{\belowdisplayskip}{0pt}\setlength{\belowdisplayshortskip}{0pt}
\begin{equation*}
\textstyle R^{\mathrm{EpochUCB}}(n)  \leq \sum_{j\neq j_\ast}\big(
\frac{4}{\Delta_j}\big(\rj{j}+\sqrt{6\log(n)}\big)^2 +3\Delta_j
+2\log(n)\Delta_j\big) +\sum_{j\in [m]}\lj{j}{n}.
\end{equation*}
\label{corr:ucb}
\end{corollary}
This gap-dependent regret bound is $\mc{O}(\log(n))$. Thus, although our problem is more general than many other bandit problems considered previously, we still obtain an gap-dependent regret bound of the same asymptotic gap-dependent order and inverse dependence on the gaps. 

The following corollary gives a gap-independent regret
bound for EpochUCB of order $\mc{O}(\sqrt{n\log(n)})$.
\begin{corollary}[Gap-Independent Regret Bound for EpochUCB]
Under the assumptions of Theorem~\ref{thm:regretbound},
\setlength{\belowdisplayskip}{0pt}\setlength{\belowdisplayshortskip}{0pt}
\begin{equation*}
\textstyle R^{\mathrm{EpochUCB}}(n) \leq \sqrt{n\sum_{j \in {[m]}}\big(4(\rj{j})^2+ 8\rj{j} \sqrt{6\log(n)} + 26\log(n)+2\big)} + \sum_{j\in [m]}\lj{j}{n}.
\end{equation*}
\label{corr:gap_independent}
\end{corollary}
The proof of Corollary~\ref{corr:gap_independent} follows from bounding $\sum_{j\neq j_\ast}\mb{E}_\alpha\big[T_j^\alpha(n)\big]\Delta_j$ using the Cauchy-Schwarz inequality, Theorem~\ref{thm:regretbound}, and $\Delta_j$ being bounded in $[0, 1]$. We defer the full proof to Appendix~\ref{app:corr:gap_independent}.

\subsubsection{Discussion of Regret Bounds}\label{sec:complexity}
The regret bounds we present have an intuitive and necessary dependence on the Markov chain statistics ($C_j, \lambda_j$) for each arm $j \in [m]$. Such a dependence is commonly found in regret bounds for  bandit problems with Markov chains. In fact, we are unaware of any papers in the related rested and restless multi-armed bandit literature~\citep[e.g., see the work of][and the references therein]{tekin:2012aa} that do not have a similar dependence. Indeed, the statistics ($C_j, \lambda_j$) are directly tied to the mixing times of the Markov chains characterized by the transition matrix for each arm $j \in [m]$, and the regret bounds that can be obtained necessarily depend on the mixing times. 

Recall from our regret decomposition that there is a penalty for selecting suboptimal arms and for the state distribution deviating from a stationary distribution. Clearly, the regret deriving from the state distribution converging to the stationary distribution of an arm must scale proportionally to the mixing time of that Markov chain. Similarly, when arms converge toward their respective stationary distributions promptly, the observed reward feedback can be exploited early on to confidently identify the expected stationary distribution reward of each arm. Conversely, as the mixing times of the Markov chains grow, any learning algorithm would require many samples to confidently estimate the expected stationary distribution reward of each arm since the feedback observed early on cannot be ensured to closely approximate rewards drawn from the stationary distributions of the arms. Hence, the dependence on the Markov chain statistics in our regret bounds should be seen as capturing a natural measure of instance-dependent complexity stemming from the Markov chains that complements the standard measure of instance-dependent complexity characterized by the reward gap. 

Finally, we remark that there exists large classes of transition matrices $P$ for which the second largest eigenvalue of the multiplicative reversiblization $\lambda_2(M(P))$ is bounded away from 1. Moreover, a significant body of work that has identified sufficient conditions for such instances~\citep[e.g., see Corollary 2.2 in][]{kirkland2009subdominant}. In practice, we also find that $\lambda_2(M(P))$ is bounded away from 1. For example, if we sample a transition matrix $P$ with 10 states from uniform and standard normal distributions (subject to normalization) 1000 times each, we find $\lambda_2(M(P))$ has respective means of 0.32 and 0.7 with 95th percentiles of 0.38 and 0.83. Thus, for the vast majority of problem instances the size of the constant factors in our regret bounds tied to the Markov chain statistics will be reasonable.

%% file: Algorithms/ucbalg.tex
\begin{algorithm}[htbp]
\caption{EpochUCB}
\label{alg:ucb}
\begin{algorithmic}[1] 
\Procedure{EpochUCB}{$\tauz$, $\zeta$, $\gamma$} 
\State $t_1\gets 0$, $T_j\gets 1 \ {\small \&} \ \barR{T_j}{j} \gets 0$ $\ \forall \ j\in [m]$
    \For{$1\leq k\leq m$} \Comment{Pull each arm once}\\
        $\qquad$$\barR{T_j}{j}\gets$ \Call{pullarm}{$k$, $k$, $\gamma$, $t_k$, $\tauz$} \Comment{Algorithm~\ref{alg:armpull}}\\
        $\qquad$$t_{k+1}\gets t_k+\tauz$ \Comment{Equation~\ref{eq:tauk}}
    \EndFor
    \While{$k>m$} \Comment{EpochUCB}\\
    $\qquad$$i=\arg\max_{j\in[m]} \barR{T_j}{j}+\cf{k}{T_j}{j}$
    \Comment{Equation~\ref{eq:epochucb_policy}}\\
    $\qquad\taukr \gets \tauz + \zeta T_i$ \Comment{Equation~\ref{eq:tauk}}\\
    $\qquad$$\br{k}{i}\gets$ \Call{pullarm}{$i$, $k$, $\gamma$,
    $t_k$, $\taukr$} \Comment{Algorithm~\ref{alg:armpull}}\\
    $\qquad$$\barR{T_i + 1}{i} \gets \barR{T_i}{i} + (\br{k}{i}-\barR{T_i}{i})/T_i$
    \\
    $\qquad$$T_i\gets T_i+1$, $t_{k+1}\gets t_k+\taukr$, $k\gets
    k+1$
    \EndWhile\label{euclidendwhile}
        \EndProcedure
    \end{algorithmic}
\end{algorithm}

%% file: Main_Sections/greedy.tex
In this section, we analyze the regret of EpochGreedy
(Algorithm~\ref{alg:greedy}). The EpochGreedy algorithm is a simple bandit policy that plays the arm with maximum empirical mean reward with probability $1-\vep_k$ and an arm selected uniformly at random with probability $\vep_k$ at an epoch $k \in [n]$. Formally, the policy at epoch $k \in [n]$ with $\alpha$ taken as the EpochGreedy algorithm is 
\begin{equation}
\alpha(k) = 
\begin{cases}
\arg \max_{j\in[m]} \barR{T_j^{\alpha}(k-1)}{j}, & \text{w.p.}~1-\vep_k \\
j, & \text{w.p.}~\frac{\vep_k}{m} \ \forall \ j \in [m] \\
\end{cases}.
\label{eq:greedy_policy}
\end{equation}

\input{Algorithms/greedyalg}

EpochGreedy is a variant of the $\vep$--greedy algorithm presented in~\citet{auer:2002aa}. The novelty in our algorithm and analysis is the construction of the random exploration probability sequence $\{\vep_k\}_{k=1}^n$ to suit our problem and obtain strong regret guarantees. The challenge in devising this sequence and analyzing the resulting algorithm stems from the need to bear in mind the multiple sources of uncertainty present in our problem. 
Consequently, critical tools in our analysis include our technique to quantify the convergence of the expected mean reward of each arm to an expected stationary distribution reward (Lemma~\ref{lem:boundforAH}) and our formulation of the observed reward feedback on each arm as a Martingale difference sequence to conform with the Azuma-Hoeffding inequality (Proposition~\ref{prop:AH}).

The random exploration probability sequence $\{\vep_k\}_{k=1}^n$ in our algorithm decays linearly as a function of the time horizon---just as is the case in the $\vep$--greedy algorithm of~\citet{auer:2002aa}. However, a fundamental distinction of the random exploration probability sequence we construct is the dependence of the constant factors found in the sequence on the Markov chain statistics ($C_j, \lambda_j$) for each arm $j \in [m]$. This dependence shows up since the Markov chain statistics act as a measure of instance-dependent complexity that complements the standard measure of instance-dependent complexity characterized by the reward gap. For a detailed discussion on this point, refer back to Section~\ref{sec:complexity}.

Toward formally posing the random exploration probability sequence we construct, fix constants $c$ and $d$ such that $0 \leq d \leq \Delta_{\min}$ where $\Delta_{\text{min}}=\min_{j\neq j_\ast}\Delta_j$ and $c \geq c'\nu^2$ where $c' > 8$, $\nu\geq \max\{\kappa, \frac{d}{\sqrt{c'}}\}$, and $\kappa = \min\{\kappa > 0: \kappa\sqrt{i} \geq \lj{j}{i} \ \forall \ i \in [n], j \in [m]\}$. Recall that $\lj{j}{\cdot}$ depends on the Markov chain statistics ($C_j, \lambda_j$) and is defined in~\eqref{eq:lj1} for discount-averaged reward feedback and in~\eqref{eq:lj12} for time-averaged reward feedback. Finally, define $\{\vep_k\}_{k=1}^n$ to be a sequence with $\vep_k=\min\{1, \frac{cm}{d^2k}\}$ for each epoch $k \in [n]$ so that for $k \geq \lceil\frac{cm}{d^2}\rceil$, $\vep_k = \frac{cm}{d^2k}$. The following theorem provides an upper bound on the probability that any suboptimal arm $j \in [m]$ will be played by the EpochGreedy algorithm at an epoch $k \geq \lceil\frac{cm}{d^2}\rceil$.

\begin{theorem}[Bound on Probability of EpochGreedy Playing a Suboptimal Arm]  Suppose Assumptions~\ref{assumption:ergodic} and~\ref{assumption:reverse} hold. Fix constant $c$ such that $c\geq c'\nu^2$ where $c' > 8$, $\nu = \max\{\kappa, \frac{d}{\sqrt{c'}}\}$, and $\kappa = \min\{\kappa > 0: \kappa\sqrt{i} \geq \lj{j}{i} \ \forall \ i \in [n], j \in [m]\}$. Define $c'' = (4c')(\sqrt{c'/2}-2)^{-2}$. Moreover, fix constant $d$ such that $0 \leq d \leq \Delta_{\min}$ where $\Delta_{\text{min}}=\min_{j\neq j_\ast}\Delta_j$.
Let $\alpha$ be the EpochGreedy algorithm with epoch length sequence $\{\tauk\}_{k=1}^n$ as given in~\eqref{eq:tauk} and random exploration probability sequence $\{\vep_k\}_{k=1}^n$ where $\vep_k=\min\{1, \frac{cm}{d^2k}\}$ for each epoch $k \in [n]$. Then, at any epoch $k\geq \lceil\frac{cm}{d^2}\rceil$ and for each suboptimal arm $j \in [m]$,
\setlength{\belowdisplayskip}{0pt}\setlength{\belowdisplayshortskip}{0pt}
\begin{align*}
\textstyle P(\alpha(k)=j) \leq  \frac{c}{d^2k} & \textstyle+ \big(\frac{2c}{d^2}\log \big(\frac{(k-1)d^2\exp(1/2)}{cm} \big)\big)\big( \frac{cm}{(k-1)d^2\exp(1/2)}\big)^{c/(5d^2)} \\
&\textstyle +\big(\frac{2c''\exp(1)}{d^2}\big)\big( \frac{cm}{(k-1)d^2\exp(1/2)} \big)^{c/c''}.
\end{align*}
\label{thm:greedy}
\end{theorem}
\noindent
The complete proof can be found in Appendix~\ref{app:thm:greedy}.
\begin{proof}(\emph{sketch.}) 
The proof hinges on the fact that the algorithm can play a suboptimal arm as a result of random exploration or when the empirical mean reward of a suboptimal arm is greater than the empirical mean reward of the optimal arm. Indeed, the probability that a suboptimal arm $j \in [m]$ is chosen at epoch $k \geq \lceil\frac{cm}{d^2}\rceil$ is given by
\begin{equation*}
\textstyle P(\alpha(k) = j) = \frac{\vep_k}{m} + (1-\vep_k)
P(\barR{T_j^{\alpha}(k-1)}{j} \geq \barR{T_{\ast}^{\alpha}(k-1)}{\ast}), 
\end{equation*}
which we bound as
\begin{equation}
\textstyle P(\alpha(k) = j) \leq \frac{c}{d^2k} + P(\barR{T_j^{\alpha}(k-1)}{j} \geq \barR{T_{\ast}^{\alpha}(k-1)}{\ast}).
\label{eq:greedy_sketch_final}
\end{equation}
Moreover,
\begin{equation*}
\textstyle P(\barR{T_j^{\alpha}(k)}{j} \geq \barR{T_{\ast}^{\alpha}(k)}{\ast}) \leq P\big(\barR{T_j^{\alpha}(k)}{j} \geq \mu_j + \frac{\Delta_j}{2}\big) + P\big(\barR{T_{\ast}^{\alpha}(k)}{\ast} \leq \mu_\ast - \frac{\Delta_j}{2}\big).
\end{equation*}
Defining $x_0 = \frac{1}{2}(\sum_{i=1}^k\frac{\vep_i}{m})$ and using techniques found in~\citet{auer:2002aa}, we can show
\begin{align}
\textstyle P\big(\barR{T_j^{\alpha}(k)}{j} \geq \mu_j + \frac{\Delta_j}{2}\big) &\leq \textstyle \sum\limits_{i=1}^{\lfloor x_0 \rfloor} P\big(T_j^{\alpha}(k)= i \big| \barR{i}{j} \geq \mu_j + \frac{\Delta_j}{2}\big)+\sum\limits_{i=\lfloor x_0\rfloor+1}^k  P\big(\barR{i}{j} \geq \mu_j + \frac{\Delta_j}{2}\big) \notag \\
 &\leq \textstyle x_0\exp\big(\frac{-x_0}{5}\big) +\sum_{i=\lfloor x_0\rfloor+1}^k  P\big(\barR{i}{j} \geq \mu_j + \frac{\Delta_j}{2}\big).
\label{eq:greedy_sketch_to_bound}
\end{align}
We can also conclude that $x_0 \geq \frac{c}{2d^2}$ and $x_0\geq \frac{c}{d^2}\log(\frac{d^2k\exp(1/2)}{cm})$.

In this proof sketch, we focus on bounding the sum present in~\eqref{eq:greedy_sketch_to_bound} since this is where our analysis deviates from that found in~\citet{auer:2002aa} most significantly. To do so, consider the event $\omega = \{\barR{i}{j} -\mu_j \geq \frac{\Delta_j}{2}\}$, which we can express as
\begin{align*}
\textstyle \omega &=\textstyle\big\{\barR{i}{j}-\frac{1}{i}\sum_{l=1}^i\mb{E}[\car{l}{j}|\ft{j}{l-1}]+\frac{1}{i}\sum_{l=1}^i\mb{E}[\car{l}{j}|\ft{j}{l-1}] - \mu_j\geq\frac{\Delta_j}{2}\big\} \\
&=\textstyle\big\{\frac{Y_{j, i}}{i}+\frac{1}{i}\sum_{l=1}^i\mb{E}[\car{l}{j}|\ft{j}{l-1}]- \mu_j\geq\frac{\Delta_j}{2}\big\},
\end{align*}
when we add and subtract the random variable $\frac{1}{i}\sum_{l=1}^i\mb{E}[\car{l}{j}|\ft{j}{l-1}]$ into the event $\omega$ and invoke the definition of $Y_{j, i}$ from~\eqref{eq:yk}. From Lemma~\ref{lem:boundforAH}, we obtain
\begin{equation}
\textstyle\omega \subset \textstyle \big\{\frac{Y_{j, i}}{i}+\frac{\lj{j}{i}}{i}\geq\frac{\Delta_j}{2}\big\}. 
\label{eq:greedy_sketch_omega}
\end{equation}
Moreover, $\frac{c}{2d^2}\geq \frac{c'\nu^2}{2\Delta_{\min}^2}$ since by construction $c \geq c'\nu^2$ and $0 \leq d \leq \Delta_{\min}$, so that
$x_0\geq \frac{c'\nu^2}{2\Delta_{\min}^2}$. This implies 
\begin{equation*}
\textstyle \frac{\Delta_j}{\sqrt{c'/2}}\geq \frac{\Delta_{\min}}{\sqrt{c'/2}}\geq \frac{\nu}{\sqrt{x_0}}\geq \frac{\lj{j}{x_0}}{x_0}.
\end{equation*}
Hence, $\frac{\Delta_j}{\sqrt{c'/2}}\geq \frac{\lj{j}{i}}{i}$ for all $i\geq \lfloor x_0\rfloor +1$ so that, recalling \eqref{eq:greedy_sketch_omega},
\begin{equation*}
\textstyle \omega\subset \big\{\frac{Y_{j,i}}{i}+\frac{\lj{j}{i}}{i}\geq \frac{\Delta_j}{2} \big\}\subset \big\{\frac{Y_{j,i}}{i}\geq \frac{\Delta_j}{2}-\frac{\Delta_j}{\sqrt{c'/2}} \big\}=\big\{\frac{Y_{j,i}}{i}\geq
\frac{\Delta_j(\sqrt{c'/2}-2)}{\sqrt{2c'}}\big\}.
\end{equation*}
Now, we apply the Azuma--Hoeffding inequality from Proposition~\ref{prop:AH} to $\omega$ to get that
\begin{equation*}
\textstyle P\big(\barR{i}{j} \geq \mu_j + \frac{\Delta_j}{2} \big) \leq P\big(\frac{Y_{j,i}}{i}\geq \frac{\Delta_j(\sqrt{c'/2}-2)}{\sqrt{2c'}}\big) \leq \exp\big(-\frac{i}{2}\big(\frac{\Delta_j(\sqrt{c'/2}-2)}{\sqrt{2c'}} \big)^2\big) = \exp\big(-\frac{i\Delta_j^2}{c''}\big).
\end{equation*}
Finally, we can show that $\sum_{i=\lfloor x_0\rfloor+1}^k\exp\big(-\frac{i\Delta_j^2}{c''}\big) \leq \frac{c''}{\Delta_j^2}\exp\big(-\frac{\Delta_j^2\lfloor x_0\rfloor}{c''}\big)$. Plugging this bound into~\eqref{eq:greedy_sketch_to_bound} and then relating that bound back to~\eqref{eq:greedy_sketch_final} yields our final result. 
\end{proof}

\begin{remark}
Theorem~\ref{thm:greedy} gives a bound on the probability of a suboptimal arm being selected at an epoch $k \geq \lceil \frac{cm}{d^2} \rceil$ that is equivalent for discount-averaged reward feedback and time-averaged reward feedback. However, the type of reward feedback does impact the bound in Theorem~\ref{thm:greedy} as a consequence of the size of the constant $c$, which derives from the constant $\kappa$ that depends on $\lj{j}{\cdot}$, and hence, the type of reward feedback. While the bound for each type of reward feedback is of identical order, when the constant $c$ grows with the constant $\kappa$ through $\lj{j}{\cdot}$ it is likely that in practice the empirical performance will degrade for benign problem instances since the algorithm will explore for a longer period of time. Refer back to Remark~\ref{remark:lj_size} for a discussion of how $\lj{j}{\cdot}$ depends on the discount factor and the type of reward feedback. 
\end{remark}

Theorem~\ref{thm:greedy} gives an instantaneous bound on the probability of selecting a suboptimal arm. To obtain a bound on the number of times a suboptimal arm will be played, we can sum the probability of the arm being played at each epoch $k \in [n]$. Observe that when $c > c''$, for $k\in [n]$ sufficiently large enough, $p(\alpha(k)=j) \leq \frac{c}{d^2k} + o(\frac{1}{k})$ for any suboptimal arm $j \in [m]$. Hence, $\mb{E}_{\alpha}[T_j^\alpha(n)] = \sum_{k=1}^np(\alpha(k)=j)  \leq \frac{c}{d^2}\mathcal{O}(\log(n))$. This implies that when we relate back to the regret proposition from Proposition~\ref{thm:regretdecomp1}, we can obtain a gap-dependent regret bound.

\begin{corollary}[Gap-Dependent Regret Bound for EpochGreedy]
Under the assumptions of Theorem~\ref{thm:greedy},
\setlength{\belowdisplayskip}{0pt}\setlength{\belowdisplayshortskip}{0pt}
\begin{equation*}
R^{\mathrm{EpochGreedy}}(n) \leq \textstyle \mathcal{O}(\log(n))\frac{c}{d^2}\sum_{j\neq j_\ast}\Delta_j+\sum_{j\in [m]}\lj{j}{n}.
\end{equation*}
\label{corr:greedy}
\end{corollary}

\begin{remark} 
The gap-dependent regret bound of EpochGreedy is of equal asymptotic order as the gap-dependent regret bound we obtained for EpochUCB in Corollary~\ref{corr:ucb}. However, as opposed to having a dependence on the reward gaps of the form $\sum_{j\neq j_{\ast}}\Delta_j^{-1}$ as in the regret bound for EpochUCB, the dependence on the reward gaps found in the regret bound for EpochGreedy is of the form $d^{-2}\sum_{j\neq j_{\ast}}\Delta_j$, which can be significantly worse for certain problem instances. Moreover, Theorem~\ref{thm:greedy} relies on knowledge of a lower bound on the minimum reward gap given by the constant $d$ that would typically not be available \emph{a priori}. That being said, EpochGreedy is simple and near-greedy approaches often perform well in practice.
\end{remark}

%% file: Algorithms/greedyalg.tex
\begin{algorithm}[htbp!]
\caption{EpochGreedy}
\label{alg:greedy}
\begin{algorithmic}[1] 
\Procedure{EpochGreedy}{$\tauz$, $\zeta$, $\gamma$, $c$, $d$}
\Comment{$c\geq c'\nu^2$, $0<d\leq \Delta_{\min}$}
\State $t_1\gets 0$, $T_j\gets 0$ {\small \&} $\barR{T_j}{j}\gets 0 \ \forall \ j\in [m]$, $k \gets 1$ 
\While{$k>0$}\Comment{EpochGreedy}\\
$\qquad$$i_k=\arg\max_j \barR{T_j}{j}$ \Comment{Find arm with maximum empirical mean reward}\\
$\qquad$$\vep_k=\min\{ 1,\frac{cm}{d^2k} \}$ \Comment{Random Exploration Probability}\\
$\qquad$\textbf{If}{ $\text{rand}(\cdot)\geq \vep_k$}: $i\gets i_k$
 \textbf{else}: $i\gets \text{randint}(m)$\Comment{Equation~\ref{eq:greedy_policy}}\\
\qquad $\taukr \gets \tauz + \zeta T_i$ \Comment Equation~\ref{eq:tauk}\\
    $\qquad$$\br{k}{i}\gets$ \Call{pullarm}{$i$, $k$, $\gamma$,
    $t_k$, $\taukr$} \Comment{Algorithm~\ref{alg:armpull}}\\
    $\qquad$$\barR{T_i + 1}{i} \gets \barR{T_i}{i} + (\br{k}{i}-\barR{T_i}{i})/T_i$ \\
    $\qquad$$T_i\gets T_i+1$, $t_{k+1}\gets t_k+\taukr$, $k\gets
    k+1$
\EndWhile\label{euclidendwhile}
\EndProcedure
\end{algorithmic}
\end{algorithm}

%% file: Main_Sections/experiments.tex
In this section, we present a set of illustrative experiments to enhance our theoretical results. To begin, we describe the class of problem instances we run our experiments on. Following this, we show theoretical and empirical regret comparisons between EpochUCB and EpochGreedy and examine each quantity as a function of the discount factor that governs the reward feedback. Finally, we compare the empirical performance of EpochUCB and EpochGreedy with a variety of alternative algorithms existing in the literature. 

Before doing so however, it is worth revisiting Example~\ref{ex:cuteex} to understand why popular bandit approaches fail to achieve good performance in a correlated Markovian environment, and to elucidate how our approach overcomes the issues plaguing conventional methods. 

\textbf{Revisiting Example~\ref{ex:cuteex}:} Why do EpochUCB and EpochGreedy work? First, suppose $\tau_0,\zeta = 1$. Then, EpochUCB and EpochGreedy initially estimate the empirical mean reward of arm $1$ to be zero in Example~\ref{ex:cuteex}. However, during the exploration phase, arm $1$ is played again for an epoch length $> 1$ and the state transitions from $\theta_1$ to $\theta_2$ within the epoch. Therefore, the observed reward feedback tends closer to stationary distribution reward, and the empirical mean reward for arm $1$ increases. This continues as the epoch length increases so that eventually the empirical mean reward for arm $1$ exceeds that of arm $2$ and each algorithm correctly identifies arm $1$ as the optimal arm prior to moving from exploration to exploitation. We can recall that UCB and $\vep$--greedy fell victim to under-estimating the stationary distribution reward of the optimal arm since the state distribution was never allowed to converge toward the stationary distribution. Effectively, UCB and $\vep$--greedy were estimating the stationary distribution reward of the optimal arm based on a reward distribution, which was arising from a state distribution deviating significantly from the stationary distribution, that had much worse expected reward.

\subsection{Problem Generation}\label{sec:generation}
\begin{figure*}[t]
\centering
\subfloat[][]{\includegraphics[width=0.4\textwidth]{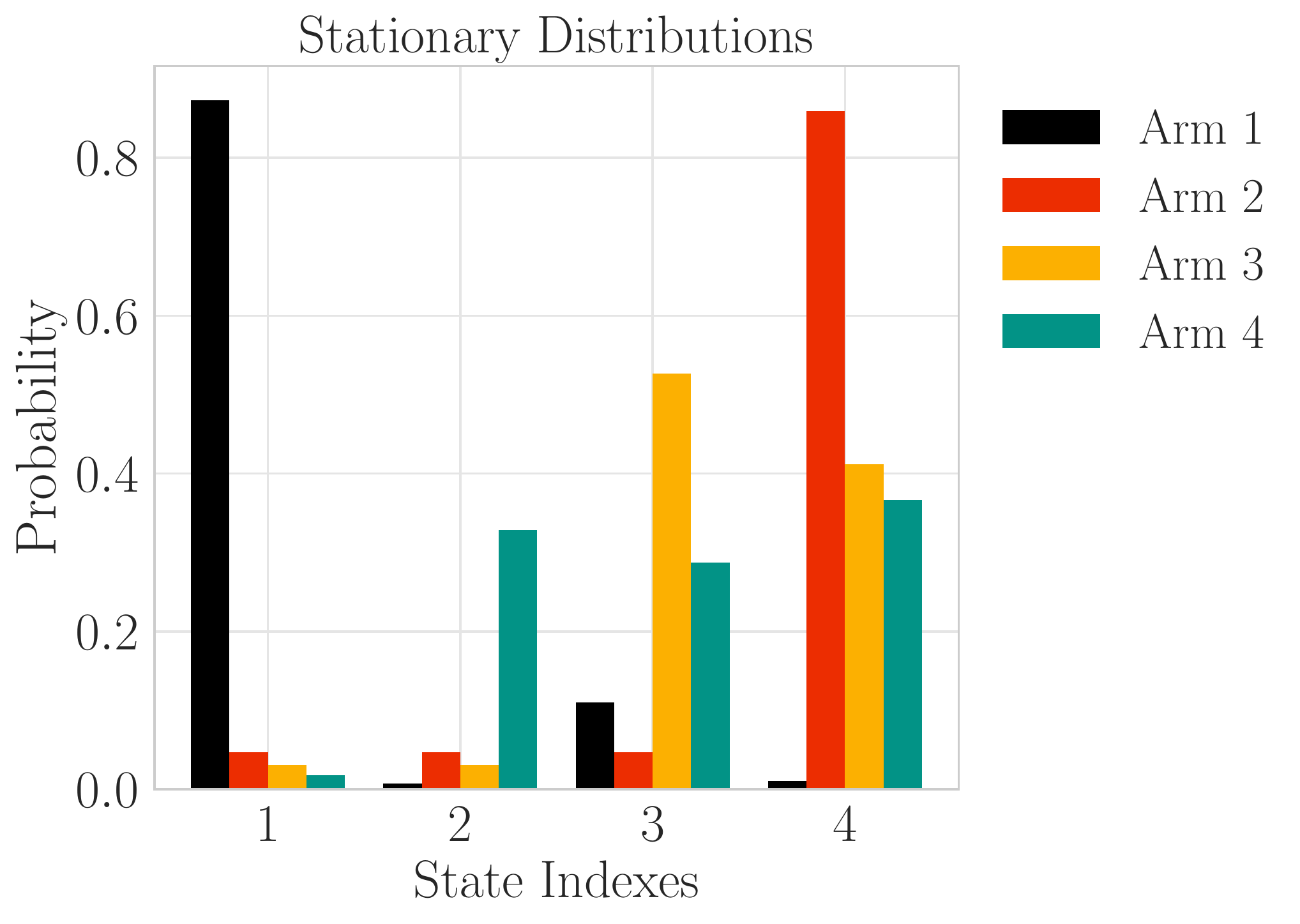}
\label{fig:stationary_dist}}
\subfloat[][]{\includegraphics[width=0.4\textwidth]{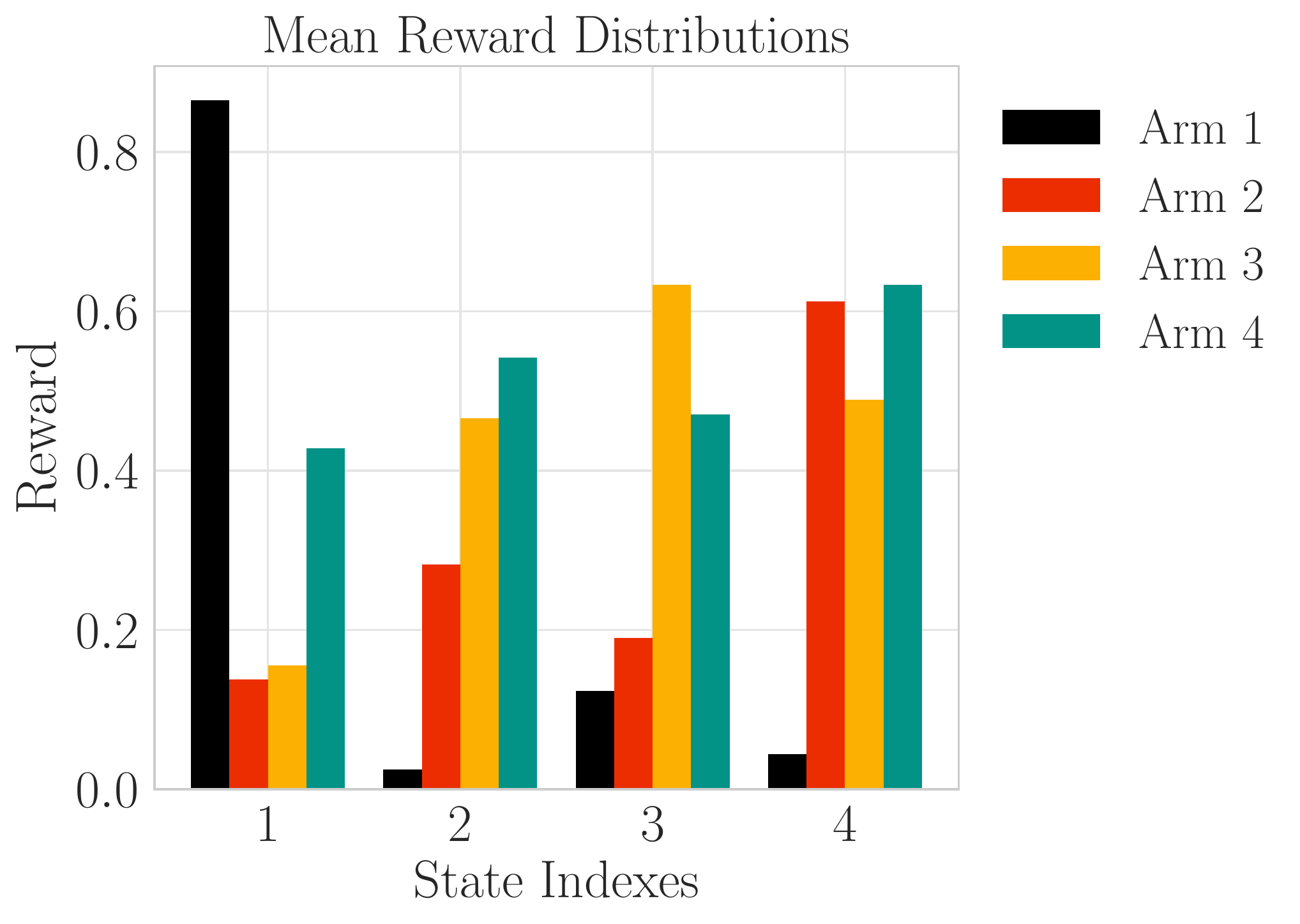}
\label{fig:reward_dist}}
\caption{(a) Stationary distributions indexed by the state for each arm in a sample $4$-arm, $4$-state instance; (b) Mean of the reward distributions indexed by the state for each arm for the same instance as (a).}
\label{fig:comparison}
\end{figure*}
For our experiments, we consider a class of problem instances exhibiting interesting structures, i.e., the arms are sufficiently unique but strongly anti-correlated. We believe this to be reasonably representative of actual instances where an agent has vastly different preferences depending the underlying state and the presented arm. We generate the problem instances with the following criteria on the transition matrices and the reward distributions.

\textbf{Transition Matrices:} We consider a scenario in which the transition matrix for the optimal arm is such that the state is constrained to be in a subset of the state space with high probability. Moreover, the transition matrices of each suboptimal arm are such that the state enters this subset of the state space with low probability. An illustration of the stationary distribution for a sample instance having this property is given in Figure~\ref{fig:stationary_dist}. Each transition matrix in our experiments was inspected to ensure Assumptions~\ref{assumption:ergodic} and~\ref{assumption:reverse} were satisfied.

\textbf{Reward Distributions:} The reward distribution for each arm, state pair is randomly chosen to be a beta, Bernoulli, or uniform distribution. The mean of each distribution is randomly selected with the caveat that the means are increasing in the stationary distribution probability of the states. Figure~\ref{fig:reward_dist} shows the mean reward for each reward distribution corresponding to an arm, state pair for a sample instance. 

\subsection{EpochUCB vs.~EpochGreedy}
\begin{figure*}[t]
\centering
\subfloat[][]{\includegraphics[width=0.31\textwidth]{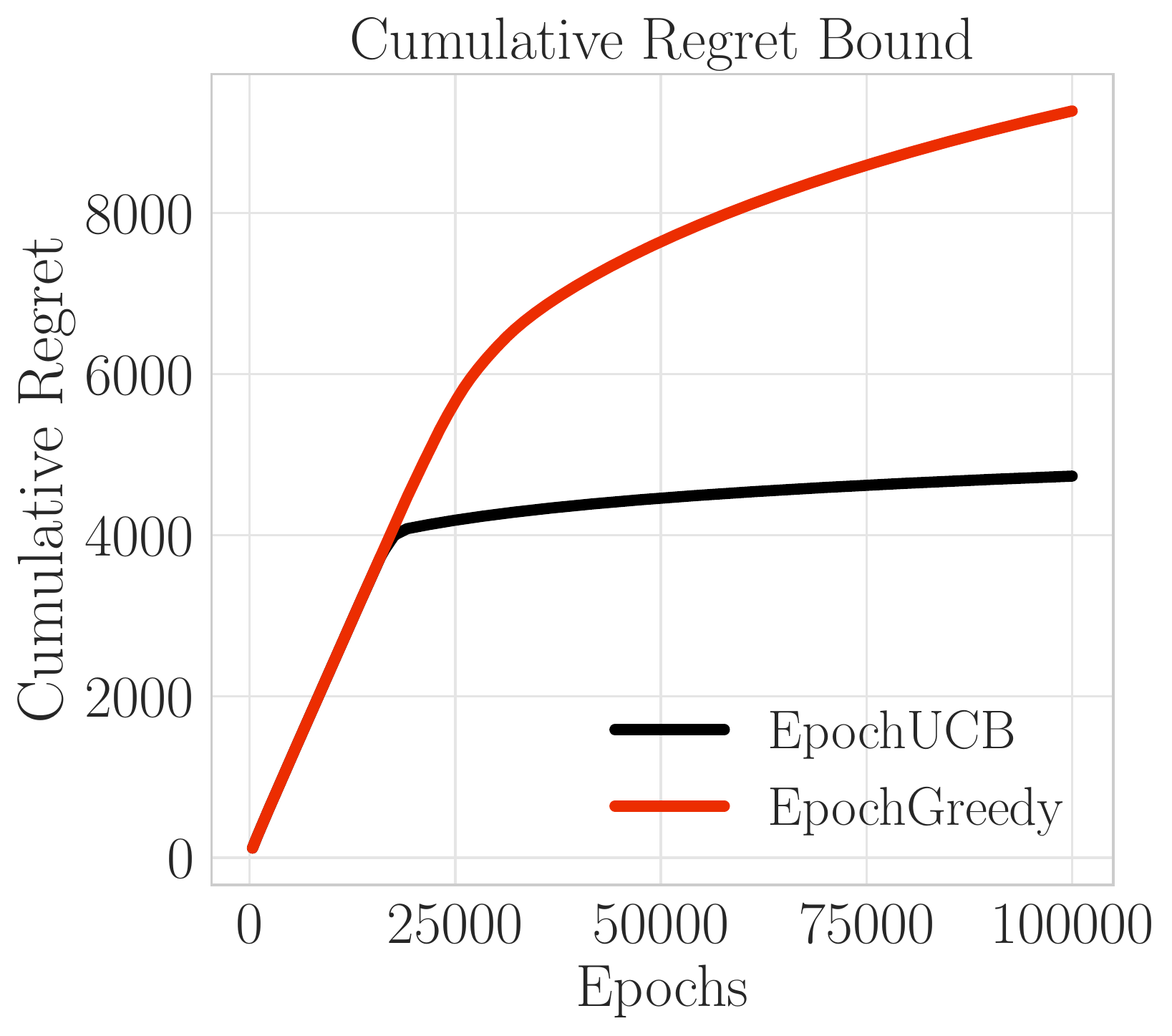}
\label{fig:Epoch_Theoretical_Comparison}}
\subfloat[][]{\includegraphics[width=0.31\textwidth]{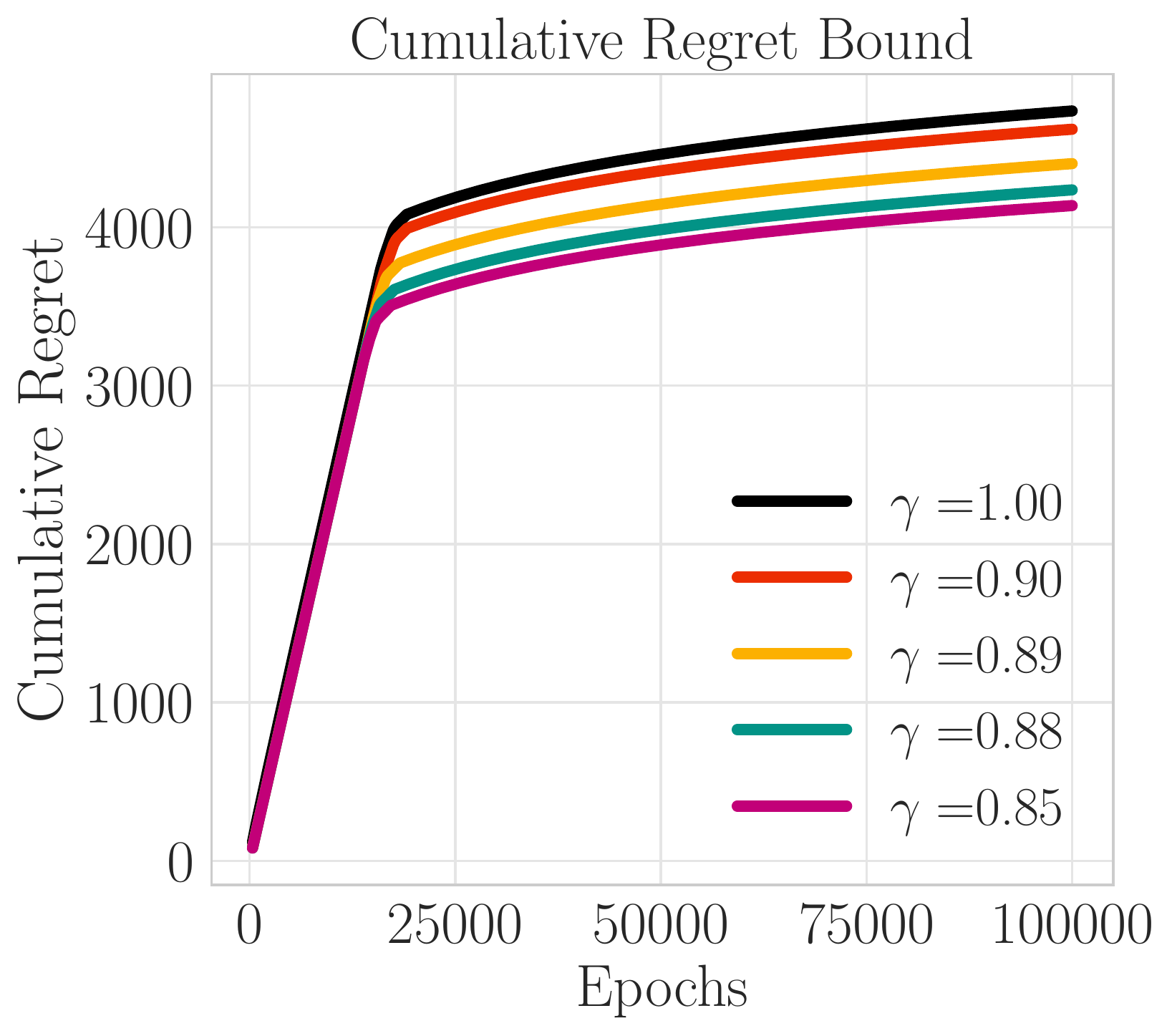}
\label{fig:EpochUCB_Theoretical_Discount}}
\subfloat[][]{\includegraphics[width=0.31\textwidth]{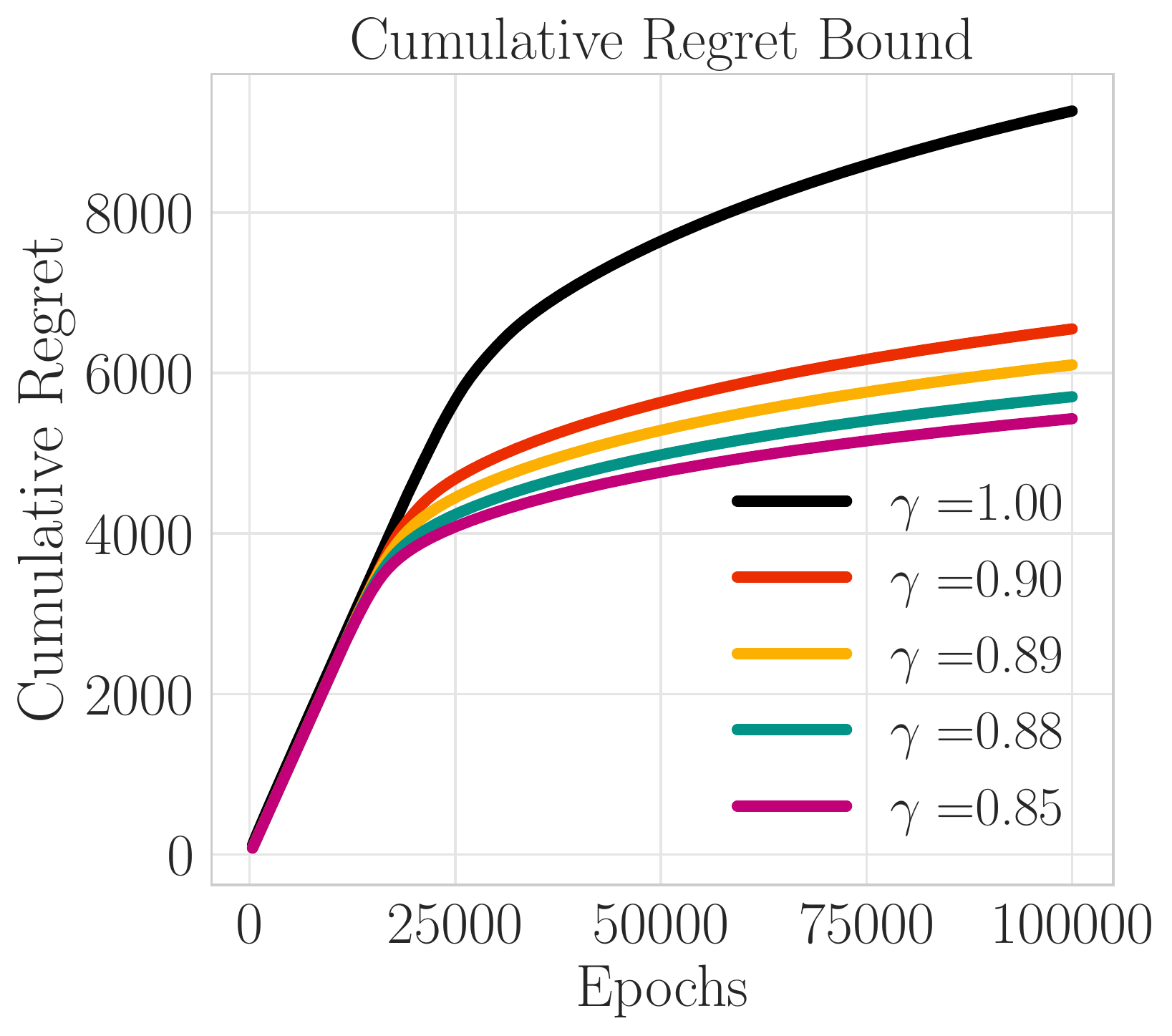}
\label{fig:EpochGreedy_Theoretical_Discount}}
\caption{(a) Mean theoretical gap-dependent regret bounds under time-averaged reward feedback for EpochUCB and EpochGreedy over $5$ problem instances from the class described in Section~\ref{sec:generation} with $4$ arms and $4$ states; (b--c) Mean theoretical gap-dependent regret bounds for EpochUCB (Figure~\ref{fig:EpochUCB_Theoretical_Discount}) and EpochGreedy (Figure~\ref{fig:EpochGreedy_Theoretical_Discount}) under discount-averaged reward feedback over the problem instances in (a) with various discount factors.}
\label{fig:theoretical}
\end{figure*}
We now compare the theoretical gap-dependent regret bounds and empirical performance of EpochUCB and EpochGreedy. In this section, we set $\tauz=40, \zeta=1$ and for EpochGreedy let the constant $c=c'\nu^2$ where $c'$ was selected to minimize the cumulative regret subject to the constraint $c' > 8$. Moreover, for each simulation we present the mean results over $5$ problem instances sampled from the class we consider as described in Section~\ref{sec:generation} with $4$ arms ($m=4$) and $4$ states ($|\Theta|=4$).

\subsubsection{Theoretical Comparison}
In Figure~\ref{fig:Epoch_Theoretical_Comparison} we compare the theoretical gap-dependent regret bounds of EpochUCB and EpochGreedy under time-averaged reward feedback. As our theoretical results indicate, each regret bound grows logarithmically and EpochUCB's regret bound is tighter than EpochGreedy's owing to the superior dependence on the reward gaps and the sharper constants. In Figures~\ref{fig:EpochUCB_Theoretical_Discount} and~\ref{fig:EpochGreedy_Theoretical_Discount} we examine the theoretical gap-dependent regret bounds of EpochUCB and EpochGreedy under discount-averaged reward feedback as a function of the discount factor on the rewards. For these problem instances, and as we would expect to be the case generally since the majority of weight is given to rewards as the state distribution tends closer to a stationary distribution (see Remark~\ref{remark:lj_size} for further discussion on this point), the gap-dependent theoretical regret decays as the discount factor decays until the improvement begins to saturate. 

\subsubsection{Empirical Comparison}
\begin{figure*}[t]
\centering
\subfloat[][]{\includegraphics[width=0.31\textwidth]{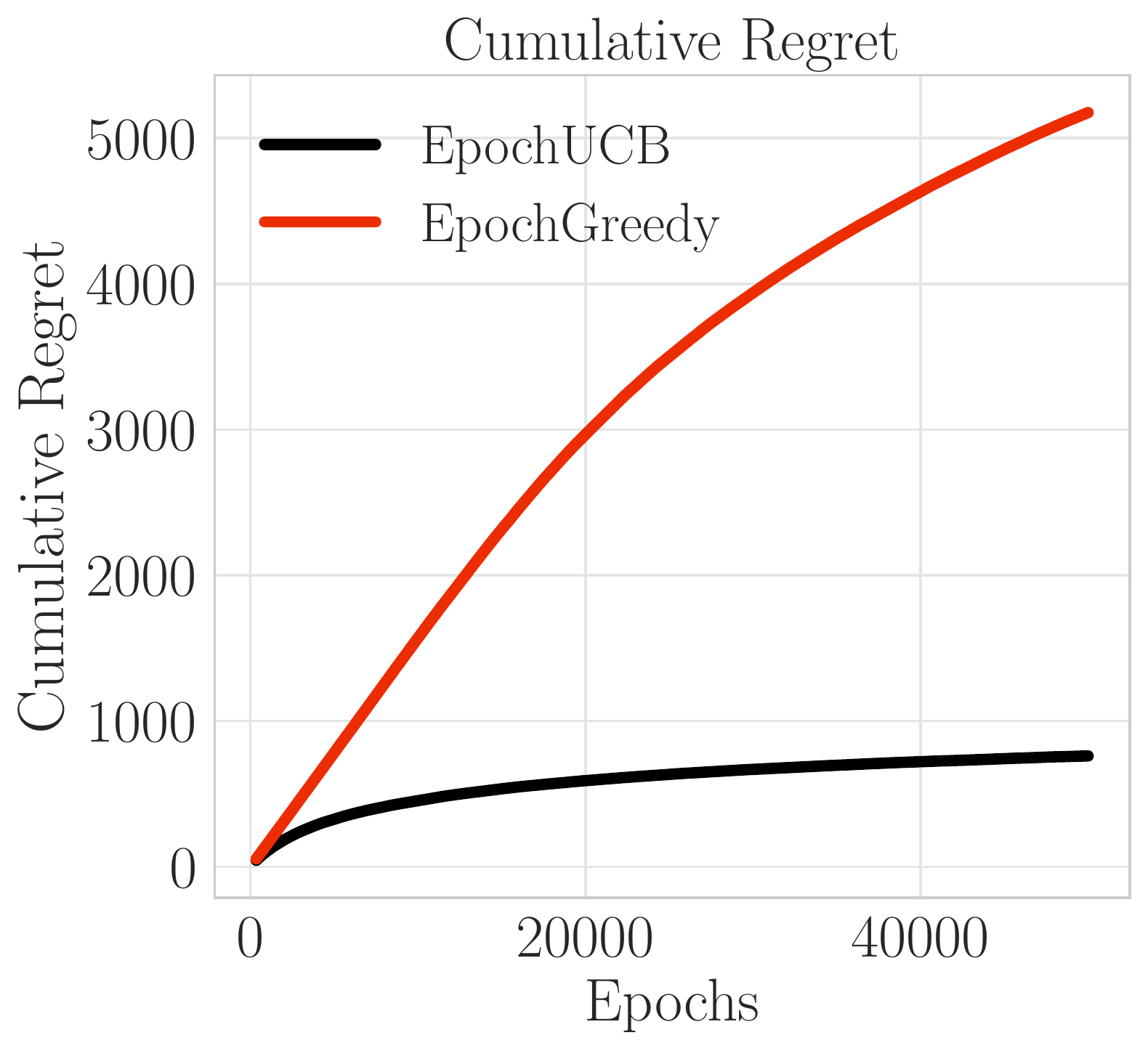}
\label{fig:Epoch_Comparison}}
\subfloat[][]{\includegraphics[width=0.43\textwidth]{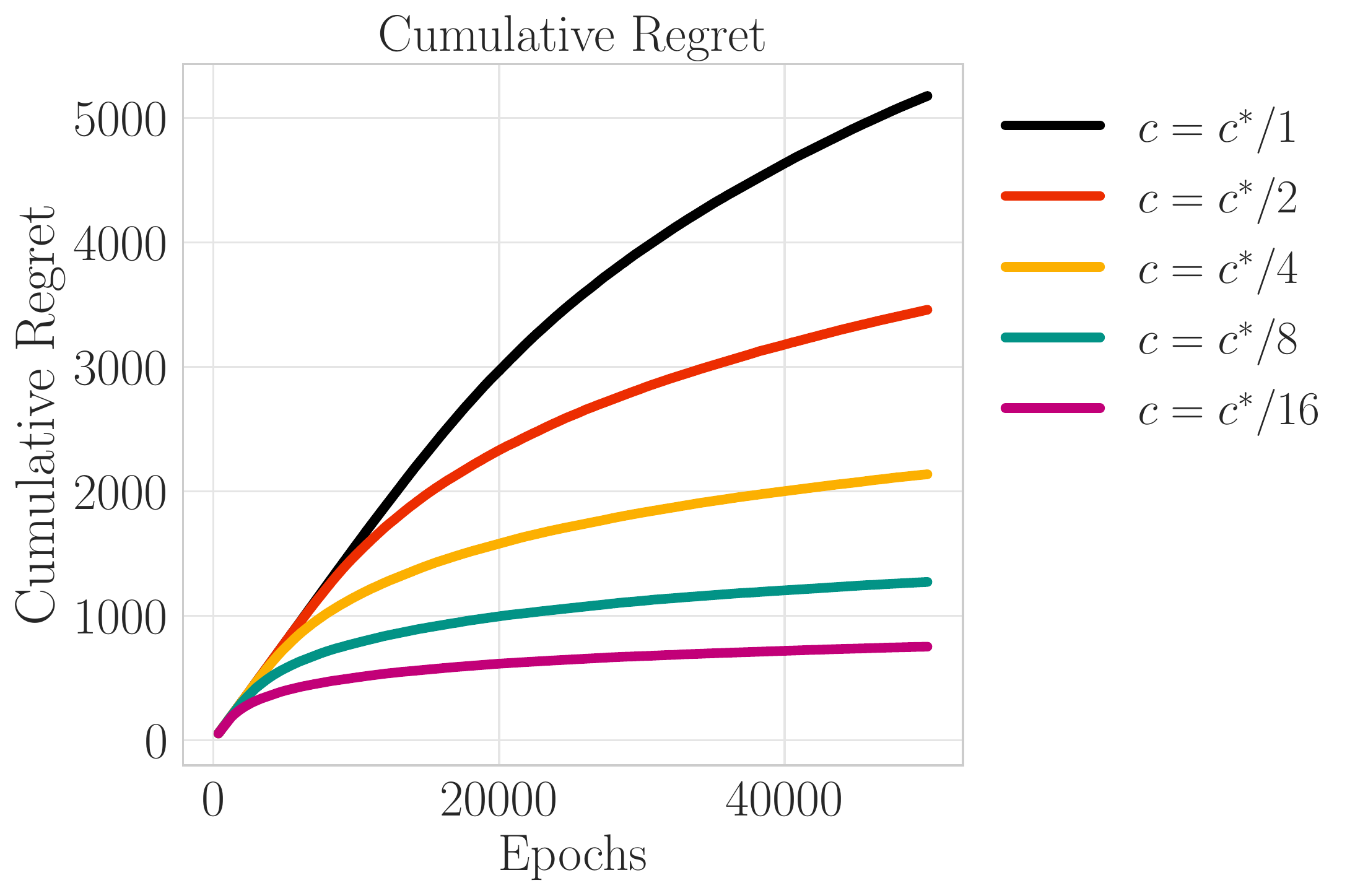}
\label{fig:EpochGreedy_Tuned}}
\caption{(a) Mean empirical regret for EpochUCB and EpochGreedy under time-averaged reward feedback over the problem instances in Figure~\ref{fig:Epoch_Theoretical_Comparison}; (b) Mean empirical regret for EpochGreedy under time-averaged reward feedback over the problem instances in (a) as the theoretically optimal constant $c^{\ast}$ to minimize the regret is decayed.}
\label{fig:Empirical_1}
\end{figure*}

\begin{figure*}[t]
\centering
\subfloat[][]{\includegraphics[width=0.31\textwidth]{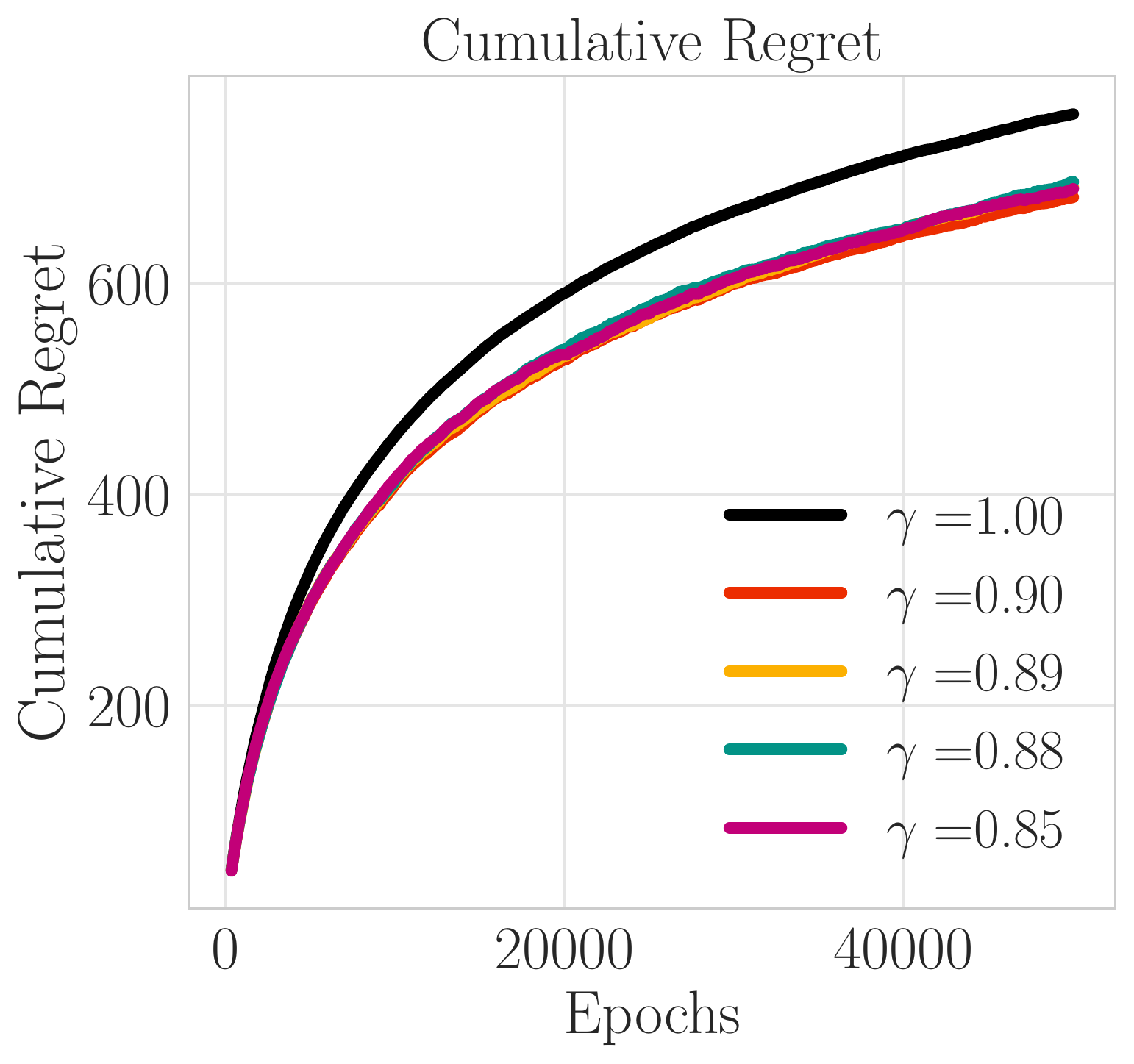}
\label{fig:EpochUCB_Discount}}
\subfloat[][]{\includegraphics[width=0.43\textwidth]{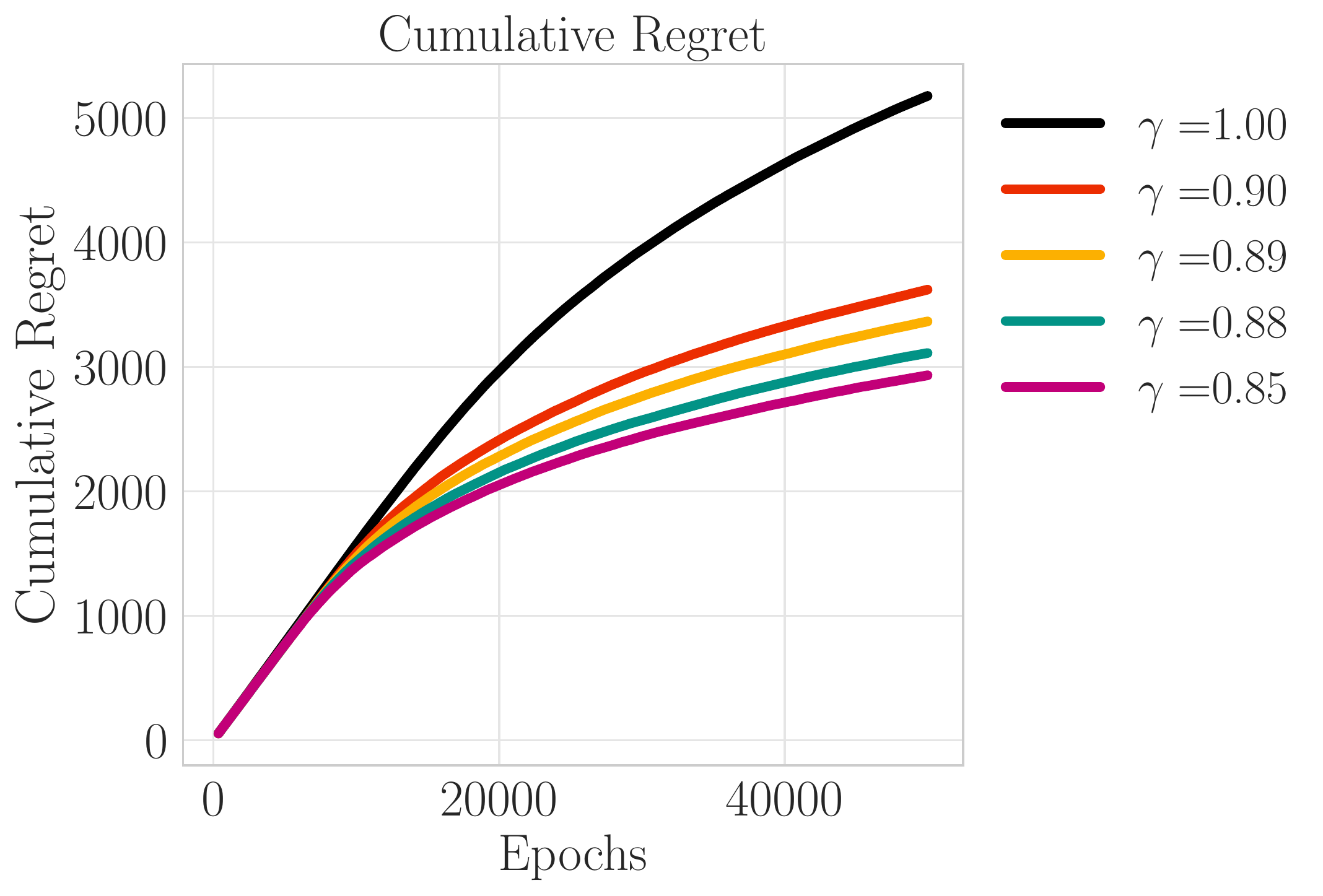}
\label{fig:EpochGreedy_Discount}}
\caption{(a--b) Mean empirical regret for EpochUCB (Figure~\ref{fig:EpochUCB_Discount}) and EpochGreedy (Figure~\ref{fig:EpochGreedy_Discount}) under discount-averaged reward feedback over the problem instances in Figure~\ref{fig:Epoch_Theoretical_Comparison} for various discount factors.}
\label{fig:Empirical_2}
\end{figure*}

In Figure~\ref{fig:Epoch_Comparison} we compare the empirical performance of EpochUCB and EpochGreedy under time-averaged reward feedback and observe that EpochUCB significantly outperforms EpochGreedy. However, in Figure~\ref{fig:EpochGreedy_Tuned} we show that EpochGreedy's empirical performance improves dramatically and tends toward the empirical performance of EpochUCB as we decay the constant parameter $c$ from that which was selected to minimize the theoretical regret. This phenomenon matches the empirical conclusions drawn with respect to the $\vep$--greedy algorithm in work of~\citet{auer:2002aa}. In Figures~\ref{fig:EpochUCB_Discount} and~\ref{fig:EpochGreedy_Discount}, we examine the empirical performance of EpochUCB and EpochGreedy under discount-averaged reward feedback as a function of the discount factor. Similar to as with the theoretical regret, we find that the cumulative regret decays as a function of the discount factor.

\subsection{Comparison to Existing Algorithms}\label{sec:comparison}
To conclude our simulations, we compare the performance of EpochUCB and EpochGreedy with UCB and $\vep$--greedy, as well as several other well-studied algorithms from the literature that we now detail.

\textbf{Variance-Tuned UCB:} Following~\citet{auer:2002aa}, we consider a tuned version of UCB where the confidence windows is replaced with a confidence window on the variance. We refer to this variant of UCB as UCB$+$. This variant is known to often perform well in practice for many instances.

\textbf{EXP3:} We compare our proposed algorithms to the EXP3 algorithm of~\citet{auer2002nonstochastic} for adversarial bandits. EXP3 has been proven to omit sublinear regret against an oblivious adversary---meaning that an adversary can select the sequence of rewards with knowledge of the algorithm but this must be done \textit{a priori}. However, since the reward feedback in the problem we study depends on the history of actions, our model can be seen as a type of adaptive adversary and therefore these bounds do not necessarily hold for the problem being considered. In fact, the regret can be linear for EXP3 against an adaptive adversary~\citep{dekelTA12}.

\textbf{Continuous Reinforcement Learning (RL):} We compare to $Q$-learning with linear function approximation as expressed in~\citet{melo2008analysis} and~\citet{geramifard2013tutorial}. We allow the reinforcement learning algorithm to observe the state distribution prior to each decision point and use the state distribution as the features of the linear model. Hence, the RL algorithm has access to a form of partial state observation that is not available to the bandit algorithms since the state is drawn from the observed state distribution. Linear function approximation is a natural choice given that the expected stationary distribution reward is a linear combination of the stationary distribution rewards and the stationary distribution. For the simulations, we decay the randomness of the $\vep$--greedy policy exponentially and for the weight updates use a step size of $1/\sqrt{k}$ to satisfy the conditions of~\citet{robbins1985stochastic}. For further details on the algorithm and implementation, see Appendix~\ref{app:experiments}.

\begin{figure*}[t]
\centering
\subfloat[][]{\includegraphics[width=0.45\textwidth]{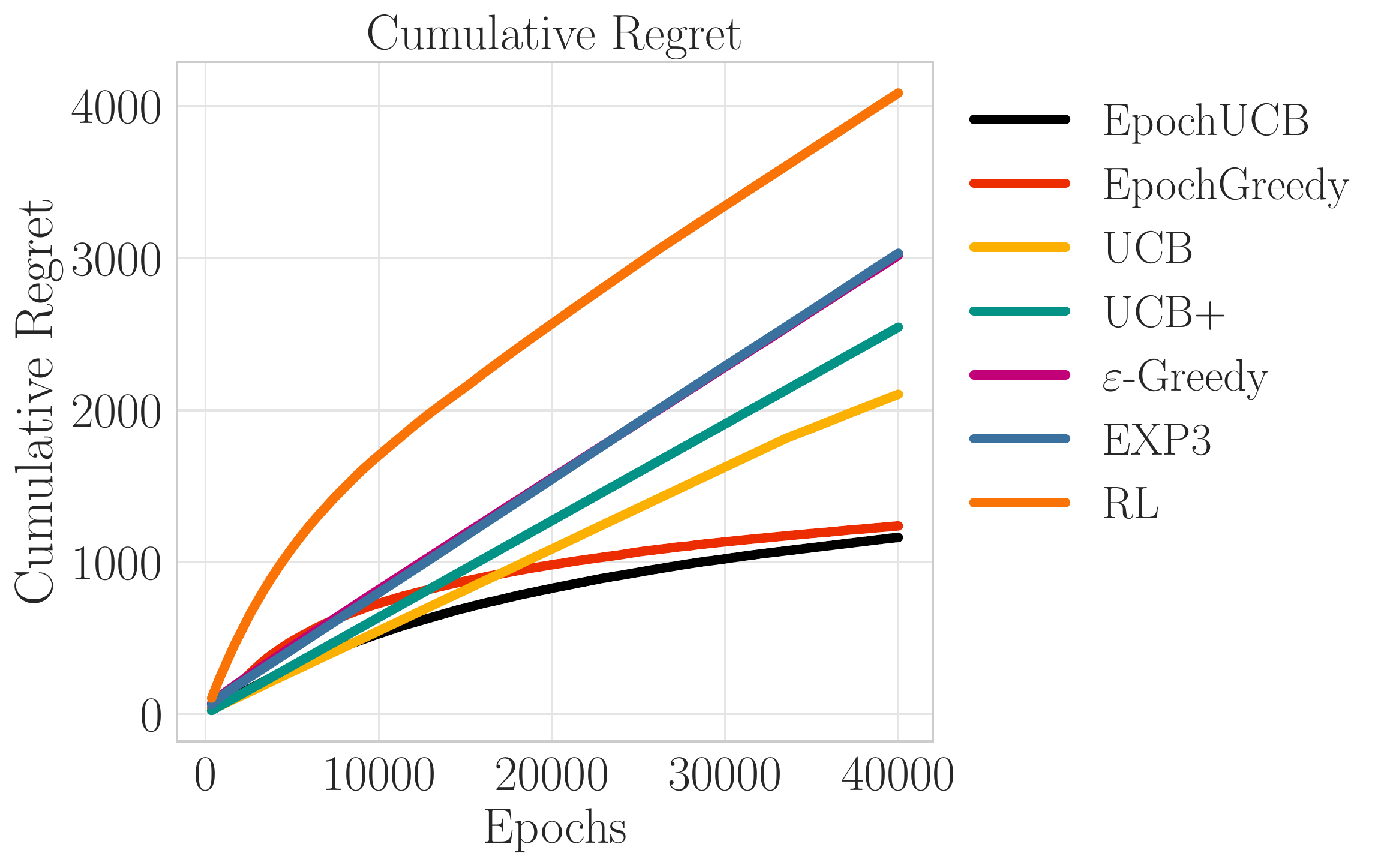}
\label{fig:Algorithm_Comparison1}}
\subfloat[][]{\includegraphics[width=0.45\textwidth]{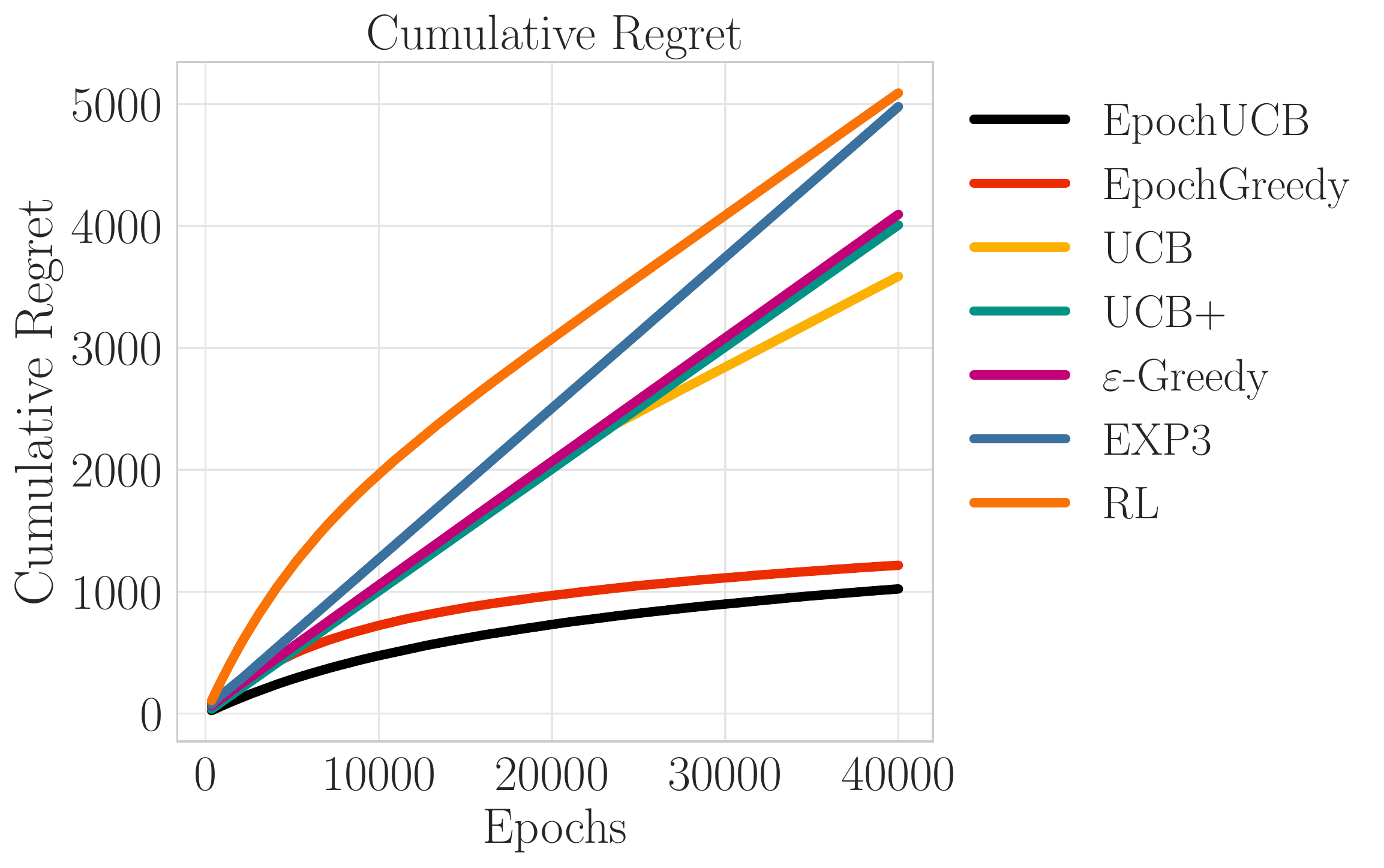}
\label{fig:Algorithm_Comparison2}}
\caption{Mean empirical regret under time-averaged reward feedback for each of the algorithms over $5$ problem instances from the class described in Section~\ref{sec:generation} with $4$ arms and $4$ states in (a) and $4$ arms and $8$ states in (b).}
\label{fig:comparison}
\end{figure*}

In this section, we again set $\tauz=40, \zeta=1$, but motived by Figure~\ref{fig:EpochGreedy_Tuned} we tune the constant $c$ for EpochGreedy to yield favorable empirical performance. We show the mean performance of each of the algorithms under time-averaged reward feedback over $5$ problem instances sampled from the class we consider as described in Section~\ref{sec:generation} with $4$ arms and $4$ states in Figure~\ref{fig:Algorithm_Comparison1} and $4$ arms and $8$ states in Figure~\ref{fig:Algorithm_Comparison2}. To allow for a fair comparison, we run each algorithm for equal number of total iterations and convert the regret from the algorithms that select actions at each iteration to epochs. As expected, EpochUCB and EpochGreedy demonstrate the sublinear convergence rates we proved. On the other hand, each of the algorithms we compare to performs poorly and suffers linear regret. Given Example~\ref{ex:cuteex}, this is unsurprising for UCB, UCB$+$, and $\vep$--greedy. For Exp3, we attribute the poor performance to the adaptive nature of the rewards, as it is known that EXP3 has linear regret against an adaptive adversary~\citep{dekelTA12}, e.g., when the reward in each round depends on the actions selected in previous rounds. Furthermore, although the RL algorithm has more information than the bandit algorithms, its performance is also poor. This is partly due to the continuous state space, the fact that the algorithm only has access to a distribution over states and not the exact state, and that it switches between actions too rapidly. 

Finally, in settings that do not have strong correlation, EpochUCB and EpochGreedy may converge slower than some of the alternatives. However, as opposed to the algorithms we compare to, EpochUCB and EpochGreedy will always perform adequately asymptotically, meaning that the optimal arm will still be identified correctly.

%% file: Main_Sections/discussion.tex
We study a multi-armed bandit problem in which a decision-maker repeatedly faces an agent with an unobserved state variable that influences the reward distribution on each arm and this state variable evolves with a Markov chain whose transition matrix depends on the action of the decision-maker. This work is motivated by applications pertaining to interactions between a decision-maker and an agent, a general example being digital platforms that actively engage with users, where the agent's underlying state is not static. The novelty and technical challenges in this problem formulation stem from the decision-maker obtaining no information about the state or distribution of the Markov chain and the fact that observed rewards are correlated with past actions. Despite the generality of the problem formulation, we developed algorithms called EpochUCB and EpochGreedy with sublinear regret guarantees. Concretely,
we proved $\mathcal{O}(\log(n))$ gap-dependent regret bounds for each of our proposed algorithms as well as an $\mathcal{O}(\sqrt{n\log(n)})$ gap-independent regret bound for EpochUCB. Moreover, our simulations empirically validate our methods and demonstrate the insufficiency of existing bandit and reinforcement learning algorithms for the problem at hand.

As is standard in bandit problems with Markovian rewards, we focus on the weak regret measure. An interesting and challenging question for future work is whether it is possible to obtain sublinear regret guarantees under stronger notions of regret. A clear example is when the benchmark policy being compared to is the optimal policy within the class of state dependent policies. However, it may be of interest to pursue dynamic regret measures that could be more attainable. A well-known dynamic regret measure is that considered in~\citet{garivier2011upper}. This regret measure, proposed for bandit problems where the reward distributions can change abruptly, compares against the arm that yields the maximum reward at each time along the horizon. Unfortunately, there is not a clear translation of this regret measure to the problem we study since the optimal arm at each time depends on the history of arm selections and state variables, and hence, is not fixed \emph{a priori}. A promising direction in identifying suitable dynamic regret measures in bandit problems with Markovian rewards is the dynamic path-dependent regret measure proposed in~\citet{cortes2017discrepancy} for rested bandit problems. This regret measure compares at each time index to the optimal action given the history of actions and feedback until that time index. An adaptation of this concept to the problem formulation we consider is worth exploring.

Finally, given that we derived a general framework for analyzing the regret of any multi-armed bandit policy interacting with a correlated Markovian environment in which the observed feedback is a smoothed reward over an epoch, there is potential to extend alternative existing algorithms from the bandit literature to the problem formulation of this paper. Of potential interest may be algorithms with weaker theoretical guarantees, but which are known to perform well in practice, such as variance-tuned algorithms.

%% file: Appendix_Sections/notation_table.tex
\begin{figure}[H]
\setlength{\tabcolsep}{4pt}
\begin{tabular}{|l|l|}\hline
\textbf{Notation} & \textbf{Meaning}\\\hline\hline

$[m]$ & Set of arms: $\{1,\dots,m\}$  \\\hline


$[n]$ & Epoch horizon: $\{1, \dots, n\}$ \\\hline


$\alpha(k)$ & Arm pulled at epoch $k$ under policy $\alpha$ \\\hline


$T_j^{\alpha}(n)$ & Number of times arm $j$ is pulled in $n$ epochs under policy $\alpha$\\\hline

$R^{\alpha}(n)$ & Cumulative regret after $n$ epochs under policy $\alpha$ \\\hline

$\tauz, \zeta$ & Initial epoch size and linear growth term in epoch length sequence \\\hline

$\taukr$ & $\#$ of iterations in epoch $k$ under policy $\alpha$: $\tauz + \zeta T_{\alpha(k)}^\alpha(k-1)$ \\\hline

$\taukj$ & $\#$ of iterations in epoch $k$ when $\alpha(k) = j$: $\tauz + \zeta T_j^\alpha(k-1)$ \\\hline

$\tau_j^i$ & $\#$ of iterations in the epoch arm $j$ is pulled for $i$--th time: $\tauz + \zeta (i-1)$ \\\hline

$t_k$ & Iteration index at the start of epoch $k$ \\\hline

$t_j^i$ & Iteration index at the start of the epoch arm $j$ is pulled for $i$--th time \\\hline

$\thetarv{t_k}$ & State variable at iteration $t_k$  \\\hline

$\thetarv{k}$ & Sequence of state variables in epoch $k$: $\{\theta_t\}_{t=t_k}^{t_{k+1}-1}$  \\\hline

$\beta_{t_k}$ & State distribution at iteration $t_k$ \\\hline

$\pi_j$ & Unique and positive stationary distribution of arm $j$ \\\hline

$P_{j}$ & Transition probability matrix for arm $j$ \\\hline

$P_{j}^{\taukr}(\theta, \theta')$ & Probability of moving from $\theta$ to $\theta'$ in $\taukr$ iterations when arm $j$ is pulled \\\hline

$\gamma$ & Discount factor on rewards $\in (0, 1]$\\\hline

$\bsgr_k$ & Sum of discount factors in epoch $k$ under policy $\alpha$: $\sum_{t=t_k}^{t_{k+1}-1}(\gamma)^{t_{k+1}-1-t}$ \\\hline

$\mc{T}_r(\theta, j)$ & Stochastic reward kernel for state $\theta$ and arm $j$  \\\hline

$r_j^\theta$ & Stationary reward of arm $j$ in state $\theta$ \\\hline

$\nr{t}{j}$ & Stochastic reward of arm $j$ at iteration $t$ drawn from $\mc{T}_r(\theta_t, j)$ \\\hline

$\br{k}{j}$ & Reward in epoch $k$ for arm $j$: $\frac{1}{\bsgr_k}\sum_{t=t_k}^{t_{k+1}-1} (\gamma)^{t_{k+1}-1-t} \nr{t}{j}$ \\\hline

$\car{i}{j}$ & Reward in the epoch arm $j$ is pulled for the $i$--th time \\\hline

$\barR{T_j}{j}$ & Mean epoch rewards for arm $j$ after $T_j$ selections: $\frac{1}{T_j}\sum_{i=1}^{T_j} \car{i}{j}$ \\\hline

$\mu_j$ & Expected stationary distribution reward for arm $j$: $\mb{E}[\sum_{\theta\in \Theta} r^\theta_j\pi_j(\theta)]$ \\\hline

$\Delta_j, \Delta_{\min},d$ & $\Delta_j = \mu_\ast - \mu_j$, $\Delta_{\min} = \min_{j\in [m]}\Delta_j$, $0 \leq d \leq \Delta_{\min}$ \\\hline

$\ft{j}{i}$ & Smallest $\sigma$-algebra generated by $(\car{1}{j}, \ldots, \car{i}{j},\thetarv{t_j^1}, \ldots,
\thetarv{t_j^i})$\\\hline

$Y_{j,T_j}$ & Martingale for arm $j$ pulled $T_j$ times: see~\eqref{eq:yk} \\\hline

$\lambda_j, \eta_j, \phi_j, \psi_j$ & $(\lambda_2(M(P_j)))^{1/2}$, $\eta_j = \min\{\gamma, \lambda_j\}$, $\phi_j = \max\{\gamma, \lambda_j\}$, and $\psi_j = \eta_j/\phi_j$ \\\hline

$C_j$ &  $1/2(1+(1-\min_{\theta}\pi_j(\theta))^2/\min_{\theta}\pi_j(\theta))^{1/2}$\\\hline

$\uj{j}{\cdot}, L_j^{\gamma}(\cdot), \rj{j}$ & Discount-averaged reward feedback: see~\eqref{eq:upsilon_discount},~\eqref{eq:lj1},~\eqref{eq:rhoj}  \\ & Time-averaged reward feedback: see~\eqref{eq:upsilon_time},~\eqref{eq:lj12},~\eqref{eq:rhoj-avg} \\\hline

$\cf{k}{\cdot}{j}$ & EpochUCB confidence window for arm $j$ at epoch $k$: see~\eqref{eq:confidence_window}  \\\hline

$\vep_k$ & EpochGreedy random exploration probability in epoch $k$: $\min\{1, \frac{cm}{d^2k}\}$  \\\hline

$c, c', c''$ & $c\geq c'\nu^2$, $c' > 8$, $c''=4c'(\sqrt{c'/2}-2)^{-2}$ \\\hline

$\nu, \kappa$ & $\nu = \max\{\kappa, \frac{d}{\sqrt{c'}}\}$, $\kappa = \min\{\kappa > 0: \kappa\sqrt{i} \geq \lj{j}{i} \ \forall \ i \in [n], j \in [m]\}$ \\ \hline

\hline
\end{tabular}
\end{figure}

%% file: Appendix_Sections/harmonic.tex
\begin{proof}(\emph{Proof of Harmonic Bound}).
For any $a > 0$ and positive integer $n$, we have that
\begin{equation}
\sum_{i=a}^{a+n}\frac{1}{i} \leq \frac{1}{a} + \log\Big(1 + \frac{n}{a}\Big).
\label{eq:sum_inequality}
\end{equation}
Indeed, rewrite the summation in~\eqref{eq:sum_inequality} as 
\begin{equation*}
\sum_{i=a}^{a+n}\frac{1}{i} = \frac{1}{a}  + \sum_{i=a+1}^{a+n}\frac{1}{i}
\end{equation*}
and apply the fundamental inequality $(i)^{-1} \leq \int_{i-1}^i (x)^{-1}dx$,
which holds for any $i > 1$ and is simply a consequence of $(i)^{-1}$ being a decreasing function, repeatedly for $i=a+1, a+2,
\ldots, a+n$ so that we have a telescoping summation of integrals---i.e.,
\begin{equation*}
\sum_{i=a}^{a+n}\frac{1}{i} = \frac{1}{a} + \sum_{i=a+1}^{a+n}\frac{1}{i} \leq
\frac{1}{a} + \int_{a}^{a+n}\frac{1}{x}dx  = \frac{1}{a} +\log\Big(1 + \frac{n}{a}\Big).
\end{equation*}
Thus, for $\tauz$ and $\zeta$ in $\mb{Z}_{+}$,
\begin{equation*}
\sum_{i=1}^{n} \frac{1}{\tauz+\zeta (i-1)} = \frac{1}{\zeta}\sum_{i=\tauz/\zeta}^{\tauz/\zeta + n - 1} \frac{1}{i} = \frac{1}{\tauz} + \frac{1}{\zeta}\sum_{i=\tauz/\zeta+1}^{\tauz/\zeta + n - 1} \frac{1}{i} 
\end{equation*}
and
\begin{equation*}
\frac{1}{\tauz} + \frac{1}{\zeta}\sum_{i=\tauz/\zeta+1}^{\tauz/\zeta + n - 1} \frac{1}{i} \leq \frac{1}{\tauz} + \frac{1}{\zeta}\int_{\tauz/\zeta}^{\tauz/\zeta + n - 1} \frac{1}{x} dx 
= \frac{1}{\tauz}+\frac{1}{\zeta}\log\Big(1+\frac{\zeta (n-1)}{\tauz}\Big).
\end{equation*}
Finally, we upper bound the preceding result by replacing $n-1$ with $n$ to simplify our analysis in the proof of Theorem~\ref{thm:regretbound}, which gives
\begin{equation*}
\sum_{i=1}^{n}\frac{1}{\tauz + \zeta (i-1)} \leq \frac{1}{\tauz}+\frac{1}{\zeta}\log\Big(1+\frac{\zeta n}{\tauz}\Big).
\end{equation*}
\end{proof}

%% file: Appendix_Sections/discountbound1.tex
\begin{proof}(\emph{Proof of Lemma~\ref{lem:discount1}}). 
Suppose $\alpha(k) =j \in [m]$ at epoch $k \in [n]$ and thus the epoch contains $\taukj$ iterations. Then, noting that $\mu_j=\sum_{\theta}r_j^\theta\pi_j(\theta)$, we have the following:
\begin{align*}
\Big|\mb{E}\Big[\sum_{\theta}r_j^\theta\pi_j(\theta)-\frac{1}{\bsg_k}\sum_{t=t_k}^{t_{k+1} - 1}(\gamma)^{t_{k+1} - 1- t}r_{j}^{\thetarv{t}}\big|\ft{j}{T_j^{\alpha}(k)-1}\Big]\Big|& \\
&\mkern-164mu\leq \frac{1}{\bsg_k}\sum_{t=t_k}^{t_{k+1} - 1}(\gamma)^{t_{k+1} - 1-t}\sum_{\theta}|\pi_j(\theta)-\beta_{t}(\theta)| \\
&\mkern-164mu= \frac{1}{\bsg_k}\sum_{t=t_k}^{t_{k+1} - 1}(\gamma)^{t_{k+1} - 1-t}\sum_{\theta}\big|\pi_j(\theta)-\sum_{\theta'}P_j^{t-t_k}(\theta',\theta)\beta_{t_k}(\theta')\big|\\
&\mkern-164mu= \frac{1}{\bsg_k}\sum_{t=t_k}^{t_{k+1} - 1}(\gamma)^{t_{k+1} - 1-t}\|\pi_j(\cdot)-\sum_{\theta'}P_j^{t-t_k}(\theta',\cdot)\beta_{t_k}(\theta')\|_{1}\\
&\mkern-164mu\leq \frac{C_j}{\bsg_k}\sum_{t=t_k}^{t_{k+1} - 1}(\gamma)^{t_{k+1} - 1-t}(\lambda_j)^{t-t_k}.
\end{align*}
The first inequality follows from the triangle inequality, the rewards being bounded in $[0, 1]$, and Fubini's theorem~\citep[Theorem 2.37]{folland:2007aa}. The second inequality is a direct application of Proposition~\ref{prop:convergence}. Now recall that we defined the following constants for arm $j$:
\begin{equation*}
\eta_j = \min\{\gamma, \lambda_j\}, \quad \phi_j = \max\{\gamma, \lambda_j\}, \quad \psi_j = \eta_j/\phi_j. 
\end{equation*}
This gives
\begin{equation*}
 \frac{C_j}{\bsg_k}\sum_{t=t_k}^{t_{k+1} - 1}(\gamma)^{t_{k+1} - 1-t}(\lambda_j)^{t-t_k} =\frac{C_j(\phi_j)^{\taukj-1}}{\bsg_k}\sum_{t=t_k}^{t_{k+1} - 1}(\psi_j)^{t-t_k}\label{eq:discount1}.
\end{equation*}
We can bound the equation overhead dependent on the discount factor $\gamma$.
\newline
\textbf{Case 1:} $\gamma \in (0, 1)$ and $\gamma \neq \lambda_j$.
\begin{equation*}
\frac{C_j(\phi_j)^{\taukj-1}}{\bsg_k}\sum_{t=t_k}^{t_{k+1} - 1}(\psi_j)^{t-t_k} =\frac{C_j(\phi_j)^{\taukj-1}(1 - (\psi_j)^{\taukj})}{\bsg_k(1-\psi_j)}.
\end{equation*}

\noindent
\textbf{Case 2:} $\gamma \in (0, 1)$ and $\gamma = \lambda_j$.
\begin{equation*}
\frac{C_j(\phi_j)^{\taukj-1}}{\bsg_k}\sum_{t=t_k}^{t_{k+1} - 1}(\psi_j)^{t-t_k} =\frac{C_j(\phi_j)^{\taukj-1}\taukj}{\bsg_k}.
\end{equation*}

\noindent
\textbf{Case 3:} $\gamma = 1$.
\begin{equation*}
\frac{C_j(\phi_j)^{\taukj-1}}{\bsg_k}\sum_{t=t_k}^{t_{k+1} - 1}(\psi_j)^{t-t_k} =\frac{C_j(1-(\lambda_j)^{\taukj})}{\bsg_k(1-\lambda)}.
\end{equation*}

\noindent
Combining each of the cases gives
\begin{equation*}
\big|\mb{E}\big[\mu_j-\br{k}{j}\big|\ft{j}{T_j^{\alpha}(k)-1}\big]\big|
\leq \frac{C_j\uj{j}{\taukj}}{\bsg_k},
\end{equation*}
where
$\uj{j}{\taukj}$ is defined as follows depending on the type of reward feedback:
\begin{enumerate}[topsep=0pt,itemsep=-2pt]
\item Discount-Averaged Reward Feedback: $\gamma\in(0,1)$.
\begin{equation*}
\uj{j}{\taukj} = 
\begin{cases} 
\frac{(\phi_j)^{\taukj-1}(1 - (\psi_j)^{\taukj})}{(1-\psi_j)}, & \text{if} \
\gamma \neq \lambda_j \\
(\phi_j)^{\taukj-1}\taukj, & \text{otherwise} 
\end{cases}.
\end{equation*}
\item Time-Averaged Reward Feedback: $\gamma=1$. 
\begin{equation*}
\textstyle\uj{j}{\taukj}=\frac{1-(\lambda_j)^{\taukj}}{1-\lambda_j}.
\end{equation*}
\end{enumerate}
\end{proof}

%% file: Appendix_Sections/discountbound2.tex
\begin{proof}(\emph{Proof of Lemma~\ref{lem:boundforAH}}).
We manipulate the expression and use the triangle inequality to convert the quantity to a sum of terms that can be bounded with Lemma~\ref{lem:discount1} as follows:
\begin{align}
\Big|\mu_j-\frac{1}{T_j}\sum_{i=1}^{T_j}\mb{E}[\car{i}{j}|\ft{j}{i-1}]\Big| &=\Big|\frac{1}{T_j}\sum_{i=1}^{T_j}(\mu_j-\mb{E}[\car{i}{j}|\ft{j}{i-1}])\Big| \nonumber\\
&=\Big|\frac{1}{T_j}\sum_{i=1}^{T_j}\mb{E}[\mu_j-\bs{r}_{j, k_j^i}^{\theta}|\ft{j}{i-1}]\Big|\nonumber \\
&\leq \frac{1}{T_j}\sum_{i=1}^{T_j}\big|\mb{E}[\mu_j-\bs{r}_{j, k_j^i}^{\theta}|\ft{j}{i-1}]\big|\nonumber\\
&\leq \frac{C_j}{T_j}\sum_{i=1}^{T_j} \frac{\uj{j}{\tau_j^i}}{\bsg_j^i},
\label{eq:discount2}  
\end{align}
where $k_j^i$ is the epoch in which arm $j$ is pulled for the $i$--th time and $\tau_j^i$ is number of iterations in that epoch. We now bound (\ref{eq:discount2}) for each function $\uj{j}{\tau_j^i}$ takes on dependent upon the discount factor $\gamma$.
\newline
\textbf{Case 1:} $\gamma \in (0, 1)$ and $\gamma \neq \lambda_j \ \forall \ j \in [m]$. 
\begin{align*}
\frac{C_j}{T_j}\sum_{i=1}^{T_j} \frac{\uj{j}{\tau_j^i}}{\bsg_j^i} 
&=\frac{C_j}{T_j}\sum_{i=1}^{T_j}\frac{(\phi_j)^{\tau_j^i-1}(1-(\psi_j)^{\tau_j^i})}{\bsg_j^i(1-\psi_j)} 
\leq \frac{C_j}{T_j}\Big(\frac{1}{1-\psi_j}\Big)\sum_{i=1}^{T_j}\frac{(\phi_j)^{\tau_j^i-1}}{\bsg_j^i} \\
&\leq \frac{C_j}{T_j}\Big(\frac{1}{1-\psi_j}\Big)\sum_{i=1}^{T_j}(\phi_j)^{\tau_j^i -1} 
= \frac{C_j}{T_j}\Big(\frac{1}{1-\psi_j}\Big)\sum_{i=1}^{T_j}(\phi_j)^{\tauz + \zeta(i-1)-1} \\
&= \frac{C_j}{T_j}\Big(\frac{(\phi_j)^{\tauz-1}}{1-\psi_j}\Big)\Big(\frac{1 - (\phi_j)^{\zeta T_j}}{1-(\phi_j)^\zeta}\Big) 
= \frac{C_j}{T_j}\Big(\frac{(\phi_j)^{\tauz}}{\phi_j-\eta_j}\Big)\Big(\frac{1 - (\phi_j)^{\zeta T_j}}{1-(\phi_j)^\zeta}\Big).
\end{align*}

\noindent
\textbf{Case 2:} $\gamma \in (0, 1)$ and $\gamma = \lambda_j$ for some $j \in [m]$.
\begin{align*}
\frac{C_j}{T_j}\sum_{i=1}^{T_j} \frac{\uj{j}{\tau_j^i}}{\bsg_j^i} &=  \frac{C_j}{T_j}\sum_{i=1}^{T_j} \frac{\tau_j^i(\phi_j)^{\tau_j^i-1}}{\bsg_j^i}  
\leq \frac{C_j}{T_j}\sum_{i=1}^{T_j}(\phi_j)^{\tau_j^i-1}\tau_j^i \\
&= \frac{C_j}{T_j}\sum_{i=1}^{T_j} (\phi_j)^{\tauz + \zeta(i-1)-1}(\tauz + \zeta (i - 1)) \\
&= \frac{C_j(\phi_j)^{\tauz - 1}}{T_j}\sum_{i=1}^{T_j}(\phi_j)^{\zeta (i-1)}(\tauz + \zeta (i-1)) \\
&= \frac{C_j(\phi_j)^{\tauz - 1}}{T_j}\Big(\frac{\tauz - (\phi_j)^{\zeta T_j}(\tauz + \zeta T_j)}{1 -(\phi_j)^{\zeta}} + \frac{\zeta (\phi_j)^{\zeta}(1 - (\phi_j)^{\zeta T_j})}{(1 - (\phi_j)^{\zeta})^2}\Big). 
\end{align*}
The final equality follows from recognizing that $\sum_{i=1}^{T_j}(\phi_j)^{\zeta (i-1)}(\tauz + \zeta (i-1))$ is an arithmetico-geometric series and then substituting the expression for the finite sum. An arithmetico-geometric series has the following general form for constants $|r| < 1$ and $a,d\in \mathbb{R}$, which can equivalently be expressed in terms of the finite sum that is bounded by the infinite sum:
\begin{equation*}
\sum_{i=1}^{n} r^{i-1}(a + d(i-1)) = \frac{a - r^n(a + dn)}{1-r} + \frac{dr(1 - r^n)}{(1 - r)^2} \leq \frac{a}{1-r} + \frac{dr}{(1 - r)^2}.
\end{equation*}
\noindent
\textbf{Case 3:} $\gamma =1$.
\begin{align*}
\frac{C_j}{T_j}\sum_{i=1}^{T_j}\frac{\uj{j}{\tau_j^i}}{\bsg_j^i} &= \frac{C_j}{T_j}\sum_{i=1}^{T_j}\frac{(1-(\lambda_j)^{\tau_j^i})}{\bsg_j^i(1-\lambda_j)} 
\leq \frac{C_j}{T_j}\Big(\frac{1}{1-\lambda_j}\Big)\sum_{i=1}^{T_j}\frac{1}{\bsg_j^i} \\
&= \frac{C_j}{T_j}\Big(\frac{1}{1-\lambda_j}\Big)\sum_{i=1}^{T_j}\frac{1}{\tau_0 + \zeta (i-1)}  \\
&\leq \frac{C_j}{T_j}\Big(\frac{1}{1-\lambda_j}\Big)\Big(\frac{1}{\tauz} + \frac{1}{\zeta}\log\Big(1 + \frac{\zeta T_j}{\tauz}\Big)\Big). 
\end{align*}
The final inequality is a direct application of harmonic bound from Section~\ref{app:harmonic}.

Combining each of the cases gives 
\begin{equation*}
\Big|\mb{E}\Big[\mu_j-\frac{1}{T_j}\sum_{i=1}^{T_j}\mb{E}[\car{i}{j}|\ft{j}{i-1}]\Big]\Big| \leq \frac{\lj{j}{T_j}}{T_j},  
\end{equation*}
where $\lj{j}{T_j}$ is defined as follows depending on the type of reward feedback:
\begin{enumerate}[topsep=0pt,itemsep=-2pt]
\item Discount-Averaged Reward Feedback: $\gamma\in(0,1)$.
\begin{equation*}
\lj{j}{T_j} = 
\begin{cases}
 C_j\big(\frac{(\phi_j)^{\tauz}}{\phi_j-\eta_j}\big)\big(\frac{1 - (\phi_j)^{\zeta T_j}}{1-(\phi_j)^\zeta}\big), & \text{if} \ \gamma \neq \lambda_j \  \forall \ j \in [m]\\
 C_j(\phi_j)^{\tauz - 1}\big(\frac{\tauz - (\phi_j)^{\zeta T_j}(\tauz + \zeta T_j)}{1 -(\phi_j)^{\zeta}} + \frac{\zeta (\phi_j)^{\zeta}(1 - (\phi_j)^{\zeta T_j})}{(1 - (\phi_j)^{\zeta})^2}\big), & \text{otherwise} 
\end{cases}.
\end{equation*}
\item Time-Averaged Reward Feedback: $\gamma=1$.
\setlength{\belowdisplayskip}{0pt}\setlength{\belowdisplayshortskip}{0pt}
\begin{equation*}
\textstyle \lj{j}{n}=\frac{C_j}{1-\lambda_j}\big(\frac{1}{\tauz}+\frac{1}{\zeta}\log\big(1+\frac{\zeta n}{\tauz} \big)\big).
\end{equation*}
\end{enumerate}
\end{proof}

%% file: Appendix_Sections/ucb_proof.tex
\begin{proof}(\emph{Proof of Theorem~\ref{thm:regretbound}}.)
The EpochUCB policy plays the arm with the maximum upper confidence bound on the empirical mean reward at each epoch. The policy at an epoch $k \in [n]$ with $\alpha$ taken as the EpochUCB algorithm is then
\begin{equation*}
\alpha(k) = \arg \max_{j\in[m]} \barR{T_j^{\alpha}(k-1)}{j} + \cf{k}{T_j^{\alpha}(k-1)}{j}.
\end{equation*}
Recall that we use the notation 
\begin{equation*}
\barR{T_j^{\alpha}(k-1)}{j}=\frac{1}{T_j^{\alpha}(k-1)}\sum_{i=1}^{T_j^{\alpha}(k-1)} \car{i}{j}
\end{equation*}
and 
\begin{equation*}
\cf{k}{T_j^{\alpha}(k-1)}{j}=\frac{\lj{j}{T_j^{\alpha}(k-1)}}{T_j^{\alpha}(k-1)}+\sqrt{\frac{6\log(k)}{T_j^{\alpha}(k-1)}}
\end{equation*}
to respectively denote the empirical mean reward and confidence window at epoch $k \in [n]$ for arm $j\in [m]$ when it has been pulled $T_j^{\alpha}(k-1)$ times prior to epoch $k$.

Given the above notation, we upper bound $T_j^{\alpha}(n)$ for each arm $j \in [m]$. Note that we replace the arm index $j$ with $\ast$ when referencing the optimal arm. For an arbitrary positive integer $\ell$, we have that
\begin{align*} 
&T_j^{\alpha}(n)= 1+\sum_{k=m+1}^nI\{\alpha(k)=j\}\\
&\leq \ell+\sum_{k=m+1}^nI\{\alpha(k)=j, T_j^{\alpha}(k-1)\geq \ell\}\\
&\leq \ell+\sum_{k=m+1}^nI\{\barR{T_\ast^{\alpha}(k-1)}{\ast}+\cf{k}{T_\ast^{\alpha}(k-1)}{\ast}\leq \barR{T_j^{\alpha}(k-1)}{j}+\cf{k}{T_j^{\alpha}(k-1)}{j},T_j^{\alpha}(k-1)\geq \ell\}\notag\\
&\leq \ell +\sum_{k=m+1}^nI\big\{\min_{0<s<k} \barR{s}{\ast}+\cf{k}{s}{\ast}
\leq \max_{\ell\leq w_j<k}\barR{w_j}{j}+\cf{k}{w_j}{j}\big\}\\
&\leq \ell+\sum_{k=1}^n\sum_{s=1}^{k-1}\sum_{w_j=\ell}^{k-1}I\{\barR{s}{\ast}+\cf{k}{s}{\ast}\leq \barR{w_j}{j}+\cf{k}{w_j}{j}\}.
\label{eq:bounds}
\end{align*}
Now, suppose that all three of the following are false: 
\begin{numcases}{} 
\label{eq:cond1}
\barR{s}{\ast} &$\leq\mu_\ast -\cf{k}{s}{\ast}$\\
\label{eq:cond2}
\barR{w_j}{j} &$\geq \mu_j+\cf{k}{w_j}{j}$\\
\label{eq:cond3}  
\mu_\ast &$<\mu_j+2\cf{k}{w_j}{j}$
\end{numcases}
Then, 
\begin{equation*}
\barR{s}{\ast}+\cf{k}{s}{\ast}>\mu_\ast\geq
\mu_j+2\cf{k}{w_j}{j}>\barR{w_j}{j}+\cf{k}{w_j}{j}.
\end{equation*}
Hence, if $\barR{s}{\ast}+\cf{k}{s}{\ast}\leq \barR{w_j}{j}+\cf{k}{w_j}{j}$, then at least one of \eqref{eq:cond1}--\eqref{eq:cond3} holds. We bound the probability of events \eqref{eq:cond1} and \eqref{eq:cond2} using the Azuma-Hoeffding inequality given in Proposition~\ref{prop:AH} and find a positive integer $\ell$ such that \eqref{eq:cond3} is always false.
        
Toward this end, suppose an arm $j$ has been played $T_j$ times. We apply Proposition~\ref{prop:AH} to the martingale $(Y_{j,T_j})_{T_j\in \mb{Z}_+}$. Note that by the law of conditional expectations,
$\mb{E}[Y_{j,T_j}]=0$ so that Proposition~\ref{prop:AH} implies that for each arm $j \in [m]$ and any $\epsilon>0$, $P(Y_{j,T_j}\leq -\epsilon)\leq \exp(-\epsilon^2/(2T_j) )$.
We need to relate the random variable $Y_{j,T_j}$ to the difference between the empirical mean reward and the expected stationary distribution reward for each arm so that we can bound this difference. Consider the event $\omega = \{\mu_j -\barR{T_j}{j}\geq \epsilon\}$, and equivalently:
\begin{align*}
\omega&=\Big\{\mu_j-\frac{1}{T_j}\sum_{i=1}^{T_j}
\mb{E}[\car{i}{j}|\ft{j}{i-1}]+\frac{1}{T_j}\sum_{i=1}^{T_j}
\mb{E}[\car{i}{j}|\ft{j}{i-1}]-\barR{T_j}{j}\geq \epsilon \Big\}\\
&=\Big\{\mu_j- \frac{1}{T_j}\sum_{i=1}^{T_j}
\mb{E}[\car{i}{j}|\ft{j}{i-1}]-\frac{Y_{j,T_j}}{T_j}\geq \epsilon\Big\}.
\end{align*}
This representation follows from adding and subtracting the random variable given by $\frac{1}{T_j}\sum_{i=1}^{T_j}\mb{E}[\car{i}{j}|\ft{j}{i-1}]$ into the event $\omega$ and the definition of $Y_{j,T_j}$ given in~\eqref{eq:yk}. By Lemma~\ref{lem:boundforAH}, 
\begin{equation*}
\omega\subset\Big\{ \frac{\lj{j}{T_j}}{T_j}-\frac{Y_{j,T_j}}{T_j}\geq \epsilon \Big\}=\Big\{ \frac{Y_{j,T_j}}{T_j}\leq \frac{\lj{j}{T_j}}{T_j}-\epsilon \Big\}.
\end{equation*}
Hence, applying Proposition~\ref{prop:AH},
\begin{equation*}
P(\mu_j- \barR{T_j}{j}\geq \epsilon)\leq P\Big(\frac{Y_{j,T_j}}{T_j}\leq \frac{\lj{j}{T_j}}{T_j}-\epsilon\Big)\leq 
\exp\Big(-\frac{T_j}{2}\Big(\epsilon-\frac{\lj{j}{T_j}}{T_j} \Big)^2\Big).
\end{equation*}
Defining 
\begin{equation*}
\epsilon=\sqrt{\frac{2}{T_j}\log\Big(\frac{1}{\delta}\Big)}+\frac{\lj{j}{T_j}}{T_j},
\end{equation*}
we have, for any fixed $\delta>0$,
\begin{equation}
P\Big(\mu_j-\barR{T_j}{j}\geq \sqrt{\frac{2}{T_j}\log\Big(\frac{1}{\delta}\Big)}+\frac{\lj{j}{T_j}}{T_j}\Big)\leq \delta.
\label{eq:failure_prob}
\end{equation}
Taking $\delta=k^{-3}$, and relating~\eqref{eq:failure_prob} to \eqref{eq:cond1} and~\eqref{eq:cond2}, we get that 
\begin{equation*}
P(\barR{s}{\ast}\leq \mu_\ast-\cf{t}{s}{\ast})\leq k^{-3}
\end{equation*}
and 
\begin{equation*}
P(\barR{w_j}{j}\geq \mu_j+\cf{t}{w_j}{\ast})\leq k^{-3}.
\end{equation*}
This implies that \eqref{eq:cond1} and \eqref{eq:cond2} hold with high probability.

Now, we choose $\ell$ to be the smallest integer such that \eqref{eq:cond3} is always false---that is, choose it such that
\begin{equation}
\mu_\ast-\mu_j-2\cf{k}{w_j}{j}>\mu_\ast-\mu_j-2\Big(\frac{\lj{j}{\ell}}{\ell}+\sqrt{\frac{6\log(k)}{\ell}} \Big)>0.
\label{eq:condition}
\end{equation}
We find $\ell$ for each function that $\lj{j}{\ell}$ can take on depending on the discount factor $\gamma$.
\newline
\newline
\textbf{Case 1:} $\gamma \in (0, 1)$ and $\gamma \neq \lambda_j \ \forall \ j \in [m]$.
 \newline
 \newline
Plugging $\lj{j}{\ell}$ into~\eqref{eq:condition}, we need to find $\ell$ to satisfy the following:
\begin{equation}
\Delta_j-2\Big(\frac{C_j}{\ell}\Big(\frac{(\phi_j)^{\tauz}}{\phi_j - \eta_j}\Big)\Big(\frac{1-(\phi_j)^{\zeta \ell}}{1-(\phi_j)^{\zeta}}\Big)+\sqrt{\frac{6\log(k)}{\ell}}\Big)>0.
\label{eq:condition1}
\end{equation}
Since $(1 - (\phi_j)^{\zeta \ell})(1-(\phi_j)^{\zeta})^{-1} \leq (1-(\phi_j)^{\zeta})^{-1}$ and $1/x<1/\sqrt{x}$ on $[1,\infty)$, we have 
\begin{equation*}
\frac{1}{\ell}\Big(\frac{(\phi_j)^{\tauz}}{\phi_j - \eta_j}\Big)\Big(\frac{1-(\phi_j)^{\zeta \ell}}{1-(\phi_j)^{\zeta}}\Big) \leq \frac{1}{\sqrt{\ell}}\Big(\frac{(\phi_j)^{\tauz}}{\phi_j - (\eta_j)^{\zeta}}\Big)\Big(\frac{1}{1-(\phi_j)^{\zeta}}\Big).
\end{equation*}
Thus,~\eqref{eq:condition1} reduces to finding the smallest integer $\ell$ such that
\begin{equation*}
\Delta_j-2\Big(\frac{C_j}{\sqrt{\ell}}\Big(\frac{(\phi_j)^{\tauz}}{\phi_j - \eta_j}\Big)\Big(\frac{1}{1-(\phi_j)^{\zeta}}\Big)+\frac{\sqrt{6\log(k)}}{\sqrt{\ell}} \Big)>0.
\end{equation*}
Rearranging and squaring terms, we get the following condition:
\begin{equation*}
\ell>\frac{4}{\Delta_j^2}\Big(C_j\Big(\frac{(\phi_j)^{\tauz}}{\phi_j - \eta_j}\Big)\Big(\frac{1}{1-(\phi_j)^{\zeta}}\Big)+\sqrt{6\log(k)} \Big)^2.
\end{equation*}
\textbf{Case 2:} $\gamma \in (0,1)$ and $\gamma = \lambda_j$ for some $j \in [m]$.
\newline
\newline
Plugging $\lj{j}{\ell}$ into~\eqref{eq:condition}, we need to find $\ell$ to satisfy 
\begin{equation}
\Delta_j-2\Big(\frac{C_j(\phi_j)^{\tau_0-1}}{\ell}\Big(\underbrace{\frac{\tauz - (\phi_j)^{\zeta \ell}(\tauz + \zeta \ell)}{1 - (\phi_j)^{\zeta}}+\frac{\zeta(\phi_j)^{\zeta}(1 - (\phi_j)^{\zeta \ell})}{(1 - (\phi_j)^{\zeta})^2}}_{=S_{\ell}}\Big)+\sqrt{\frac{6\log(k)}{\ell}}\Big)>0,
\label{eq:condition2}
\end{equation}
where $S_{\ell}$ is the finite sum of an arithmetico-geometric series. Since the finite sum of a series is upper bounded by the infinite sum, implying $S_{\ell} \leq S_{\infty}$, and $1/x<1/\sqrt{x}$ on $[1,\infty)$, we have 
\begin{equation*}
\Big(\frac{1}{\ell}\Big(\frac{\tauz - (\phi_j)^{\zeta \ell}(\tauz + \zeta \ell)}{1 - (\phi_j)^{\zeta}}+\frac{\zeta(\phi_j)^{\zeta}(1 - (\phi_j)^{\zeta \ell})}{(1 - (\phi_j)^{\zeta})^2}\Big)\Big)< \Big(\frac{1}{\sqrt{\ell}}\Big(\frac{\tauz }{1 - (\phi_j)^{\zeta}}+\frac{\zeta(\phi_j)^{\zeta}}{(1 - (\phi_j)^{\zeta})^2}\Big)\Big).
\end{equation*}
Consequently, satisfying Equation~\eqref{eq:condition2} reduces to finding the smallest integer $\ell$ such that
\begin{equation*}
\Delta_j-2\Big(\frac{C_j(\phi_j)^{\tauz-1}}{\sqrt{\ell}}\Big(\frac{\tauz }{1 - (\phi_j)^{\zeta}}+\frac{\zeta(\phi_j)^{\zeta}}{(1 - (\phi_j)^{\zeta})^2}\Big)+\frac{\sqrt{6\log(k)}}{\sqrt{\ell}} \Big)>0.
\end{equation*}
Rearranging and squaring terms, we get the following condition:
\begin{equation*}
\ell>\frac{4}{\Delta_j^2}\Big(C_j(\phi_j)^{\tauz-1}\Big(\frac{\tauz }{1 - (\phi_j)^{\zeta}}+\frac{\zeta(\phi_j)^{\zeta}}{(1 - (\phi_j)^{\zeta})^2}\Big)+\sqrt{6\log(k)} \Big)^2.
\end{equation*}
\textbf{Case 3:} $\gamma = 1$.
\newline
\newline
Plugging $\lj{j}{\ell}$ into~\eqref{eq:condition}, we need to find $\ell$ to satisfy the following equation:
\begin{equation}
    \Delta_j-2\Big(
\frac{C_j}{\ell (1-\lambda_j)}\Big(\frac{1}{\tauz}+\frac{1}{\zeta}\log\Big(
    1+\frac{\ell\zeta}{\tauz} \Big) \Big)+\sqrt{\frac{6\log(k)}{\ell}}
\Big)>0.
\label{eq:condition3}
\end{equation}
Let $\tilde{\ell}=\ell\zeta/\tauz$, so that the condition needing to be satisfied in~\eqref{eq:condition3} is equivalently expressed as 
\begin{equation*}
\Delta_j-2\Big(\frac{C_j}{\tauz(1-\lambda_j)}\Big(\frac{1}{\tilde{\ell}}\frac{\zeta}{\tauz}+\frac{1}{\tilde{\ell}}\log(1+\tilde{\ell}) \Big)+\sqrt{\frac{6\log(k)}{\ell}}\Big)>0.
\end{equation*}
Since $1/x<1/\sqrt{x}$ and $(1/x)\log(1+x)<1/\sqrt{x}$ on $[1,\infty)$, we have the inequality
\begin{equation*}
\frac{1}{\tilde{\ell}}\frac{\zeta}{\tauz}+\frac{1}{\tilde{\ell}}\log(1+\tilde{\ell})<\frac{1}{\sqrt{\tilde{\ell}}}\frac{\zeta}{\tauz}+\frac{1}{\sqrt{\tilde{\ell}}}.
\end{equation*}
Hence, an $\ell$ such that the following equation holds will satisfy~\eqref{eq:condition3}:
\begin{equation*}
\Delta_j-2\Big( \frac{C_j}{\tauz (1-\lambda_j)}\Big(\frac{\sqrt{\tauz}}{\sqrt{\ell\zeta}}\frac{\zeta}{\tauz}+\frac{\sqrt{\tauz}}{\sqrt{\ell\zeta}}
\Big)+\frac{\sqrt{6\log(k)}}{\sqrt{\ell}} \Big)>0.
\end{equation*}
Rearranging and squaring terms, we get the following condition:
\begin{equation*} 
\ell>\frac{4}{\Delta_j^2}\Big(\frac{C_j}{\sqrt{\zeta \tauz}(1-\lambda_j)}\Big(1+ \frac{\zeta}{\tauz}\Big)+\sqrt{6\log(k)} \Big)^2.
\end{equation*}
Combining each of the cases, \eqref{eq:cond3} is false for
\begin{equation*}
\ell=\Big\lceil \frac{4}{\Delta_j^2}\big(\rj{j}
+\sqrt{6\log(n)} \big)^2\Big\rceil,
\end{equation*}
and for all 
\begin{equation*}
w_j\geq \frac{4}{\Delta_j^2}\big(\rj{j}
+\sqrt{6\log(n)} \big)^2,
\end{equation*}
where the constant $\rj{j}$ is defined as follows depending on the type of reward feedback:
\begin{enumerate}[topsep=0pt,itemsep=-2pt]
\item Discount-Averaged Reward Feedback: $\gamma \in (0, 1)$.
\begin{equation*}
\rj{j} = 
\begin{cases}
C_j\big(\frac{(\phi_j)^{\tauz}}{\phi_j - \eta_j}\big)\big(\frac{1}{1-(\phi_j)^{\zeta}}\big), & \text{if} \ \gamma \neq \lambda_j \ \forall \ j \in [m] \\
C_j(\phi_j)^{\tauz-1}\big(\frac{\tauz }{1 - (\phi_j)^{\zeta}}+\frac{\zeta(\phi_j)^{\zeta}}{(1 - (\phi_j)^{\zeta})^2}\big), & \text{otherwise}
\end{cases}.
\end{equation*}
\item Time-Averaged Reward Feedback: $\gamma=1$.
\begin{equation*}
\textstyle \rj{j}=\frac{C_j}{\sqrt{\zeta \tauz}(1-\lambda_j)}\big(1 + \frac{\zeta}{\tauz} \big).
\end{equation*}
\end{enumerate}
Hence, 
\begin{align*}
    \mb{E}_{\alpha}[T_j^{\alpha}(n)]&\leq \ell +
    \sum_{k=1}^n\sum_{s=1}^{k-1}\sum_{w_j=\ell}^{k-1}\big(
    P(\barR{s}{\ast}\leq \mu_\ast-\cf{k}{s}{\ast}) +P(\barR{w_j}{j}\geq
    \mu_j+\cf{k}{w_j}{j})\big)\\
    &\leq \Big\lceil \frac{4}{\Delta_j^2}\big(
    \rj{j}+\sqrt{6\log(n)} \big)^2\Big\rceil+
       \sum_{k=1}^n\sum_{s=1}^{k}\sum_{w_j=1}^{k}2k^{-3}\\
       &\leq \frac{4}{\Delta_j^2}\big(
    \rj{j}+\sqrt{6\log(n)} \big)^2+3+2\log(n).
    \label{eq:Tbound}
\end{align*}
Note that the final inequality follows from~\eqref{eq:sum_inequality}.
\end{proof}

%% file: Appendix_Sections/gap_independent.tex
\begin{proof}(\emph{Proof of Corollary~\ref{corr:gap_independent}}). The regret decomposition given in Proposition~\ref{thm:regretdecomp1} says that the regret of any algorithm $\alpha$ with epoch length sequence $\{\taukr\}_{k=1}^n$ as defined in~\eqref{eq:tauk} is bounded as follows:
\begin{equation*}
R^\alpha(n) \leq \sum_{j\neq j_\ast}\mb{E}_\alpha[ T_j^\alpha(n)]\Delta_j + \sum_{j\in[m]}\lj{j}{n}.
\end{equation*}
To obtain a gap-independent regret bound for EpochUCB, it is sufficient to derive a gap-independent bound on the term $\sum_{j\neq j_\ast}\mb{E}_\alpha[T_j^\alpha(n)]\Delta_j$ since $\sum_{j\in[m]}\lj{j}{n}$ has no dependence on the reward gaps. We derive such a bound using the Cauchy-Schwarz inequality, the upper bound on the number of times a suboptimal arm is played by EpochUCB given in Theorem~\ref{thm:regretbound}, and the fact that $\Delta_j \in [0, 1]$. Indeed, 
\begin{align*}
\sum_{j\neq j_\ast}\mb{E}_\alpha[T_j^\alpha(n)]\Delta_j &= \sum_{j\in [m]}\mb{E}_\alpha[T_j^\alpha(n)]\Delta_j = \sum_{j\in [m]}\big(\sqrt{\mb{E}_\alpha[T_j^\alpha(n)]}\big)\Big(\sqrt{\mb{E}_\alpha[T_j^\alpha(n)}]\Delta_j\Big) \\
&\leq \sqrt{\Big(\sum_{j\in [m]}\mb{E}_\alpha[T_j^\alpha(n)]\Big)\Big(\sum_{j\in [m]}\mb{E}_\alpha[T_j^\alpha(n)]\Delta_j^2\Big)} \\
&= \sqrt{n\sum_{j\in [m]}\mb{E}_\alpha[T_j^\alpha(n)]\Delta_j^2} \\
&\leq \sqrt{n\sum_{j\in [m]}\Big(\frac{4}{\Delta_j^2}\big(\rj{j}+\sqrt{6\log(n)} \big)^2+3+2\log(n)\Big)\Delta_j^2 }\\
&= \sqrt{n\sum_{j\in [m]}\big(4\big(\rj{j}+\sqrt{6\log (n)} \big)^2+3\Delta_j^2 +2\Delta_j^2\log(n)\big)}\\
&\leq \sqrt{n\sum_{j\in [m]}\big(4\big(\rj{j}+\sqrt{6\log(n)} \big)^2+3+2\log(n)\big)}\\
&= \sqrt{n\sum_{j\in [m]}\big(4(\rj{j})^2+ 8\rj{j} \sqrt{6\log(n)} + 26\log (n)+3\big)}.
\end{align*}
Substituting the preceding result into Proposition~\ref{thm:regretdecomp1}, we have
\begin{equation*}
R^{\mathrm{EpochUCB}}(n) \leq \sqrt{n\sum_{j \in [m]}\big(4(\rj{j})^2+ 8\rj{j} \sqrt{6\log(n)} + 26\log (n)+3\big)} + \sum_{j\in[m]}\lj{j}{n}.
\end{equation*}
\end{proof}

%% file: Appendix_Sections/greedy_proof.tex
\begin{proof}(\emph{Proof of Theorem~\ref{thm:greedy}}).
The EpochGreedy policy plays the arm with maximum empirical mean reward with probability $1-\vep_k$ and an arm selected uniformly at random with probability $\vep_k$ at an epoch $k \in [n]$. Formally, the policy at epoch $k \in [n]$ with $\alpha$ taken as the EpochGreedy algorithm is 
\begin{equation*}
\alpha(k) = 
\begin{cases}
\arg \max_{j\in[m]} \barR{T_j^{\alpha}(k-1)}{j}, & \text{w.p.}~1-\vep_k \\
j, & \text{w.p.}~\frac{\vep_k}{m} \ \forall \ j \in [m] \\
\end{cases}.
\end{equation*}

Define $\Delta_{\text{min}}=\min_{j\neq j_\ast}\Delta_j$ and fix constants $c$
and $d$ such that $0 \leq d \leq \Delta_{\min}$ and $c>c'\nu^2$ where $c' > 8$, $\nu\geq \max\{\kappa, \frac{d}{\sqrt{c'}}\}$, and $\kappa = \min\{\kappa > 0: \kappa\sqrt{i} \geq \lj{j}{i} \ \forall \ i \in [n], j \in [m]\}$. Recall that $\lj{j}{\cdot}$ depends on the Markov chain statistics ($C_j, \lambda_j$) and is defined in~\eqref{eq:lj1} for discount-averaged reward feedback and in~\eqref{eq:lj12} for time-averaged reward feedback.
Let $\{\vep_k\}_{k=1}^n$ be a sequence with $\vep_k=\min\{1, \frac{cm}{d^2k}\}$ for each epoch $k \in [n]$ so that for $k \geq \lceil\frac{cm}{d^2}\rceil$, $\vep_k = \frac{cm}{d^2k}$. Moreover, let $x_0 = \frac{1}{2}(\sum_{i=1}^k\frac{\vep_i}{m})$, and observe that the value in the parenthesis is the expected number of times an arm will be pulled from random exploration. Given the above notation, our objective is to upper bound the probability of a suboptimal arm $j \in [m]$ being selected at an epoch $k \geq \lceil\frac{cm}{d^2}\rceil$. 

The probability that a suboptimal arm $j \in [m]$ is chosen at epoch $k \geq \lceil\frac{cm}{d^2}\rceil$ is given by
\begin{align*}
P(\alpha(k) = j) &= \frac{\vep_k}{m} + (1-\vep_k)
P\big(\barR{T_j^{\alpha}(k-1)}{j} \geq \barR{T_{\ast}^{\alpha}(k-1)}{\ast}\big), \notag 
\end{align*}
which we bound as
\begin{align}
P(\alpha(k) = j) &\leq \frac{c}{d^2k} + P\big(\barR{T_j^{\alpha}(k-1)}{j} \geq \barR{T_{\ast}^{\alpha}(k-1)}{\ast}\big).
\label{eq:prob_suboptimal}
\end{align}
For notational simplicity, we proceed bounding $P(\barR{T_j^{\alpha}(k)}{j} \geq \barR{T_{\ast}^{\alpha}(k)}{\ast})$ and obtain a bound on $P(\barR{T_j^{\alpha}(k-1)}{j} \geq \barR{T_{\ast}^{\alpha}(k-1)}{\ast})$ from merely swapping the epoch index in our final result. From a union bound over the events that result in the event $\{\barR{T_j^{\alpha}(k)}{j} \geq \barR{T_{\ast}^{\alpha}(k)}{\ast}\}$, we can bound $P(\barR{T_j^{\alpha}(k)}{j} \geq \barR{T_{\ast}^{\alpha}(k)}{\ast})$. Indeed,
\begin{align}
P\big(\barR{T_j^{\alpha}(k)}{j} \geq \barR{T_{\ast}^{\alpha}(k)}{\ast}\big) \leq P\Big(\barR{T_j^{\alpha}(k)}{j} \geq \mu_j + \frac{\Delta_j}{2}\Big) + P\Big(\barR{T_{\ast}^{\alpha}(k)}{\ast} \leq \mu_\ast - \frac{\Delta_j}{2}\Big).
\label{eq:union_split}
\end{align}
We expand $P(\barR{T_j^{\alpha}(k)}{j} \geq \mu_j + \frac{\Delta_j}{2})$ using the conditional probability to factor out the randomness of $T_j^{\alpha}(k)$ from $\barR{T_j^{\alpha}(k)}{j}$ as follows:
\begin{align}
\mkern-13muP\Big(\barR{T_j^{\alpha}(k)}{j} \geq \mu_j + \frac{\Delta_j}{2}\Big) 
&=  \sum_{i=1}^{k} P\Big(T_j^{\alpha}(k)=i, \barR{i}{j} \geq \mu_j + \frac{\Delta_j}{2}\Big) \notag \\
&= \sum_{i=1}^{k} P\Big(T_j^{\alpha}(k)=i \big| \barR{i}{j} \geq
\mu_j + \frac{\Delta_j}{2}\Big) P\Big(\barR{i}{j} \geq \mu_j +
\frac{\Delta_j}{2}\Big) \notag \\
&= \sum_{i=1}^{\lfloor x_0 \rfloor} P\Big(T_j^{\alpha}(k)=i \Big|
\barR{i}{j} \geq \mu_j + \frac{\Delta_j}{2}\Big) P\Big(\barR{i}{j}
\geq \mu_j + \frac{\Delta_j}{2}\Big) \notag \\
&+ \sum_{i=\lfloor x_0 \rfloor+1}^{k} P\Big(T_j^{\alpha}(k)=i \Big| \barR{i}{j} \geq \mu_j + \frac{\Delta_j}{2}\Big) P\Big(\barR{i}{j} \geq \mu_j + \frac{\Delta_j}{2}\Big).
\label{eq:todrop}
\end{align} 

Now, consider the event $\omega = \{\barR{i}{j} -\mu_j \geq \frac{\Delta_j}{2}\}$, and equivalently:
\begin{align*}
\omega &=\Big\{\barR{i}{j}-\frac{1}{i}\sum_{l=1}^i\mb{E}[\car{l}{j}|\ft{j}{l-1}]+\frac{1}{i}\sum_{l=1}^i\mb{E}[\car{l}{j}|\ft{j}{l-1}] - \mu_j\geq\frac{\Delta_j}{2}\Big\} \\
&=\Big\{\frac{Y_{j, i}}{i}+\frac{1}{i}\sum_{l=1}^i\mb{E}[\car{l}{j}|\ft{j}{l-1}]- \mu_j\geq\frac{\Delta_j}{2}\Big\}.
\end{align*}
This representation follows from adding and subtracting $\frac{1}{i}\sum_{l=1}^i\mb{E}[\car{l}{j}|\ft{j}{l-1}]$ into the event $\omega$ and the definition of $Y_{j, i}$ from~\eqref{eq:yk}. From Lemma~\ref{lem:boundforAH}, we obtain
\begin{align}
\omega \subset \Big\{\frac{Y_{j, i}}{i}+\frac{\lj{j}{i}}{i}\geq\frac{\Delta_j}{2}\Big\}. 
\label{eq:omega}
\end{align}
Equation~\ref{eq:omega} holds for discount-averaged and time-averaged reward feedback. Fundamentally, the type of reward feedback impacts the final result as a consequence of the lower bound on the constant $c$, which derives from the constant $\kappa$ that depends on the functional form $\lj{j}{\cdot}$ adopts as a consequence of the type of reward feedback.

For a positive integer $k\geq k'=\frac{cm}{d^2}$ and $\vep_k=\frac{cm}{d^2 k}$, we have that 
\begin{align*}
x_0=\frac{1}{2m}\sum_{i=1}^k\vep_i = \frac{1}{2m}\Big(\sum_{i=1}^{\lfloor k'\rfloor}\vep_i+\sum_{i=\lceil k'\rceil}^k \vep_i\Big)\geq
\frac{c}{2d^2}+\frac{c}{d^2}\log\Big(\frac{k}{k'}\Big)\geq \frac{c}{2d^2}.
\end{align*}
We also conclude that
\begin{equation}
x_0\geq \frac{c}{d^2}\log\Big(\frac{d^2k\exp(1/2)}{cm}\Big).
\label{eq:x0_lower_bound}
\end{equation}
Moreover, $\frac{c}{2d^2}\geq \frac{c'\nu^2}{2\Delta_{\min}^2}$ since by construction $c \geq c'\nu^2$ and $0 \leq d \leq \Delta_{\min}$, so that
$x_0\geq \frac{c'\nu^2}{2\Delta_{\min}^2}$. This implies
\[\frac{\Delta_j}{\sqrt{c'/2}}\geq \frac{\Delta_{\min}}{\sqrt{c'/2}}\geq \frac{\nu}{\sqrt{x_0}}\geq \frac{\lj{j}{x_0}}{x_0}.\]
Hence, $\frac{\Delta_j}{\sqrt{c'/2}}\geq \frac{\lj{j}{i}}{i}$ for all $i\geq \lfloor x_0\rfloor +1$ so that, recalling \eqref{eq:omega},
\[\omega\subset \Big\{\frac{Y_{j,i}}{i}+\frac{\lj{j}{i}}{i}\geq \frac{\Delta_j}{2} \Big\}\subset \Big\{\frac{Y_{j,i}}{i}\geq \frac{\Delta_j}{2}-\frac{\Delta_j}{\sqrt{c'/2}} \Big\}=\Big\{\frac{Y_{j,i}}{i}\geq
\frac{\Delta_j(\sqrt{c'/2}-2)}{\sqrt{2c'}}\Big\}.\]
Now, we apply the Azuma-Hoeffding inequality from Proposition~\ref{prop:AH} to $\omega$ to get that
\begin{align}
P\Big(\barR{i}{j} \geq \mu_j + \frac{\Delta_j}{2} \Big) \leq P\Big(\frac{Y_{j,i}}{i}\geq
\frac{\Delta_j(\sqrt{c'/2}-2)}{\sqrt{2c'}}\Big) \leq \exp\Big(-\frac{i}{2}\Big(\frac{\Delta_j(\sqrt{c'/2}-2)}{\sqrt{2c'}} \Big)^2\Big).
\label{eq:azuma_greedy}
\end{align}
Thus,
\begin{align}
\mkern-4mu P\Big(\barR{T_j^{\alpha}(k)}{j} \geq \mu_j + \frac{\Delta_j}{2}\Big) &\leq
\sum_{i=1}^{\lfloor x_0 \rfloor} P\Big(T_j^{\alpha}(k)= i \Big| \barR{i}{j} \geq \mu_j + \frac{\Delta_j}{2}\Big)+\sum_{i=\lfloor x_0\rfloor+1}^k  P\Big(\barR{i}{j} \geq \mu_j + \frac{\Delta_j}{2}\Big) \label{eq:drop} \\
&\mkern-150mu\leq \sum_{i=1}^{\lfloor x_0 \rfloor} P\Big(T_j^{\alpha}(k)= i \Big| \barR{i}{j} \geq \mu_j + \frac{\Delta_j}{2}\Big)+\sum_{i=\lfloor x_0\rfloor+1}^k  \exp\Big(-\frac{i}{2}\Big(\frac{\Delta_j(\sqrt{c'/2}-2)}{\sqrt{2c'}} \Big)^2\Big) \label{eq:apply_azuma} \\
&\mkern-150mu\leq \sum_{i=1}^{\lfloor x_0 \rfloor} P\Big(T_j^{\alpha}(k)= i \Big|\barR{i}{j} \geq \mu_j + \frac{\Delta_j}{2}\Big)+\frac{c''}{\Delta_j^2}\exp\Big( -\frac{\Delta_j^2\lfloor x_0\rfloor}{c''} \Big), \label{eq:exp_inequality}
\end{align}
where in~\eqref{eq:drop} we upper bound the terms that were dropped from~\eqref{eq:todrop} by one, in~\eqref{eq:apply_azuma} we apply the inequality of~\eqref{eq:azuma_greedy}, and in~\eqref{eq:exp_inequality} we define $c'' = 4c'(\sqrt{c'/2}-2)^{-2}$ and employ the inequality 
\begin{equation*}
\sum_{i=\lfloor x_0 \rfloor+1}^n \exp\Big(-\frac{i\Delta_j^2}{c''}\Big) \leq \sum_{i=\lfloor x_0 \rfloor+1}^\infty \exp\Big(-\frac{i\Delta_j^2}{c''}\Big) \leq \frac{c''}{\Delta_j^2}\exp\Big(-\frac{\Delta_j^2 \lfloor x_0 \rfloor}{c''}\Big).
\end{equation*} 
The choice of which term to keep in each sum of~\eqref{eq:drop} stems from the fact that the concentration bound on the empirical mean will be the dominating factor when $i \leq x_0$ and will tend toward zero for $i \geq \lfloor x_0 \rfloor + 1$. 

Defining $T_j^{\mathrm{R}}(k)$ to be the number of epochs arm $j$ has been chosen from random exploration by the EpochGreedy algorithm in $k$ epochs, we have
\begin{align*}
 P\Big(\barR{T_j^{\alpha}(k)}{j} \geq \mu_j + \frac{\Delta_j}{2}\Big) &\leq
\sum_{i=1}^{\lfloor x_0 \rfloor} P\Big(T_j^{\alpha}(k)\leq i \Big|\barR{i}{j} \geq \mu_j + \frac{\Delta_j}{2}\Big)+\frac{c''}{\Delta_j^2}\exp\Big( -\frac{\Delta_j^2\lfloor x_0\rfloor}{c''} \Big)  \\
&\leq
\sum_{i=1}^{\lfloor x_0 \rfloor} P\Big(T_j^{\mathrm{R}}(k)\leq i \Big|\barR{i}{j} \geq \mu_j + \frac{\Delta_j}{2}\Big)+\frac{c''}{\Delta_j^2}\exp\Big( -\frac{\Delta_j^2\lfloor x_0\rfloor}{c''} \Big) \\
& \leq x_0P\Big(T_j^{\mathrm{R}}(k)\leq x_0 \Big)+\frac{c''}{\Delta_j^2}\exp\Big( -\frac{\Delta_j^2\lfloor x_0\rfloor}{c''} \Big).
\end{align*}
The final inequality follows from recognizing that epochs consisting of random exploration are independent of the empirical mean rewards. 
To determine a bound on $P(T_j^{\mathrm{R}}(k)\leq x_0)$ we need the following Bernstein inequality. 
\begin{proposition}[Bernstein Inequality~\citep{uspensky:1937aa}]
Let $Z_1,\dots, Z_n$ be independent random variables with range $[0, 1]$. Define $S_n = \sum_{i=1}^n X_i$ and $\sigma^2 = \text{Var}(S_n) = \sum_{i=1}^n\text{Var}(X_i)$. Then for all $\epsilon \geq 0$,
\setlength{\belowdisplayskip}{0pt}\setlength{\belowdisplayshortskip}{0pt}
\begin{equation*}
P(S_n -\mb{E}[S_n] \leq -\epsilon) \leq \exp\Big(-\frac{\epsilon^2}{2\sigma^2 + \epsilon}\Big).
\end{equation*}
\label{prop:bernstein}
\end{proposition}
We can express $T_j^{\mathrm{R}}(k)$ as a sum of indicator variables $I_{j,i}^{\mathrm{R}}$ for $i \in [k]$ given by $T_j^{\mathrm{R}}(k) = \sum_{i=1}^k I_{j, i}^{\mathrm{R}}$. Each indicator variable $I_{j,i}^{\mathrm{R}}$ represents the event arm $j$ is pulled at an epoch $i \in [k]$ due to random exploration. Moreover, each indicator variable $I_{j,i}^{\mathrm{R}}$ is a Bernoulli random variable with parameter $p= \frac{\vep_i}{m}$ for $i \in [k]$, so that $\mb{E}[T_j^{\mathrm{R}}(k)] = \sum_{i=1}^k \frac{\vep_i}{m}$ and $\mb{E}[T_j^{\mathrm{R}}(k)] = 2x_0$. The variance of a Bernoulli random variable is upper bounded by the expected value, implying that $\text{Var}(T_j^{\mathrm{R}}(k)) \leq \mb{E}[T_j^{\mathrm{R}}(k)] = 2x_0$. Leveraging the preceding analysis, we apply the Bernstein inequality from Proposition~\ref{prop:bernstein} to $P(T_j^{\mathrm{R}}(k)\leq x_0)$. This gives
\begin{equation*}
P\big(T_j^{\mathrm{R}}(k)\leq x_0 \big) = P\big(T_j^{\mathrm{R}}(k)\leq 2x_0 - x_0 \big) = P\big(T_j^{\mathrm{R}}(k) - \mb{E}[T_j^{\mathrm{R}}(k)]\leq - x_0 \big) \leq \exp\Big(-\frac{x_0}{5}\Big).
\end{equation*}
Hence, 
\begin{align*}
 P\Big(\barR{T_j^{\alpha}(k)}{j} \geq \mu_j + \frac{\Delta_j}{2}\Big) 
& \leq x_0\exp\Big(-\frac{x_0}{5}\Big)+\frac{c''}{\Delta_j^2}\exp\Big( -\frac{\Delta_j^2\lfloor x_0\rfloor}{c''} \Big).
\end{align*}
Equivalent analysis can be used to show that 
\begin{align*}
 P\Big(\barR{T_\ast^{\alpha}(k)}{\ast} \leq \mu_\ast - \frac{\Delta_j}{2}\Big) 
& \leq x_0\exp\Big(-\frac{x_0}{5}\Big)+\frac{c''}{\Delta_j^2}\exp\Big( -\frac{\Delta_j^2\lfloor x_0\rfloor}{c''} \Big).
\end{align*}
Relating back to~\eqref{eq:union_split}, we obtain
\begin{equation*}
P\big(\barR{T_j^{\alpha}(k-1)}{j} \geq \barR{T_{\ast}^{\alpha}(k-1)}{\ast}\big)\leq 2x_0\exp\Big(-\frac{x_0}{5}\Big)+\frac{2c''}{\Delta_j^2}\exp\Big( -\frac{\Delta_j^2\lfloor x_0\rfloor}{c''} \Big).
\end{equation*}
Substituting the above inequality into~\eqref{eq:prob_suboptimal}, we have
\begin{equation*}
P(\alpha(k)=j) \leq \frac{c}{d^2k} + 2x_0\exp\Big(-\frac{x_0}{5}\Big)+\frac{2c''}{\Delta_j^2}\exp\Big( -\frac{\Delta_j^2\lfloor x_0\rfloor}{c''} \Big).
\end{equation*}
Finally, the lower bound constructed for $x_0$ in~\eqref{eq:x0_lower_bound} yields 
\begin{align*}
P(\alpha(k)=j) \leq \frac{c}{d^2k}&+\Big(\frac{2c}{d^2}\log \Big(\frac{(k-1)d^2\exp(1/2)}{cm} \Big)\Big)\Big( \frac{cm}{(k-1)d^2\exp(1/2)}\Big)^{c/(5d^2)}\\
&+\Big(\frac{2c''\exp(1)}{d^2}\Big)\Big( \frac{cm}{(k-1)d^2\exp(1/2)} \Big)^{c/c''}.
\end{align*} 
\end{proof}

%% file: Appendix_Sections/experiments.tex
In this section, we provide details on the continuous reinforcement learning approach that we compare with our proposed algorithms. For a complete treatment, we refer the reader to the works of~\citet{melo2008analysis} and~\citet{geramifard2013tutorial}. The $Q$-function is approximated by $Q(x, a)= w^T\phi(x, a)$ where $w$ is the weight vector, $x$ is the feature vector, and $a$ is the action. The loss function is then given by $L(w) = \frac{1}{2}(Q^+(x, a) - Q(s, a))^2$ where $Q^+(x, a) = r + \gamma \max_{a'}Q(x', a')$. To update the weights, gradient descent is used on the loss function so that we have $w_{k+1} = w_k -\alpha \nabla L(w)$ where $\alpha$ is the step size. We allow the reinforcement learning algorithm to use the state distribution $\beta_{t_k}$ as the features for the model and $\phi(\beta_{t_k}, a)$ is a vector of dimension $|\Theta|m$ which maps $\beta_{t_k}$ into coordinates reserved for action $a$ and makes each other coordinate have value zero. For our simulations, we use a step size of $1/\sqrt{k}$ where $k$ is the round. The policy that the reinforcement learning algorithm follows is an $\vep$--greedy policy, meaning that at each iteration $\max_a Q(x, a)$ is played with probability $1-\vep$ and a random action is played with probability $\vep$. We exponentially decayed $\vep$ to give sufficient opportunity to explore and then exploit. Specifically, we decayed $\vep$ so that it was $< .05$ after half of the horizon.

There are several reasons why we consider comparing to this form of reinforcement learning. To begin with, we note that it would be an extremely unfair comparison between discrete $Q$-learning that was able to observe the precise state---as opposed to state distribution---and the bandit algorithms. This is because the discrete $Q$-learning algorithm would then just be solving a Markov decision process and converge to the optimal state-dependent policy in the limit. For this reason we were interested in a reinforcement learning approach that can handle partially observable problems. Indeed, we can consider the state distribution as a partial state observation since the state will be drawn from this distribution, but the reinforcement learning algorithm is not privy to the exact state. Linear function approximation is a natural choice given that the expected stationary distribution reward is a linear combination of the stationary distribution rewards and the stationary distribution.

%% file: Appendix_Sections/commentary.tex
In this section, we revisit our discussion from Section~\ref{sec:regretdecomp} on why studying algorithms employing constant length epochs would be problematic for bandit problems in a correlated Markovian environment. The purpose of this discussion is to provide reasoning beyond motivating applications for analyzing algorithms based on an arm-dependent linearly increasing epoch length sequence, as well as to give theoretical justification for the failure of traditional bandit algorithms that ostensibly employ epochs of a fixed duration equal to one. In the final step to obtain a bound on the Markovian regret penalty for each regime of the discount factor $\gamma$, we were able to evaluate or bound the sums in~\eqref{eq:c1sum}--\eqref{eq:preharmonic} to terms that approach a constant or grow only logarithmically in the time horizon. However, a constant epoch length would make the summands of~\eqref{eq:c1sum}--\eqref{eq:preharmonic} constant. Consequently, each respective sum would be of the form $\sum_{i=1}^{n} c$ where $c > 0$ represents the constant summand, so that each respective sum would have a linear dependence on the time horizon. This means that the bounds on the Markovian regret penalty appearing in Proposition~\ref{thm:regretdecomp1} would grow linearly in the time horizon and we could no longer hope to obtain sublinear regret bounds regardless of the algorithm. Intuitively, this is to be expected: since the regret benchmark is with respect to the stationary distribution rewards and, owing to the correlated nature of the problem, any algorithm that is not guaranteed to select an arm repeatedly may never guide the state distribution toward a stationary distribution. In a similar manner, epochs of fixed duration would lead to problems when considering specific choices of algorithms and attempting to bound either the probability of playing each suboptimal arm or the expected number of times each suboptimal arm is played. This is because if the observed rewards are not guaranteed to approach the stationary distribution rewards, then attempting to discriminate between the stationary distribution reward of each arm would be infeasible.